\documentclass[12pt]{article}

\usepackage[top=1in,bottom=1in,left=1in,right=1in]{geometry}
\usepackage{tcolorbox}
\usepackage[colorlinks, linkcolor=blue, citecolor=blue]{hyperref}           
\usepackage{graphicx,dsfont}
\usepackage{subcaption}
\usepackage{amsmath}
\usepackage{textcomp, gensymb}
\usepackage[numbers, square, sort&compress]{natbib} 
\usepackage{tikz}
\usepackage{pgfplots}
\usepackage{pgf-pie}
\usepackage{mathrsfs}  
\usepackage{amssymb}
\usepackage{cancel}
\usepackage{tabularx} 
\usepackage{enumitem}
\usepackage{physics}
\usepackage{import}
\usepackage{pdfpages}
\usepackage{mathtools}
\usepackage{xparse}
\usepackage{bm}
\usepackage{amsthm}
\usepackage[colorlinks, linkcolor=blue, citecolor=blue]{hyperref}  
\usepackage{amsthm,wrapfig,comment}
\usepackage{booktabs,siunitx}

\definecolor{amber}{rgb}{1.0,0.75,0.0}
\definecolor{americanrose}{rgb}{1.0,0.01,0.24}

\makeatletter
\renewcommand{\paragraph}{%
  \@startsection{paragraph}{4}%
  {\z@}{0ex \@plus 1ex \@minus .2ex}{-1em}%
  {\normalfont\normalsize\bfseries}%https://www.overleaf.com/project/605b4eba36e1e271f7c1dfa9
}
\makeatother

\usepackage{soul}
\usepackage{algorithm}
\usepackage[noend]{algpseudocode}

\usepackage{tikz}
\usetikzlibrary{shapes,arrows}
\usepackage{verbatim}

\tikzstyle{block} = [draw, fill=blue!20, rectangle, 
    minimum height=3em, minimum width=6em]
\tikzstyle{sum} = [draw, fill=blue!20, circle, node distance=1cm]
\tikzstyle{input} = [coordinate]
\tikzstyle{output} = [coordinate]
\tikzstyle{pinstyle} = [pin edge={to-,thin,black}]

\usetikzlibrary{positioning,arrows}

\usetikzlibrary{calc}

\newcommand{\cmmnt}[1]{\ignorespaces}

\newcommand{\wasim}[1]{\textcolor{red}{[Wasim: #1]}}

\newcommand{\daniel}[1]{\textcolor{red}{[Daniel: #1]}}

\newcommand{\R}{\mathbb{R}}

\newcommand{\Rmnum}[1]{\expandafter\@slowromancap\romannumeral #1@}
\newcommand{\s}[1]{\mathsf{#1}}
\newcommand{\p}[1]{\left(#1\right)}
\newcommand{\pp}[1]{\left[#1\right]}
\newcommand{\ppp}[1]{\left\{#1\right\}}

\newcommand{\calG}{\mathcal{G}} %graph
\newcommand{\calV}{\mathcal{V}} %set of all vertices in the graph 
\newcommand{\calE}{\mathcal{E}} %set of all edges in the graph
\newcommand{\calP}{\mathcal{P}} %set of all paths from source to an observation
\newcommand{\calH}{\mathcal{H}} %Hypothesis
\newcommand{\calI}{\mathcal{I}} %Information type
\newcommand{\calF}{\mathcal{F}} %Martingale filtration
\newcommand{\calS}{\mathcal{S}} %scan test
\newcommand{\calA}{\mathcal{A}} 
 
\newcommand{\calT}{\mathcal{T}} 
\newcommand{\calD}{\mathcal{D}}

\newcommand{\calZ}{\mathcal{Z}}
\newcommand{\calL}{\mathcal{L}}

\newcommand{\bE}{\mathbb{E}}

\newtheorem{definition}{Definition}

\newtheorem{theorem}{Theorem}

\newtheorem{lemma}{Lemma}

\newenvironment{fminipage}%
  {\begin{Sbox}\begin{minipage}}%
  {\end{minipage}\end{Sbox}\fbox{\TheSbox}}

\NewDocumentCommand{\E}{e{_}e{^}g}{%
	\operatorname{\mathbb{E}}%
	\IfValueT{#1}{_{#1}}
	\IfValueT{#2}{^{#2}}%
	\IfValueT{#3}{\pp{#3}}%
}
\RenewDocumentCommand{\P}{e{_}e{^}g}{%
	\operatorname{\mathbb{P}}%
	\IfValueT{#1}{_{#1}}
	\IfValueT{#2}{^{#2}}%
	\IfValueT{#3}{\p{#3}}%
}

\RenewDocumentCommand{\var}{e{_}e{^}g}{%
	\operatorname{\mathbb{V}\textnormal{ar}}%
	\IfValueT{#1}{_{#1}}
	\IfValueT{#2}{^{#2}}%
	\IfValueT{#3}{\p{#3}}%
}

\makeatletter
\def\thanks#1{\protected@xdef\@thanks{\@thanks
        \protect\footnotetext{#1}}}
\makeatother

\allowdisplaybreaks

\DeclareMathOperator*{\argmax}{arg\,max}

\pgfplotsset{compat=1.18}

\begin{document}
\newtheorem*{my_corollary}{Corollary}

%\title{Online Multiclass Classification of Information Flow}
%\title{Sequential Multiclass Classification of Information Flow}
\title{Sequential Classification of Misinformation}

\author{Daniel Toma~~~~~~~~~Wasim Huleihel\thanks{D. Toma and  W. Huleihel are with the Department of Electrical Engineering-Systems at Tel Aviv university, {T}el {A}viv 6997801, Israel (e-mails:  \texttt{danielto@mail.tau.ac.il, vered.azr@gmail.com, wasimh@tauex.tau.ac.il}). This work is supported by the ISRAEL SCIENCE FOUNDATION (grant No. 1734/21).}}
%\title{Sequential Misinformation Classification}
%\title{Sequential Classification of Misinformation Spread}
%\title{Optimal Sequential Detection of Misinformation Spread}
\maketitle
%
%\sloppy

\begin{abstract}

In recent years there have been a growing interest in online auditing of information flow over social networks with the goal of monitoring undesirable effects, such as, misinformation and fake news. Most previous work on the subject, focus on the binary classification problem of classifying information as fake or genuine. Nonetheless, in many practical scenarios, the multi-class/label setting is of particular importance. For example, it could be the case that a social media platform may want to distinguish between ``true", ``partly-true", and ``false" information. Accordingly, in this paper, we consider the problem of online multiclass classification of information flow. To that end, driven by empirical studies on information flow over real-world social media networks, we propose a probabilistic information flow model over graphs. Then, the learning task is to detect the label of the information flow, with the goal of minimizing a combination of the classification error and the detection time. For this problem, we propose two detection algorithms; the first is based on the well-known multiple sequential probability ratio test, while the second is a novel graph neural network based sequential decision algorithm. For both algorithms, we prove several strong statistical guarantees. We also construct a data driven algorithm for learning the proposed probabilistic model. Finally, we test our algorithms over two real-world datasets, and show that they outperform other state-of-the-art misinformation detection algorithms, in terms of detection time and classification error.
\end{abstract}

\section{Introduction}\label{section:intro}

As social media gained popularity, the spread of misinformation and disinformation in social networks has become a widespread phenomenon. While some of this spread may be due to innocent errors or differing opinions, some is deliberately intended to manipulate individuals' beliefs and behavior, often to serve economic or political agendas. Therefore, it is considered a significant societal concern \cite{Harsin2018PostTruthAC}. 

One promising branch of solutions to the above issues is the development of automated misinformation detection algorithms \cite{Harsin2018PostTruthAC}. These algorithms are primarily based on binary classifiers that categorize news items and user posts as either ``real" or ``fake". However, this binary view of misinformation overlooks the complex nature of the problem. For instance, in \cite{Arendt2005-ARETAP}, true information is contrasted not only with deliberate lies but also with opinions, and \cite{iq} proposes various taxonomies of disinformation, one of which includes lies, spins, and bullshit.

Classifying the type of misinformation/disinformation is not merely a philosophical exercise; it has real-world applications. For instance, a social media platform could respond differently to various types of misinformation. Pop culture misinformation, such as ``Titanic 3 will be filmed in New Zealand", is generally harmless, while other types of false information could be used to manipulate readers. Therefore, the problem we aim to address is the multi-class classification of misinformation. Since early detection is critical in mitigating the spread of misinformation, we are particularly interested in quickest sequential multi-class classification of misinformation.

\paragraph{Related work.} The problem of misinformation classification has been widely studied. Early attempts to tackle this issue suggested using a set of predefined features to identify misinformation \cite{Kwon13}, while other methods focused solely on the linguistic content of news items \cite{10.1007/s11042-020-10183-2, DBLP:journals/corr/RiedelASR17}. A significant improvement came with the introduction of graph neural networks (GNNs) \cite{DBLP:journals/corr/KipfW16}, which learn node representations by recursively embedding a node's features with those of its neighbors. These models can be applied to both node classification and whole graph classification \cite{DBLP:journals/corr/abs-1810-00826}. User preference-aware fake news detection (UPFD) \cite{dou2021user} and graph convolutional networks fake news (GCNFN) \cite{DBLP:journals/corr/abs-1902-06673} are examples of GNN-based models used to classify propagation graphs. Specifically, GCNFN employs profile features as node inputs and utilizes a graph convolutional networks (GCN) based model for classification \cite{DBLP:journals/corr/KipfW16}, while UPFD incorporates the content of the news item by encoding it using NLP techniques like BERT \cite{devlin2019bertpretrainingdeepbidirectional} or SpaCy \cite{honnibal_spacy_2018}, and has three different variations that are based on GCN, graph attention networks (GAT) \cite{veličković2018graphattentionnetworks}, and GraphSage convolutional layers \cite{DBLP:journals/corr/HamiltonYL17}. Although these models were not specifically designed for early detection, they can be employed during the propagation phase.

Another group of methods focuses on early detection of fake news. Quickstop \cite{wei2019quickstop} models the temporal series of user profiles sharing a news item as a Markov chain and applies the sequential probability ratio test (SPRT) rule \cite{10.1214/aoms/1177730197} to reach a decision. Similarly, \cite{orenloberman2023online} uses a more complex Markovian modeling of user profiles and applies the SPRT decision rule for early detection. Both approaches present the problem as a Bayesian optimization problem, aiming to minimize both propagation time and the probability of misclassification \cite{bayesian}. Finally, Heterogeneous graph GNNs represent another set of solutions. Hierarchical graph attention network (HGAT) \cite{DBLP:journals/corr/abs-2002-04397}, for example, incorporates both news and user nodes, connecting them when a user shares a specific news item. A related method, hypergraph for fake news detection (HGFND) \cite{10020234}, similarly models news items as nodes connected by hyper-edges \cite{DBLP:journals/corr/abs-1809-09401}, which link sets of nodes when the same users share the news items around the same time or mention the same entities. While these models achieve state-of-the-art accuracy in real-world datasets, they require extensive preprocessing, making them unsuitable for early detection of misinformation. Furthermore, none of these methods address the challenge of classifying misinformation into multiple classes.

\paragraph{Main contributions.} Following \cite{wei2019quickstop,oren2023onlineconf}, in this paper, we suggest a real-time multiclass misinformation detection framework, based on a certain Markovian probabilistic information spreading model over a social network modeled by a graph. Under this framework, we derive two data-driven and model-driven algorithms, for online multiclass misinformation detection. Specifically, our first algorithm is based on the well-known multiple SPRT (MSPRT) decision rule \cite{340472}, for sequential classification of in a multi-class setting. Classical theory of MSPRT typically relies on simple observational models, where the data is generated as independent and identically distributed (i.i.d.). In our case, we deviate from this assumption, and prove several theoretical statistical guarantees under the Markovian model mentioned above; among those results are high probability upper bounds on the stopping/decision times, and misclassification error probabilities. As it turns out, while the MSPRT enjoys desirable statistical properties, it suffers from a relatively high sample complexity, it is quite sensitive to errors resulted during the estimation of the model parameters, and it is not robust to model mismatch. As a remedy for these problematic issues, we propose also a GNN based sequential decision rule, which proves quite robust to the type of errors mentioned above, and at the same time, provably shares similar statistical guarantees as the MSPRT. Finally, we compare our algorithms against several state-of-the-art misinformation detection techniques over real-world datasets, and show the superiority of our algorithms in terms of accuracy and detection time.

\paragraph{Notation.} We use the calligraphic font to indicate sets, and sans serif font with uppercase and lowercase letters, e.g., $\s{X}$ and $x$, to indicate random variables and their realizations, respectively. We let $\P(\cdot)$ and $\E{\cdot}$ indicate the probability and expectation functions. We denote by $\mathds{1}_{\calE}$ the indicator function that gets $1$ when an event $\calE$ is true and $0$, otherwise. We denote the cardinality of some set $\calS$ by $\abs{\calS}$, and for a non-negative integer $M$ we let $[M]=\{0,1,\ldots,M-1\}$. For a set $\mathcal{X}$, we let $\mathcal{X}^n$ denote the $n$-fold Cartesian product of $\mathcal{X}$. An element of $\mathcal{X}^n$ is denoted by $x^n =(x_1,x_2,\ldots,x_n)$. A substring of $x^n\in\mathcal{X}^n$ is designated by $x_i^j = (x_i, x_{i+1},\ldots,x_j)$, for $1\leq i \leq j \leq n$; when $i = 1$, the subscript is omitted. A directed walk is a finite or infinite sequence of edges directed in the same direction that joins a sequence of vertices. Let $\calG = (\calV, \calE)$ be a directed graph. A finite directed walk is a sequence of edges $e_1,e_2,\ldots,e_{n-1}$ for which there is an associated sequence of vertices $(v_1,v_2,\ldots,v_n)$ such that $e_i = (v_i,v_{i+1})$, for $i = 1,2,\ldots,n-1$. The sequence $(v_1,v_2,\ldots,v_n)$ is the vertex sequence of the directed walk. A directed path is a directed walk in which all vertices and edges are distinct.

\section{Problem Definition} \label{section:StatisticalModel}

In this section, we present the problem of multi-label misinformation detection. We start by describing the underlying probabilistic information flow model, and then formulate mathematically the inference problem we aim to solve in this paper.

\subsection{Information flow model}\label{subsec:infflow}

%\subsubsection{Underlying graph} 

\paragraph{Information traces and decisions.}
%We follow a similar setting to the one in \cite{wei2019quickstop,orenloberman2023online}. 

Let $\calG = (\calV, \calE)$ be a directed graph, where $\calV = [n]$ is the set of nodes and $\calE$ is the set of directed edges. One can think of this graph as representing a certain social network platform, where the nodes are ``users" and the edges are the ``connections" between those users. We associate each node $u\in\calV$ with a $d$-dimensional ``feature" vector $\mathbf{x}_u\in \mathbb{R}^d$; commonly used features are, for example, user profile, user social engagements, etc. The platform goal is to monitor the spread of a message (or, ``piece of information"), denoted by $\calI$, and assumed to be categorized/labeled into/by $M\geq2$ values, i.e., $\calI\in[M]$. It is convenient to think of $\calI=0$ as representing completely genuine information, while $\calI\in[M]\setminus\{0\}$ represents different degrees of in-genuine (or, fake) information. The fact that $M\geq2$ allows for ``partially true" information, or, for different types of fake information, e.g., fake celebrity news versus fake political news. Given a directed edge $e = (u,v)$, we shall refer to user $v$ as a \emph{follower} of user $u$, while user $u$ is a \emph{followee} of user $v$. Over $\calG$, user $v\in\calV$ can forward $\calI$ from user $u \in\calV$. Although implicitly in our model, user $u$ decides whether to forward $\calI$ or not based on: (i) the information type $\mathcal{I} \in [M]$; (ii) its features $\mathbf{x}_u$; and (iii) the set of its neighbors $\mathcal{N}_u \triangleq \{v \in \calV: (u , v) \in \calE\}$, who forwarded the information earlier. 

Given the above setup for the underlying network, we next define the concept of \emph{information traces}, i.e., the way information flows over the network. We focus on a single information source $s\in\calV$, and we let $\calP$ denote the set of all possible directed paths in $\calG$ starting at $s$. For any path $\s{P}\in\calP$ in $\calG$, let its vertex sequence be denoted by $\s{P} = (v_{1}^{\s{P}},v_{2}^{\s{P}},\ldots,v_{|\s{P}|}^{\s{P}})$. The information $\calI$ can flow over one or more of these paths. Generally speaking, we say that an event occurs when a follower forwards the information from one of its followees. This is represented by the edge over which this event occurs, which in turn is modeled by a pair of features $(\mathbf{x}_u,\mathbf{x}_v)$ that correspond to the end users. We treat $\ell\in\mathbb{N}$ as the time index, and at the $\ell$th time step we denote by $(u_\ell,v_\ell)$ the followee-follower pair over which an event happened. Then, the \emph{information trace} is the sequence $\{(\mathbf{x}_{u_\ell},\mathbf{x}_{v_\ell})\}_{\ell\geq1}$, where $(u_\ell,v_\ell)\in\calE$, and $v_{\ell}$ is a child of $u_{\ell}$, i.e., if $u_{\ell} = v_i^{\s{P}}$ for some $i\in[|\s{P}|]$, then, $v_{\ell} = v_{i+1}^{\s{P}}$. An illustration of possible information traces over $\calG$ can be seen in Fig.~\ref{Fig:GraphExmp}. 

\begin{figure*}[ht!]
\centering
\begin{tikzpicture}[thick,black!70!black]
\draw [-stealth,dashed,thick](-5.2,0) -- (-1.5,0) node[below,black,xshift=-1.8cm] {$\s{Information}\;\s{trace}$};
\node[text width=3cm] at (-3,0.4) 
    {$\{(\mathbf{x}_{u_\ell},\mathbf{x}_{v_\ell})\}_{\ell\geq1}$};
\node[cloud, draw,fill = gray!10,aspect=2,
    cloud puffs = 16] (c) at (0,0) {$\s{Black}\;\s{Box}$};
\draw [-stealth,dashed,thick](1.5,0) -- (4.1,0) node[below,black,xshift=-1.2cm] {$\s{Decision}$};
\node[text width=3cm] at (4.3,0.4) 
    {$\hat{\calI}$};
\end{tikzpicture}  
\caption{An illustration of the input-output information flow detection problem.}
\label{fig:blackBox}
\end{figure*}
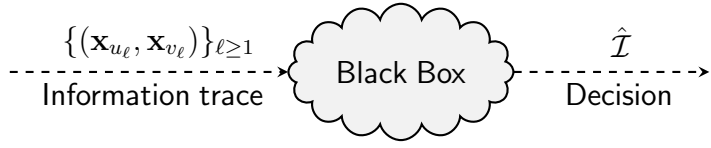

Given information traces our goal is to infer the message $\calI$, as illustrated in Fig.~\ref{fig:blackBox}, where ``black box" refer to any probabilistic modeling and algorithmic operations we apply for inference. The general meta question we would like to answer is:
\begin{quote}
   \emph{How many information traces over $\calG$ we should see, and how lengthy should they be, so as to infer the message $\calI$ with ``high confidence"?}
\end{quote}
We will formalize this question as a sequential multiple hypothesis testing problem, and to that end, we will consider two probabilistic information spreading models; later on, will show how these models can be extracted from actual data. 

%\subsubsection{Information traces and decisions}

\paragraph{Edge/Node compressed representations.} As mentioned above, our inference algorithms are driven by certain probabilistic models for the information flow. As will be seen in the sequel, these models basically place probabilistic/statistical assumptions on compressed node/edge representations of feature-pairs over the information traces. We will refer to these representations as the \emph{edge-based} and \emph{node-based} models, and we next define those in detail. 

\begin{figure}[t!]
    % \centering
    \captionsetup[subfigure]{justification=centering}
   % \centering
     \begin{subfigure}[b]{0.3\textwidth}
         %\centering
         \begin{tikzpicture}[scale=0.5]
  \begin{axis}[
        ybar,
        axis on top,
        height=8cm, width=15.5cm,
        bar width=0.4cm,
        ymajorgrids, 
        xminorgrids,
        yminorgrids,
        tick align=inside,
        major grid style={line width=0.5pt,draw=gray!50},
        minor grid style={line width=0.2pt,draw=gray!20},
        enlarge y limits={value=.1,upper},
        ymin=0, ymax=100,
        axis x line*=bottom,
        axis y line*=left,
        y axis line style={opacity=0},
        tickwidth=0pt,
        enlarge x limits=true,
        legend style={
            at={(0.5,-0.2)},
            anchor=north,
            legend columns=-1,
            /tikz/every even column/.append style={column sep=0.5cm}
        },
        ylabel={Frequency (\%)},
        xlabel={$\s{Z}$},
        symbolic x coords={0, 1, 2, 3, 4, 5},
        xtick=data,
        minor x tick num=1,
        legend image code/.code={
        \draw [#1] (0cm,-0.1cm) rectangle (0.2cm,0.25cm); },
    ]
    \addplot [draw=none, fill=blue!80] coordinates {(0,79.15) (1,0.) (2,17.58) (3,0.0) (4,2.88) (5,0.00)};
    \addplot [draw=none, fill=red!80] coordinates {(0,26.11) (1,0.00) (2,73.89) (3,0.00) (4,0.00) (5,0.00)};
    \addplot [draw=none, fill=black!70] coordinates {(0,0.0) (1,2.66) (2,4.16) (3,1.83) (4,90.53) (5,0.0)};

    \legend{True News, Fake Gossip News, Fake Politics News}
  \end{axis}
\end{tikzpicture}
         \caption{3-class UPFD dataset.}
         \label{fig:upfd_edge_histogram}
     \end{subfigure}
%\hfill
\hspace{3cm}
\begin{subfigure}[b]{0.3\textwidth}
       %  \centering
         \begin{tikzpicture}[scale=0.5]
  \begin{axis}[
        ybar, axis on top,
        height=8cm, width=15.5cm,
        bar width=0.4cm,
        ymajorgrids, 
        xminorgrids,
        yminorgrids,
        tick align=inside,
        major grid style={line width=0.5pt,draw=gray!50},
        minor grid style={line width=0.2pt,draw=gray!20},
        enlarge y limits={value=.1,upper},
        ymin=0, ymax=100,
        axis x line*=bottom,
        axis y line*=left,
        y axis line style={opacity=0},
        tickwidth=0pt,
        enlarge x limits=true,
        legend style={
            at={(0.5,-0.2)},
            anchor=north,
            legend columns=-1,
            /tikz/every even column/.append style={column sep=0.5cm}
        },
        ylabel={Frequency (\%)},
        xlabel={$\s{Z}$},
        symbolic x coords={0, 1, 2, 3, 4, 5},
        xtick=data,
        minor x tick num=1,
        legend image code/.code={
        \draw [#1] (0cm,-0.1cm) rectangle (0.2cm,0.25cm); },
    ]
    \addplot [draw=none, fill=blue!80] coordinates 
    {(0,73.19) (1,11.18) (2,5.72) (3,4.97) (4,1.24) (5,3.71)};
    \addplot [draw=none, fill=red!80] coordinates 
    {(0,42.96) (1,11.76) (2,10.16) (3,17.97) (4,3.16) (5,14.00)};
    \addplot [draw=none, fill=black!70] coordinates 
    {(0,8.33) (1,12.63) (2,3.09) (3,11.79) (4,8.78) (5,55.38)};

    \legend{True News, Fake Subjective News, Fake Objective News}
  \end{axis}
\end{tikzpicture}
         \caption{3-class Weibo dataset.}
         \label{fig:weibo_edge_histogram}
     \end{subfigure}
        \caption{Distribution of edge types across different hypotheses.}
        \label{fig:ZDistribution}
\end{figure}
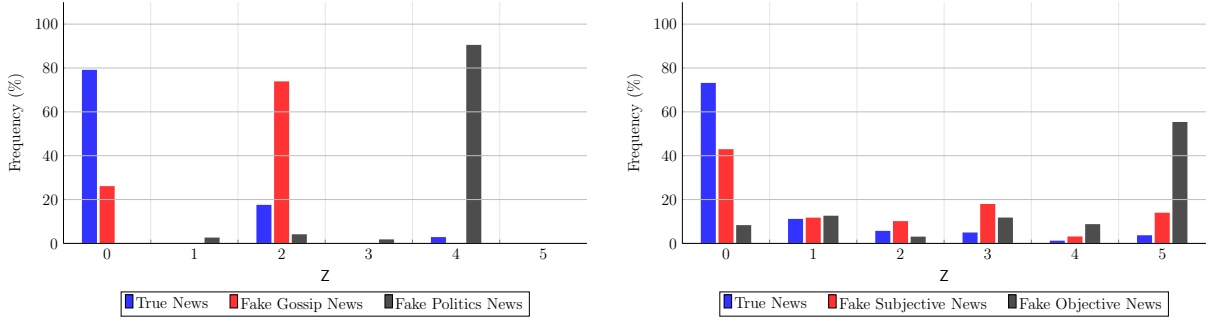

%The edge-based model, views each edge $e\in \calE$ as a communication channel that is associated with a given weight, which we denote by $\s{W}_{u,v}\triangleq \s{W}_e\in\calZ$, where $\calZ\triangleq\{0,1,\ldots,|\calZ|-1\}$, is the set of possible edge types. Intuitively speaking, we think of edges with weight ``$0$" as those who are more likely to spread genuine information, while edges with weight $|\calZ|-1$ are those who are more likely to spread fake information. Accordingly, we assume that $\s{W}_e = g(\mathbf{x}_u , \mathbf{x}_v)$, where $e \triangleq (u,v) \in \calV^2$, and $g:\calE\to[|\calZ|]$ is the edge-classifier function. 
The edge-based model, views each edge $e\in \calE$ as a communication channel that is associated with a given weight/type, which we denote by $\s{W}_{u,v}\triangleq \s{W}_e\in\calZ$, where $\calZ\subseteq\mathbb{R}_{+}$. For example, when $\calZ$ is finite, e.g., $\calZ=[|\calZ|]$ with $|\calZ|<\infty$, then, intuitively, we can think of edges with weight ``$0$" as those which are more likely to spread genuine information, while edges with weight $|\calZ|-1$ are those which are more likely to spread fake information. Accordingly, we assume that $\s{W}_e = g(\mathbf{x}_u , \mathbf{x}_v)$, where $e \triangleq (u,v) \in \calV^2$, and $g:\calE\to\calZ$ is the ``edge-classifier" function. To motivate the edge-based model above, we consider the following two experiments over real-world databases. Specifically, Fig.~\ref{fig:upfd_edge_histogram} shows the distribution of edge-types as a function of the edge type, under each one of the hypotheses, in the user preference-aware fake news detection (UPFD) \cite{dou2021user} dataset. Here, there are three classes whose types correspond to ``true news", ``fake gossip news", and ``fake politics news". Similarly, Fig.~\ref{fig:weibo_edge_histogram} shows a similar histogram for the three-class Weibo dataset \cite{10.5555/3061053.3061153}, in which the classes correspond to ``true news", ``fake subjective news", and ``fake objective news". We see a positive correlation between the hypothesis and its edge-type.
%illustrates that the scores of edges involved in spreading genuine information are concentrated around zero, while Fig. 1b shows that the scores of edges involved in spreading fake information are concentrated around one. This is consistent with the fact that an edge with a higher weight is more likely to spread fake information

Next, we move forward to the node-based model. The underlying assumption here is that given $\calI$, the probability that a certain user spreads $\calI$ to its neighbors depends on its own features and the features of his neighbors. Accordingly, in contrast to the edge-model above, here each node is associated with a weight/type, i.e., users are classified based on their features. Intuitively, an honest user has a lower probability to spread misinformation than a malicious user. Mathematically, each node $u\in\calV$ is associated with a given weight, denoted by $\bar{\s{W}}_{u}\in\bar{\calZ}\subseteq\mathbb{R}$. For example, when $\bar{\calZ}\triangleq[|\bar{\calZ}|]$ is finite, then, intuitively speaking, we think of nodes with weight ``$0$" as those which are more likely to spread genuine information, while nodes with weight $|\bar{\calZ}|-1$ are those which are more likely to spread fake information. As for the edge-type model, we assume that $\bar{\s{W}}_{u} = \bar{g}(\mathbf{x}_u,\{\mathbf{x}_v\}_{v\in\mathcal{N}_u})$, where $\bar{g}:\calV\to\bar{\calZ}$ is the ``node-classifier" function. Later on, we will explain how the edge and node classifiers $g$ and $\bar{g}$ above, can be generated from data. In our problem, the inference algorithm declares its decision on $\calI$ based on the edge/node weights representations, i.e., $\{\s{W}_{u_\ell,v_{\ell}}\}_{\ell\geq1}$ or $\{\bar{\s{W}}_{u_\ell}\}_{\ell\geq1}$, respectively; these can be thought as compressed versions of the information trace $\{(\mathbf{x}_{u_\ell},\mathbf{x}_{v_\ell})\}_{\ell\geq1}$. 

\paragraph{Probabilistic edge/node paths.} Given the above setup for the information trace and its edge/node compressed representations, we now set a probabilistic information spreading model, using which we formulate the decision problem we aim to solve. Later on, we will explain how this probabilistic model can be extracted from data. For each possible path in $\calG$, we model the sequence of edges/nodes representations in the information trace as a union of \emph{Markov processes}. Specifically, under the edge-model, each path $\s{P}\in\calP$ in $\calG$ is associated with a first-order homogeneous Markov edge process $\s{W}^{\s{P}} \triangleq \{\s{W}_{v_{i}^{\s{P}},v_{i+1}^{\s{P}}}\}_{i=1}^{|\s{P}|-1}$. Similarly, under the node-model, each path $\s{P}\in\calP$ in $\calG$ is associated with a first-order homogeneous Markov node process $\bar{\s{W}}^{\s{P}} \triangleq \{\bar{\s{W}}_{v_{i}^{\s{P}}}\}_{i=1}^{|\s{P}|}$.

Given $\calI$, we let $\alpha_{\calI}(\s{z}\vert\s{z}')$ and $\bar{\alpha}_{\calI}(\bar{\s{z}}\vert\bar{\s{z}}')$ for $\s{z},\s{z}'\in\calZ$ and $\bar{\s{z}},\bar{\s{z}}'\in\bar{\calZ}$, denote the edge and node Markov transition kernels probabilities, respectively. Furthermore, for any neighbor of $s$, $v\in\mathcal{N}_s$, we define the initial edge probabilities $\eta_{\calI}(\s{z})\triangleq\P(\s{W}_{s,v}=\s{z}\vert\calI)$ and node probabilities $\bar{\eta}_{\calI}(\bar{\s{z}})\triangleq\P(\bar{\s{W}}_{s}=\bar{\s{z}}\vert\calI)$, for any $\calI\in[M]$, $\s{z}\in\calZ$ and $\bar{\s{z}}\in\bar{\calZ}$, respectively. Later on, we will devise a data-based offline algorithm to learn the transition kernels and initial probabilities from training. %Finally, with some abuse of notation, for a given path $\s{P}\in\calP$, we let $\s{W}_{u\to v}^{\s{P}}$ denote the trajectory of edge weights starting from user $u\in\calV$ and ending at user $v\in\calV$. 
Fig.~\ref{Fig:GraphExmp} gives an example of a social media graph with multiple possible Markov chains paths. 
%Given $\calI$, we let the transition probability kernel be $\mathcal{K}_\calI(\s{X}\in\s{A} \vert\s{X}')=\P(\s{X}\in \s{A} \vert\s{X}',\calI)$ for $\s{X},\s{X}'\in\mathbb{R}^d$, and $\s{A} \subseteq \mathbb{R}^d$ is a measureable set in $\mathbb{R}^d$. We can define the \textbf{transition density function} $\kappa_\calI(\s{X}, \s{X}') $ as the Radon-Nikodym derivative of the Markov kernel with respect to \(\s{X}\). Formally, the transition density function is given by,
%\begin{align}
%    \kappa_\calI(\s{X} \mid \s{X}') = \frac{\partial \mathcal{K}_\calI(\s{X} \in \s{A} \mid \s{X}')}{\partial \s{X}} \bigg|_{\s{X} \in \s{A}}.
%\end{align}
%This density function $\kappa(\s{X} \mid \s{X}')$ satisfies,
%\begin{align}
%    \mathcal{K}_\calI(\s{X} \in \s{A} \mid \s{X}') = \int_{\s{A}} p_\calI(\s{X} \mid \s{X}') \, d\s{X}.
%\end{align}
%Furthermore, for $s$ any neighbor of $s$, $v\in\mathcal{N}_s$, we define the initial probabilities $\kappa_\calI(\s{X}\in\s{A})\triangleq\P(\s{W}_s\in\s{A}\vert\calI)$.

%\input{edge_transition_probability_matrices_IEEE}

\subsection{Problem statement}

\begin{figure*}[t!]
    \centering
    \begin{tikzpicture}[
        every node/.style={circle, draw, thin, minimum size=8mm, scale=0.7},
        grow=down,
        level distance=1.4cm,
        edge from parent/.style={->, -latex, draw},
        edge/.style = {->, -latex, draw},
        level 1/.style={sibling distance=50mm},
        level 2/.style={sibling distance=17mm},
        level 3/.style={sibling distance=6.5mm}
    ]
    \node (1) {1}
        child {node (3) {3}
            child {node (9) {9}
                child {node (21) {21}}
                child {node (22) {22}}
            }
            child {node (8) {8}
                child {node (20) {20}}
                child {node (19) {19}}
            }
            child {node (7) {7}
                child {node (17) {17}}
                child {node (18) {18}}
                child {node (16) {16}}
            }
        }
        child {node (4) {4}
            child {node (11) {11}
                child {node (27) {27}}
                child {node (26) {26}}
                child {node (25) {25}}
            }
            child {node (10) {10}
                child {node (24) {24}}
                child {node (23) {23}}
            }
        }
        child {node (2) {2}
            child {node (6) {6}
                child {node (15) {15}}
            }
            child {node (5) {5}
                child {node (14) {14}}
                child {node (12) {12}}
                child {node (13) {13}}
            }
        }
        ;

        \draw[edge] (1) to (6);
        \draw[edge] (6) to (14);
        %\draw[edge] (6) to (16);
        \draw[edge] (7) to (19);
        %\draw[edge] (9) to (23);
        \draw[edge] (4) to (24);
        \draw[edge] (11) to (27);
        \draw[edge] (1) to (16);

        % Delete colored edges
        \draw[edge, white] (1) -- (2);
        \draw[edge, white] (2) -- (6);
        \draw[edge, white] (6) -- (14);
        \draw[edge, white] (1) -- (3);
        \draw[edge, white] (3) -- (7);
        \draw[edge, white] (7) -- (19);
        \draw[edge, white] (1) -- (23);
        \draw[edge, white] (1) -- (4);
        \draw[edge, white] (4) -- (24);

        % Color branch with rotated colors
        \draw[edge, blue, very thick] (1) -- (2)  node[midway, left, yshift=20pt, draw=none] {$\s{W}^{\s{P}_1}_{2}$};
        \draw[edge, blue, very thick] (2) -- (6) node[midway, right, yshift=10pt, xshift=-30pt, draw=none] {$\s{W}^{\s{P}_1}_{6}$};
        \draw[edge, blue, very thick] (6) -- (14) node[midway, left, yshift=8pt, xshift=30pt, draw=none] {$\s{W}^{\s{P}_1}_{14}$};
        
        \draw[edge, orange, very thick] (1) -- (3) node[midway, left, xshift=6pt, yshift=10pt , draw=none] {$\s{W}^{\s{P}_2}_{3}$};
        \draw[edge, orange, very thick] (3) -- (7) node[midway, left, yshift=8pt, xshift=30pt, draw=none] {$\s{W}^{\s{P}_2}_{7}$};
        \draw[edge, orange, very thick] (7) -- (19) node[midway, right, yshift=14pt, xshift=-25pt, draw=none] {$\s{W}^{\s{P}_2}_{19}$};

        \draw[edge, green, very thick] (1) -- (16) node[midway, right, xshift=-45pt , draw=none] {$\s{W}^{\s{P}_3}_{23,1}$};
        
        \draw[edge, red, very thick] (1) -- (4) node[midway, right, yshift=2pt, draw=none] {$\s{W}^{\s{P}_4}_{4}$} ;
        \draw[edge, red, very thick] (4) -- (24) node[midway, right, yshift=4pt, xshift=-35pt, draw=none] {$\s{W}^{\s{P}_4}_{24}$};
    \end{tikzpicture}
    \caption{A partial social media graph with a single information source at $s = 1$. Each weighted path in the graph corresponds to a different Markov chain.}
    \label{Fig:GraphExmp}
\end{figure*}
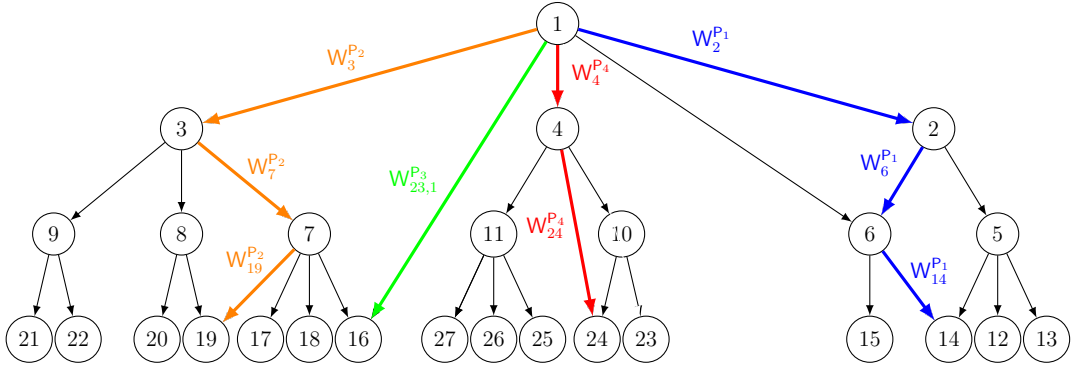

\paragraph{Learning problem.} 
We formulate the misinformation detection problem as a sequential multiple hypothesis testing problem. First, we define the type of observations available for testing. It is clear that when complete network and diffusion information are known, the information spreading trace forms a tree; we assume that its edge/node representation paths are distributed as first-order Markov chain. Recall that at the $\ell$th time step, we denote by $(u_\ell,v_\ell)$ the followee-follower pair over which an event happened (information spread). Then, we denote the sequence of edge observations by $\{\s{Z}_\ell\}_{\ell\geq1}$, where $\s{Z}_\ell = \s{W}_{u_\ell,v_\ell}^{\s{P}}$, for $(u_\ell,v_\ell)\in\calE$, and some path $\s{P}\in\calP$; in fact, over path $\s{P}$, the vertex $v_{\ell}$ must be a child of $u_{\ell}$, and thus, if $u_{\ell} = v_i^{\s{P}}$ for some $i\in[|\s{P}|]$, then, $v_{\ell} = v_{i+1}^{\s{P}}$. Similarly, we denote the sequence of node observations by $\{\bar{\s{Z}}_\ell\}_{\ell\geq1}$, where $\bar{\s{Z}}_\ell = \bar{\s{W}}_{v_\ell}^{\s{P}}$, for $v_\ell\in\calV$ and some path $\s{P}\in\calP$. Again, we would like to emphasize here the important fact that $\{\s{Z}_\ell\}_{\ell\geq1}$ ($\{\bar{\s{Z}}_\ell\}_{\ell\geq1}$) does not necessarily form a single first-order Markov processes, but is rather composed of multiple Markov processes, depending on the number of paths involved in forming $\{\s{Z}_\ell\}_{\ell\geq1}$ ($\{\bar{\s{Z}}_\ell\}_{\ell\geq1}$). In particular, keeping Fig.~\ref{Fig:GraphExmp} in mind, at the $\ell$th time step, we can represent the sequence of edge observations $\s{Z}_1^\ell$ as,
\begin{align}
    \s{Z}_1^\ell = \bigcup_{\s{P}\in\calP_\ell}\bigcup_{i\in[|\s{P}|]}\s{W}^{\s{P}}_{v_i^{\s{P}},v_{i+1}^{\s{P}}},\label{eqn:bigCupUnion}
\end{align}
where $\calP_\ell$ is a set of disjoint paths spanned by $\s{Z}_1^\ell$ in $\calG$. Accordingly, given $\calI$, the joint probability law of $\s{Z}_1^\ell$ is,
\begin{align}
    \P\p{\s{Z}_1^\ell\vert\calI} = \prod_{\s{P}\in\calP_\ell}\prod_{i\in[|\s{P}|]}\alpha_{\calI}(\s{W}_{v_i^{\s{P}},v_{i+1}^{\s{P}}}^{\s{P}}\vert\s{W}_{v_{i-1}^{\s{P}},v_{i}^{\s{P}}}^{\s{P}}).\label{eqn:jointprobLAW}
\end{align} 
To simplify notation, we define the $j$th edge-type in path $\s{P}$ as $\s{Z}_{j}^{\s{P}}\triangleq \s{W}_{v_j^{\s{P}},v_{j+1}^{\s{P}}}^{\s{P}}$. Furthermore, if the $\ell$th observation $\s{Z}_\ell$ is the $j$th edge in path $\s{P}\in\calP_\ell$, we define the ancestor of $\s{Z}_\ell$ as $\calA_\ell\triangleq\s{Z}_{j-1}^{\s{P}}$, namely, it is the observation that precedes $\s{Z}_\ell$ in its corresponding path. With these notations, \eqref{eqn:jointprobLAW} can be written as,
\begin{align}
     \P\p{\s{Z}_1^\ell\vert\calI}
    &= \prod_{\s{P}\in\calP_\ell}\prod_{i\in[|\s{P}|]}\alpha_{\calI}(\s{Z}_{i}^{\s{P}}\vert\s{Z}_{i-1}^{\s{P}})\\
    &=\prod_{i=1}^\ell \alpha_{\calI}(\s{Z}_{i}\vert\calA_i)\label{eqn:jointprobLAW2}.
\end{align}
For the node representation we have the same relations as in \eqref{eqn:bigCupUnion}--\eqref{eqn:jointprobLAW2}, but with $\s{Z}_1^\ell$, $\calA_\ell$, and $\alpha_\calI(\cdot\vert\cdot)$, replaced by $\bar{\s{Z}}_1^\ell$, $\bar{\calA}_\ell\triangleq\bar{\s{Z}}_{j-1}^{\s{P}}$, and $\bar{\alpha}_\calI(\cdot\vert\cdot)$, respectively. 

Our learning problem is formulated as follows: we are given an information trace sequence $\{(\mathbf{x}_{u_\ell},\mathbf{x}_{v_\ell})\}_{\ell\geq1}$, which induces an edge-representation sequence $\{\s{Z}_{\ell}\}_{\ell\geq1}$, and a node-representation sequence $\{\bar{\s{Z}}_{\ell}\}_{\ell\geq1}$ that obey one of the hypotheses. The audit is tasked with distinguishing between
\begin{align}\label{eqn:learning}
    &\calH_0:\calI=0\quad\quad\s{vs.}\quad\quad\calH_1:\calI=1 \quad\quad \cdots  \quad\quad\s{vs.}\quad\quad\calH_{M-1}:\calI=M-1.
\end{align}
We denote the prior probability of the $j$th hypothesis by $\pi_j$. The audit is tasked with distinguishing between the hypotheses above, in a way that minimizes a combination of the error probability and the propagation cost, as we define in the sequel. For the rest of this paper, the underlying probability space is $(\Omega,\calF,\mathbb{P}_\pi)$, where $\P_{\pi}$ is the probability measure defined as follows,
\begin{align}
    \P_{\pi} = \sum_{j=0}^{M-1} \pi_j\P_j,
\end{align}
with $\P_j$ being the probability measures under the $j$th hypothesis, namely, under $\P_{j}$, the sequence $\{\s{Z}_{\ell}\}_{\ell\geq1}$ ($\{\bar{\s{Z}}_{\ell}\}_{\ell\geq1}$) is Markovian with transition probabilities $\alpha_j(\cdot\vert\cdot)$ ($\bar{\alpha}_j(\cdot\vert\cdot)$).

At this point we would like to emphasize again that the edge and node representations, as well as the Markovian assumption associated with either one of these representations, is done only as part of an algorithmic solution for the decision problem we faced with. To wit, the input-output relation is exactly as described in Fig.~\ref{fig:blackBox}, with the inputs/observations being the information traces, while the edge/node representations are only an artificial algorithmic operations done as part of the ``black-box" relation which produce the decision.    

\paragraph{Sequential decision rule.} Starting with the edge-based representation, as mentioned above, assume we are in the situation where we observe $\{\s{Z}_{\ell}\}_{\ell\ge1}$ sequentially, generating the natural filtration $\{\calF_{\ell}\}_{\ell\ge1}$, with, 
\begin{align}
    \calF_{\ell} \triangleq \{\s{Z}_1, \s{Z}_2, \ldots , \s{Z}_{\ell}\},
\end{align}
and $\calF_0 \triangleq (\Omega, \emptyset)$. Let $\s{T} \in \calT$ denote the random stopping time at which a decision is taken; here $\calT$ designates the set of all such stopping times w.r.t. the filtration $\ppp{\calF_{\ell}}_{\ell\geq1}$, and a sequence $\ppp{\delta_{\ell}}_{\ell\geq1}$ of tests/decision rules, where $\delta_{\ell}$ is an $\calF_{\ell}$-measure function taking values in $[M]$. Let $\calD$ denote the set of all such functions. With the above definitions in mind, a decision rule is defined as,
\begin{align}
    \delta_{\s{T}} \triangleq \sum_{\ell = 0}^{\infty} \delta_{\ell} \mathds{1}_{\{ \s{T} = \ell \}},    
\end{align}
where the indicator function $\mathds{1}_{\{ \s{T} = \ell \}}$ is unity when $\s{T} = \ell$, and zero otherwise. A \emph{sequential decision rule} (SDR) is a pair $(\s{T},\delta)$, where $\s{T}$ declares the time to stop sampling, and once $\s{T}$ is given, $\delta_{\s{T}}$ takes one of the values in $[M]$, declaring which hypotheses to accept. Finally, for the node-based representation, the above definitions remain the same but with $\{\s{Z}_\ell\}_{\ell\geq1}$ and $\calF_\ell$ replaced by $\{\bar{\s{Z}}_\ell\}_{\ell\geq1}$ and $\bar{\calF}_{\ell} \triangleq \{\bar{\s{Z}}_1,\ldots, \bar{\s{Z}}_{\ell}\}$, respectively.

\begin{comment}
\begin{tcolorbox}[colback=blue!5!white, colframe=blue!75!black, title=Decision Rule Definition]
Given filtration $\calF_l$, decide which hypothesis $\calI$ is most probable, at the earliest time $\s{T}$ in which the posterior probability crosses the significance level $\frac{1}{1+a_k}$, for a given set of positive constants $\{a_k\}_k$,
\begin{align*}
    \hat{\calI} &= \argmax_{k} \P(\calF_T | \calH_k)\\
    \s{T} &\triangleq \min\{\ell\in\mathbb{N}:\s{s.t.} \P(\calF_T\vert\calH_\ell) \geq A_\ell\}.
\end{align*}
\end{tcolorbox}
\end{comment}

\paragraph{Approximate Bayesian optimality.} To capture the inherent tradeoff between the decision accuracy and the potential damage of spreading misinformation, we formulate the problem under a Bayesian framework. Specifically, for a given SDR $(\s{T},\delta)$, we define its total risk by,
\begin{align}
    \s{Risk}(\s{T},\delta) \triangleq \P_{\pi}(\delta_{\s{T}} \neq \calI) + \sum_{j=0}^{M-1} c_j\cdot\E\pp{\s{T}\mathds{1}_{\calH_\ell}},\label{eqn:riskTest}
\end{align}
where $c_j \in \mathbb{R}_+$, for $j\in[M]$. Here, the first term at the right-hand-side of \eqref{eqn:riskTest} is the average probability of error due to misdetection, explicitly given by,
\begin{align}
    \P_{\pi}(\delta_{\s{T}} \neq \calI) = \sum_{j=0}^{M-1}\pi_j\P_j\p{\delta_{\s{T}} \neq\ell}.
\end{align}
The second term at the right-hand-side of \eqref{eqn:riskTest} is the \emph{propagation cost} due to spreading misinformation, where $c_j$ is the cost of spreading the $j$th misinformation. Crucially, note that we do not penalize true information, following the underlying assumption that spreading news does not occur any cost. This makes the cost asymmetric and depends on the hypothesis, which is in contrast to classical theory of sequential testing problems where this cost is simply the average stopping time, i.e., $c\cdot\bE[\s{T}]$, for some $c>0$. A main goal in the theory of sequential testing is to find the SDR that minimizes the risk. In our case, this is formulated as the Bayesian optimization problem,
\begin{align} 
    \inf\limits_{\mathsf{T} \in \calT,\delta\in\calD} \s{Risk}(\s{T},\delta).\label{eq:OptimalStoppingProblem}
\end{align}
Finding the optimal SDR in the multiple hypothesis case is challenging even for a symmetric propagation cost and an i.i.d. probabilistic model.  
%Since MSPRT \cite{340472} (lemma 3.3) did not assume the data to be i.i.d in their proof of approximate Bayesian optimality, their conclusions still apply.
%MSPRT \cite{340472} proves in lemma 3.3 the MSPRT algorithm provides an approximate solution to 
In the following sections, we propose ``approximately optimal" SDRs, and prove several statistical guarantees on their performance. We also devise data driven algorithms for learning the edge types and transition probabilities from training data, as these are typically unknown in practical scenarios.

%\paragraph{Information Cascade.} The information cascade (AKA retweet trace) is a subgraph of $\calG$ which consists of all users who shared a certain post. Each information cascade has a single information source $s \in \calV$. Let $\calP$ denote the set of all possible paths starting at the source $s$. We assume each path $P_s = (s, v_1,\ldots,v_{|P|}) \in \calP$ is associated with a first-order homogeneous Markov process. In other words, the set of edge features across the path, $w^P = \{w_{e_i}\}_{i=0}^{|P|-1}$ are sampled from a Markov random process. Consequently, the set of edge types across the path are sampled from a homogeneous Markov Chain (HMC). We denote $\alpha_j(z, z')$ the transition matrix from edge z to z' under hypothesis $\calH_j$. The probability of seeing an edge type z emanating from $s$ is denoted $\eta_j(z)$.

\section{Main Results} \label{section:main}

In this section, we present our main results. In the first part of this section, we start with the edge-based representation. Specifically, assuming that the Markov kernels and edge types are known, we propose a model based sequential detection algorithm for the optimization problem in \eqref{eq:OptimalStoppingProblem}, and prove several theoretical guarantees on its performance. In the second part of this section, we move forward to the node-based representation. Here, we propose a novel GNN SDR architecture, and prove theoretical guarantees on its performance. As will be explained later on, this architecture implicitly learns the Markov dependencies between consecutive nodes, and proves more robust to estimation errors. Finally, we construct offline routines for training the edge classifier, and estimating the edge Markov kernels, under each hypothesis, under the edge-based representation model.

%As we will explain later on, the classical MSPRT suffers from a relatively high computational complexity and is, perhaps more importantly, not robust to imprecision in the estimations of the transition probabilities and in edge types. To that end, we propose a novel GNN architecture, which implicitly learns the Markov dependencies between consecutive nodes and as so more robust to estimation errors, and exhibit better computational complexity. 

\subsection{Edge-based representation}\label{subsection:MSPRT}

%\subsubsection{Sequential test and performance guarantees}\label{subsection:MSPRT}

%%Since MSPRT \cite{340472} (lemma 3.3) did not assume the data to be i.i.d in their proof of approximate Bayesian optimality, their conclusions still apply. %MSPRT \cite{340472} proves in lemma 3.3 the MSPRT algorithm provides an approximate solution to 
Recall the edge-based representation described in the previous section. We next propose a SDR for the testing problem in \eqref{eqn:learning}, assuming that the Markov kernels and edge types are known. Later on, in Subsection~\ref{subsection:dataAlgo}, we propose a data-based algorithm for learning these parameters. Recall that in the edge-based representation, we transform the information trace sequence $\{(\mathbf{x}_{u_\ell},\mathbf{x}_{v_\ell})\}_{\ell\geq1}$ into an edge sequence $\{\s{Z}_\ell\}_{\ell\geq1}$, and in this subsection we assume that $|\calZ|<\infty$, namely, $\s{Z}_\ell$ is discrete for any $\ell\geq1$; this implies that $\{\s{Z}_\ell\}_{\ell\geq1}$ forms a union of Markov chains. For our theoretical results only, we further assume that these Markov chains are irreducible.

Our SDR for solving \eqref{eq:OptimalStoppingProblem} is the well-known MSPRT \cite{340472}. To describe this test, we introduce some notations. For $m\in[M]$ and $\ell\in\mathbb{N}$, let 
\begin{align}
    \Pi_{\ell}^{(m)}&\triangleq \P{\calH_m \vert \calF_{\ell}},
\end{align}
denote the posterior probability of the $m$th hypothesis. Bayes theorem gives,
\begin{equation}\label{eq:msprt}
    \Pi_\ell^{(m)} = \frac{\pi_m f_m(\s{Z}_1^\ell)}{\sum_{j=0}^{M-1} \pi_{j} f_j(\s{Z}_1^\ell)},
\end{equation}
where for $m\in[M]$ and $\ell\in\mathbb{N}$ we define $f_m(\s{Z}_1^\ell) \triangleq \P(\s{Z}_1^\ell \vert \calH_m)$; see \eqref{eqn:jointprobLAW}--\eqref{eqn:jointprobLAW2} for an explicit expression for $f_m(\cdot)$. We are now in a position to state how the MSPRT works. Specifically, the stopping time $\s{T}_{\s{MSPRT}}$, and the final decision $\delta_{\s{MSPRT}}$, can be described as follows,
\begin{equation}
\boxed{\begin{aligned}\label{eqn:decproblem}
    \s{T}_{\s{MSPRT}} &= \inf\ppp{\ell\in\mathbb{N}:\;\Pi_\ell^{(m)} \geq \frac{1}{1+a_m},\;\s{for}\;\s{some}\;m},\\
    \delta_{\s{MSPRT}}& = \calH_{m^\star},\;\s{where}\;m^\star=\arg\max_m \Pi_{\s{T}_{\s{MSPRT}}}^{(m)}.
\end{aligned}}
\end{equation}
In the above, the (hyper-) parameters $\{a_m\}_{m\in[M]}$ control the significance threshold; to ensure that only a single hypothesis is chosen we set $0\leq a_m < 1$, for all $m\in[M]$. A pseudo-code of the above MSPRT procedure is given in Algorithm~\ref{alg:online}. We would like to emphasize here that in the second step of Algorithm~\ref{alg:online}, we implicitly use some portion of the information trace $\{(\mathbf{x}_{u_\ell},\mathbf{x}_{v_\ell})\}_{\ell\geq1}$ as a training in order to learn the edge representation, transition probabilities, and initial probabilities, using Algorithm~\ref{alg:offline}, which we explain in detail in Subsection~\ref{subsection:dataAlgo}. For the rest of this paper, we denote by $\hat{\calH}_{\s{MSPRT}}$, $\delta_{\s{MSPRT}}$, and $\s{T}_{\s{MSPRT}}$, the decision, the rule, and the stopping time, associated with the MSPRT algorithm, respectively.

\begin{comment}
    The likelihood is given by:
\begin{equation}\label{eq:likelihood}
    f_j(Z_1^l) = \prod_{i=1}^l A_j(Z_i | F_{i-1})
\end{equation}
The conditional probabilities $A_j$ are derived similarly to \cite{orenloberman2023online}:
\begin{equation}\label{eq:A_j}
    A_j(Z_l | F_{l-1}) \triangleq \Pr(Z_l | H_j, F_{l-1}) = 
    \mathbb{E}_{\mu_j} \left[\sum_{Z_l(P)} \prod_{t=I_i(P)+1}^i \alpha(\s{Z}_t | \s{Z}_{t-1})\right]
\end{equation}
$P \in \calP_l$, where $\calP_l$ denotes all possible paths starting at $s$ and ending with the current observation $Z_l$. $Z_l(P) \triangleq Z_{I_i(P)),\ldots,Z_{i-1}}$ denotes the unobserved edges along path P, from the last observed edge in P up to the current observation. $Z_i$ is therefore the current observation, and $Z_{I_i(P)}$ the previous observation along P.
The probability of path $P \in \calP_l$ is $\mu_j(P_s)$, where $\calP_l$ denotes all possible paths starting at $s$ and ending with the current observation $Z_l$. We use the same equation as in \cite{orenloberman2023online}.
\end{comment}

\begin{algorithm}[t!]
\caption{MSPRT}\label{alg:online}
\textbf{Input} {$\{(\mathbf{x}_{u_\ell},\mathbf{x}_{v_\ell})\}_{\ell\geq1}$}.\\
\textbf{Obtain} {$\{\s{Z}_\ell\}_{\ell\geq1}$, $\{\alpha_j(\cdot|\cdot)\}_{j\in[M]}$, and $\{{\eta}_j(\cdot)\}_{j\in[M]}$ using Algorithm~\ref{alg:offline}.}
\begin{algorithmic}
    \State \textbf{Initialize:} $0 < a_j < 1, \Pi_0^{(m)} \leftarrow \pi_m, \calF_0 \leftarrow \emptyset$.
    \State \textbf{While}  $\Pi_\ell^{(m)} < \frac{1}{1+a_m} \quad \forall m\in[M]$ \textbf{do:}
        \State\hspace{\algorithmicindent} $\ell \leftarrow \ell+1, \calF_\ell \leftarrow (\calF_{\ell-1}, \s{Z}_\ell)$
%        \State\hspace{\algorithmicindent} Find $A_m(\s{Z}_\ell\vert\calF_{\ell-1})\; \forall m$ using \eqref{eq:A_j}. 
        \State\hspace{\algorithmicindent} Find $\Pi_{\ell}^{(m)}\; \forall m$ using \eqref{eq:msprt}.
\end{algorithmic}
\textbf{Return} Hypothesis $m$ for which $\Pi_{\ell}^{(m)} \geq \frac{1}{1+a_m}$.
\end{algorithm}

Many statistical guarantees for the MSPRT algorithm are well-known in the literature, under the i.i.d. model (see, e.g., \cite{340472}). For example, it is known that the MSPRT reaches a posterior probability estimate within the specified significance level, at a bounded time. Furthermore, asymptotic guarantees regarding the stopping time, the probability of error, and the approximation of Bayesian optimality, are known as well. We prove several similar guarantees, but under the more general Markovian model presented in the previous section. For simplicity, in order to prove the following theoretical results, we assume that any pair of matrix transition probabilities are sufficiently different. This is formulated in terms of the conditional Hellinger distance. Specifically, recall that for probability measures $\mathbb{P}$ and $\mathbb{Q}$, $\calH^2(\mathbb{P}||\mathbb{Q}) = \frac{1}{2}\cdot\bE_{\mathbb{Q}}(1-\sqrt{\mathrm{d}\mathbb{P}/\mathrm{d}\mathbb{Q}})^2$, denotes the Hellinger distance between $\mathbb{P}$ and $\mathbb{Q}$. In our case, for any $z\in\calZ$ and $k\neq j\in[M]$, define,
\begin{align}
    S_{k,j}(z)&\triangleq1-\calH^2(\alpha_k(\cdot \vert z), \alpha_j(\cdot\vert z))\\
    &= \sum_{z'\in\calZ}\sqrt{\alpha_k(z' \vert z)\alpha_j(z' \vert z)}.\label{eqn:SkjDef}
\end{align}
Then, we assume that $\max_{z\in\calZ}\max_{k\neq j\in[M]}S_{k,j}(z)<1$. To wit, for any $k\neq j$ and any $z\in\calZ$, the transition probability distributions $\alpha_k(\cdot\vert z)$ and $\alpha_j(\cdot\vert z)$ are not the same, that is, there exist $z'\in\calZ$ such that $\alpha_k(z'\vert z)\neq \alpha_j(z'\vert z)$. 

We start with the following result, which shows that the probability that $\s{T}_{\s{MSPRT}}$ exceeds $t$ decreases exponentially with $t$.
\begin{theorem}[Exponentially bounded stopping time]\label{th:bounded_stop_time}
Fix $k\in[M]$, and assume that
\begin{align}
\min_{j\neq k\in[M]}\max_{z\in\calZ}S_{k,j}(z)<1.
\end{align}
Then, for any $t\in\mathbb{R}_+$, we have,
\begin{align}
    \P\pp{\left.\s{T}_{\s{MSPRT}}>t\right|\calH_k}\leq \s{C}_1\cdot\exp(-\s{C}_2\cdot t),
\end{align}
for some $\s{C}_1,\s{C}_2\in\mathbb{R}_+$.
%Conditioned on each of the hypotheses $\calH_0, H_1,\ldots,\calH_{M-1}$, the stopping time $\s{T}$ is exponentially bounded.
\end{theorem}
Theorem~\ref{th:bounded_stop_time} implies that with probability at least $1-\delta$, we have $\s{T}_{\s{MSPRT}}\leq \frac{1}{\s{C}_2}\log\frac{\s{C}_1}{\delta}$. In particular, $\s{T}_{\s{MSPRT}}$ is finite with probability one. To prove this result, we follow a similar approach as in \cite{340472}. To wit, we first show that
\begin{align}
    \P\pp{\left.\s{T}_{\s{MSPRT}}>t\right|\calH_k}\leq 1 - \min_{j\neq k}\calH^2 (f_j(\s{Z}_1^t), f_k(\s{Z}_1^t)).\label{eqn:1minusHel}
\end{align}
As it turns out, the term at the right-hand-side of \eqref{eqn:1minusHel} decreases exponentially with $t$, as $t\to\infty$, which implies that the run-time of Algorithm ~\ref{alg:online} is bounded. Next, we analyze the probability of error. To that end, for $j\neq k\in[M]$, define $\calE^{\s{MSPRT}}_{j,k}\triangleq\P_{\calH_j}(\hat{\calH}_{\s{MSPRT}} =\calH_k)$, $\calE_{k}^{\s{MSPRT}}\triangleq\P(\hat{\calH}_{\s{MSPRT}} =\calH_k)$, and $\calE^{\s{MSPRT}}$ as the total probability of incorrect decision. We mention here that $\{\calE^{\s{MSPRT}}_k\}_{k\in[M]}$ are also known as the \emph{frequentist} error probabilities, and in general, they are different from the standard conditional error probabilities.
\begin{theorem}[Error guarantees]\label{th:error_guarantee}
For all $k\in[M]$, we have,
\begin{align}
    \calE_{k}^{\s{MSPRT}}&=\sum_{j:j \neq k}\pi_j\calE^{\s{MSPRT}}_{j,k}
  \le \pi_k a_k,\\
  \calE^{\s{MSPRT}} &\triangleq \sum_{k\in[M]} \calE^{\s{MSPRT}}_k
  \leq \sum_{k\in[M]} \pi_k a_k.
\end{align}
If, in addition, $a_\ell = \nu$, for all $\ell\in[M]$, and for some $\nu\in[0,1]$, then,
\begin{align}
    \calE^{\s{MSPRT}} \leq \frac{\nu}{\nu+1}.
\end{align}
\end{theorem}
The bounds above are derived using techniques similar to the ones used by Wald for the SPRT \cite{10.1214/aoms/1177730197}, and therefore, its proof is relegated. Next, we investigate the asymptotic behaviour of the MSPRT stopping time. Specifically, we find the asymptotic behavior of $\s{T}_{\s{MSPRT}}$, in the regime where the error probabilities are ``small", and accordingly, the stopping time is ``large". In light of Theorem~\ref{th:error_guarantee}, this regime corresponds to the case where $\norm{a}_\infty\triangleq\max_{\ell\in[M]}|a_\ell|\to0$. To present our main finding, we need a few definitions. The Kullback-Leibler (KL) divergence between two probability measures $\mathbb{P}$ and $\mathbb{Q}$ is defined as let $d_{\s{KL}}(\mathbb{P}||\mathbb{Q})\triangleq \bE_{\mathbb{P}}\log\frac{\mathrm{d}\mathbb{P}}{\mathrm{d}\mathbb{Q}}$. Accordingly, in our case, for any $z\in\calZ$, we define,
\begin{align}
    d_{\s{KL}}(\alpha_k(\cdot\vert z)||\alpha_{j}(\cdot\vert z)) \triangleq \sum_{z'\in\calZ}\alpha_k(z'\vert z)\log\frac{\alpha_k(z'\vert z)}{\alpha_j(z'\vert z)}.
\end{align}
%Let $X$ and $Y$ be two discrete random variables over $\mathcal{X}$ and $\mathcal{Y}$, respectively. Let $\mathbb{P}_{XY}$ and $\mathbb{Q}_{XY}$ be two joint probability distributions. Then, we define $d_{\s{KL}}(\mathbb{P}_{XY}||\mathbb{Q}_{XY}) = \bE_{\mathbb{P}_{XY}}\log(\mathrm{d}\mathbb{P}_{XY}/\mathrm{d}\mathbb{Q}_{XY})$ as the Kullback-Leibler (KL) divergence between $\mathbb{P}_{XY}$ and $\mathbb{Q}_{XY}$. Also, the conditional KL divergence is defined as,
%\begin{align}
%    d_{\s{KL}}\p{P_{Y|X}||Q_{Y|X}\vert P_X} &= \bE_{X\sim P_X}\pp{d_{\s{KL}}(P_{Y|X}||Q_{Y|X})}\\
%    & =\sum_{x\in\mathcal{X}}P_X(x)d_{\s{KL}}(P_{Y|X}(\cdot|x)||Q_{Y|X}(\cdot|x))\\
%    & = \sum_{x\in\mathcal{X}}P_X(x)\sum_{y\in\mathcal{Y}}P_{Y|X}(y|x)\log\frac{P_{Y|X}(y|x)}{Q_{Y|X}(y|x)}.
%\end{align}
Also, we denote by $\pi^{\s{stat}}_k$ the stationary distribution of the $k$th irreducible Markov chain with transition probabilities $\alpha_k(\cdot\vert\cdot)$. Then, in accordance to the above notations, we define the stationary conditional KL divergence as,
\begin{align}
    d_{\s{KL}}(\alpha_k||\alpha_j\vert\pi_k^{\s{stat}}) = \sum_{z'\in\mathcal{Z}}\pi^{\s{stat}}_k(z')\cdot d_{\s{KL}}(\alpha_k(\cdot\vert z')||\alpha_j(\cdot\vert z')).
\end{align}
Furthermore, the $\chi^2$-divergence between two probability measures $\mathbb{P}$ and $\mathbb{Q}$ is defined as $\chi^2(\mathbb{P},\mathbb{Q})\triangleq\E_{\mathbb{P}}\frac{\mathrm{d}\mathbb{P}}{\mathrm{d}\mathbb{Q}}-1$. In our case, we define for any $z\in\calZ$,
\begin{align}
    \chi^2(\alpha_k(\cdot\vert z),\alpha_{j}(\cdot\vert z)) \triangleq \sum_{z'\in\calZ}\frac{\alpha^2_k(z'\vert z)}{\alpha_j(z'\vert z)}-1.
\end{align}
We are now in a position to state our main result.

%\begin{align}\label{eq:conditional_rel_entropy}
%d_{\s{KL}}\left(\alpha_k(\s{Z}_{i}^P\vert\s{Z}_{i-1}^P)||\alpha_j(\s{Z}_{i}^P\vert\s{Z}_{i-1}^P)\right) \triangleq \sum_{z \in \calZ} \pi_s(\s{Z}^P=z\vert \calH_k)d_{\s{KL}}\left(\alpha_k(\s{Z}_{i}^P\vert\s{Z}_{i-1}^P=z)||\alpha_j(\s{Z}_{i}^P\vert\s{Z}_{i-1}^P=z)\right)
%\end{align}
%$\pi_s(\s{Z}_{u_i} \vert \calH_k)$ here denotes the stationary distribution of $\alpha_k$.
\begin{comment}
Since our data is a random process, we are interested in the KL divergence \textit{rate} between the likelihoods $f_k$ and $f_j$, which is defined as:
\begin{align}
    \bar{d}_{\s{KL}}(f_k||f_j) \triangleq\lim_{n\to\infty}\frac{d_{\s{KL}}(f_k(\s{Z}_1^n),f_j(\s{Z}_1^n))}{n}.
\end{align}
In our case, this can be simplified to,
\begin{align}
\bar{d}_{\s{KL}}(f_k||f_j) = \lim_{n \to \infty} \sum_{\s{Z}_{1}^{n} \in \calZ^n} f_k(\s{Z}_{1}^{n}) \frac{1}{n} \sum_{\s{P}\in\calP_\ell}\sum_{i\in[|\s{P}|]} \log\left( \frac{\alpha_k(\s{Z}_{i}^P\vert\s{Z}_{i-1}^P)}{\alpha_j(\s{Z}_{i}^P\vert\s{Z}_{i-1}^P)} \right).
\end{align}
\end{comment}
%In Lemma \ref{lem:AEP} we show our observations $\s{Z}_1^\ell$ posses the following \textit{Asymptotic Equipartition Property}:
%\begin{align}
%    \frac{1}{\ell}\log{\frac{f_k(\s{Z}_1^\ell)}{f_j(\s{Z}_1^\ell)}}
%    \xrightarrow[\ell\to\infty]{}    d_{\s{KL}}\left(\alpha_k(\s{Z}_{i}^P\vert\s{Z}_{i-1}^P)||\alpha_j(\s{Z}_{i}^P\vert\s{Z}_{i-1}^P)\right)
%\end{align}
%At this point we defined All the necessities to introduce the Asymptotic stopping time theorem.
\begin{theorem}[Asymptotic stopping time]\label{th:asymptotic_time}
Fix $k\in[M]$, and assume that
\begin{align}
\min_{j\neq k\in[M]}\max_{z\in\calZ} \chi^2(\alpha_k(\cdot\vert z),\alpha_{j}(\cdot\vert z)) < \infty,
\end{align}
and
\begin{align}
\min_{j\neq k\in[M]}\max_{z\in\calZ}S_{k,j}(z)<1.
\end{align}
%\begin{align}
%    0<\min_{j \neq k} d_{\s{KL}}\left(\alpha_k||\alpha_j\vert\pi_k^{\s{stat}}\right)<\infty.
%\end{align}
Then,
\begin{align}    
    \lim_{\norm{a}_\infty\to0} \frac{\s{T}_{\s{MSPRT}}}{- \log a_k} = \frac{1}{\min_{j \neq k} d_{\s{KL}}\left(\alpha_k||\alpha_j\vert\pi_k^{\s{stat}}\right)},
\end{align}
$f_k$-almost surely. Furthermore,
\begin{align}
    \lim_{\norm{a}_\infty\to0} \frac{\E\left[\s{T}_{\s{MSPRT}}\right]}{- \log a_k} = \frac{1}{\min_{j \neq k} d_{\s{KL}}\left(\alpha_k||\alpha_j\vert\pi_k^{\s{stat}}\right)}.
\end{align}
\end{theorem}
Theorem~\ref{th:asymptotic_time} connects between the stopping time $\s{T}_{\s{MSPRT}}$, the required significance levels $\{a_k\}_k$, and the KL-divergence between the closest hypothesis to the actual decision. Intuitively, as the required significance level increases, the algorithm requires more iterations to reach that level. Furthermore, as the hypotheses gets ``closer", namely, as the KL-divergence gets smaller, the algorithm requires more steps to reach a decision, as expected.

\subsection{Node-based representation}\label{subsection:GNN}
%\subsection{GNN-based SDR and performance guarantees}\label{subsection:GNN}

While Algorithm~\ref{alg:online} enjoys desirable statistical guarantees, it suffers from a relatively high sample complexity. Indeed, as the number of hypotheses increase, Algorithm~\ref{alg:online} runs quickly into the \textit{curse of dimensionality}; since there are $\vert \calZ \vert = M(M-1)$ possible states, the transition matrix has $\mathcal{O}(M^4)$ entries, and since Algorithm~\ref{alg:online} requires the evaluation of each one of these entries, its computational complexity is governed by these number of states, which can be huge in practice. Even more importantly, as it turns out from our experimental study, it seems as though that Algorithm~\ref{alg:online} is quite sensitive to errors resulted during the estimation of the transition probabilities and edge types. Finally, in practice, the underlying observational processes may not truly be homogeneous/time-invariant, while Algorithm~\ref{alg:online} relies strongly on this assumption. As a remedy to these issues, we propose a GNN-based architecture, which implicitly learns the Markovian dependencies between consecutive nodes on the fly. For this algorithm we were able to prove that Theorems~\ref{th:bounded_stop_time}--\ref{th:asymptotic_time} remain true (and as so exhibit favorable statistical guarantees), but at the same time is more robust to estimation and model errors, and is faster than Algorithm~\ref{alg:online}. Next, we describe the algorithm and present its statistical guarantees.

\begin{figure}[t!]
    \centering
    \usetikzlibrary{positioning, calc, decorations.pathreplacing}
%\begin{document}
\begin{tikzpicture}[scale=0.7]

  \node[fill=orange!50] (dense) {\texttt{Dense}};
  \node[blue!50!black, right=0.5cm of dense] (relu) {\texttt{ReLU}};
  \node[fill=teal!50, right=1cm of relu] (ginconv) {\texttt{GINConv}};
  \node[blue!50!black, right=0.5cm of ginconv] (softmax1) {$\log(\texttt{softmax})$};
  \node[right=of softmax1, font=\Large, label={below:add}, inner sep=0] (add) {$\oplus$};
  \node[blue!50!black, right=0.5cm of add] (softmax2) {\texttt{softmax}};
  \node [right=0.5cm of softmax2] (output) {$\left(\begin{array}{c}
  \Phi_0\\
  \Phi_1\\
  \Phi_2\\
  \vdots \\
  \Phi_{\mathsf{M}-1}\\
  \end{array}\right)$
};

  %\node[right=of softmax1, font=\Large, label={below:add}, inner sep=0, pin={60:$\mathcal F(\vec x) + \vec x$}] (add) {$\oplus$};
  %\node[blue!50!black, right=of add, label={below:activation}] (softmax1) {$a(\vec x)$};

  \draw[->] (dense) -- (relu);
  \draw[->] (relu) -- (ginconv);
  \draw[<-] (dense) -- ++(-2,0) node[below, pos=0.8](begin){};%{\input{graph.tex}};
  \draw[->] (ginconv) -- (softmax1) node[above, pos=0.8] {};
  \draw[->] (softmax1) -- (add) node[above, pos=0.8] {};
  \draw[->] (add) -- (softmax2) node[above, pos=0.8] {};
  \draw[->] (softmax2) -- (output) node[above, pos=0.8] {};

  \draw[decorate, decoration={brace, amplitude=1ex, raise=1cm}] (dense.west) -- node[midway, above=1.2cm] {Edge classifier $g(\cdot)$} (ginconv.east);

\coordinate[, pin={60:$x_i$}] (begin_pos) at ($(begin.west)!0.5!(dense.west)$);
\node[below=1cm of begin_pos] (image0) {%\documentclass[tikz]{standalone}
%\usetikzlibrary{positioning, calc, decorations.pathreplacing}
%\begin{document}
\begin{tikzpicture}[
        every node/.style={circle, draw, thin, minimum size=2mm, scale=0.3},
        grow=right,
        level distance=0.4cm,
        edge from parent/.style={-, draw},
        edge/.style = {-, draw},
        level 1/.style={sibling distance=4mm},
        level 2/.style={sibling distance=2mm},
        level 3/.style={sibling distance=1mm}
    ]
    \node (1) {1}
        child {node (2) {2}
            child {node (5) {5}}
            child {node (6) {6}}
        }
        child {node (3) {3}
            child {node (7) {7}}
            child {node (4) {4}}
        }
        child {node (8) {8}
            child {node (10) {10}}
            child {node (9) {9}}
        }
;
\end{tikzpicture}

%\end{document}
};
  \draw[-] (begin_pos) -- (image0);
\coordinate[pin={60:$x'_i$}] (graph1_pos) at ($(relu)!0.5!(ginconv)$);
  \node[below=1cm of graph1_pos] (image1) {%\documentclass[tikz]{standalone}
%\usetikzlibrary{positioning, calc, decorations.pathreplacing}
%\begin{document}
\begin{tikzpicture}[
        every node/.style={circle, draw, thin, minimum size=2mm, scale=0.3},
        grow=right,
        level distance=0.4cm,
        edge from parent/.style={-, draw},
        edge/.style = {-, draw},
        level 1/.style={sibling distance=4mm},
        level 2/.style={sibling distance=2mm},
        level 3/.style={sibling distance=1mm}
    ]
    \node (1) {1}
        child {node (2) {2}
            child {node (5) {5}}
            child {node (6) {6}}
        }
        child {node (3) {3}
            child {node (7) {7}}
            child {node (4) {4}}
        }
        child {node (8) {8}
            child {node (10) {10}}
            child {node (9) {9}}
        }
;
\end{tikzpicture}

%\end{document}
};
  \draw[-] (graph1_pos) -- (image1);
\coordinate[, pin={60:$\log\varphi_i$}] (graph2_pos) at ($(softmax1.east)!0.5!(add)$);
  \node[below=1cm of graph2_pos] (image2) {%\documentclass[tikz]{standalone}
%\usetikzlibrary{positioning, calc, decorations.pathreplacing}
%\begin{document}
\begin{tikzpicture}[
        every node/.style={circle, draw, thin, minimum size=2mm, scale=0.3},
        grow=right,
        level distance=0.4cm,
        edge from parent/.style={-, draw},
        edge/.style = {-, draw},
        level 1/.style={sibling distance=4mm},
        level 2/.style={sibling distance=2mm},
        level 3/.style={sibling distance=1mm}
    ]
    \node (1) {1}
        child {node (2) {2}
            child {node (5) {5}}
            child {node (6) {6}}
        }
        child {node (3) {3}
            child {node (7) {7}}
            child {node (4) {4}}
        }
        child {node (8) {8}
            child {node (10) {10}}
            child {node (9) {9}}
        }
;
\end{tikzpicture}

%\end{document}
};
  \draw[-] (graph2_pos) -- (image2);
\coordinate[, pin={60:$\bar{\mathsf{Z}}_i$}] (ginconv_pos) at ($(ginconv.east)!0.5!(softmax1.west)$);

\node[below=0.01cm of image0] {\small $\mathbb{R}^{\vert\mathcal{V}\vert\times\vert x_i\vert}$};
\node[below=0.01cm of image1] {\small $\mathbb{R}^{\vert\mathcal{V}\vert\times h}$};
\node[below=0.01cm of image2] {\small $\mathbb{R}^{\vert\mathcal{V}\vert\times \mathsf{M}}$};

\end{tikzpicture}
%\end{document}
    \caption{The msprtGNN architecture. \texttt{Dense} operates locally on node features. \texttt{GINConv} embeds each node with its ancestor. The add pooling layer aggregates the embeddings of all nodes to a graph level vector.}
    \label{fig:msprtGNN}
\end{figure}
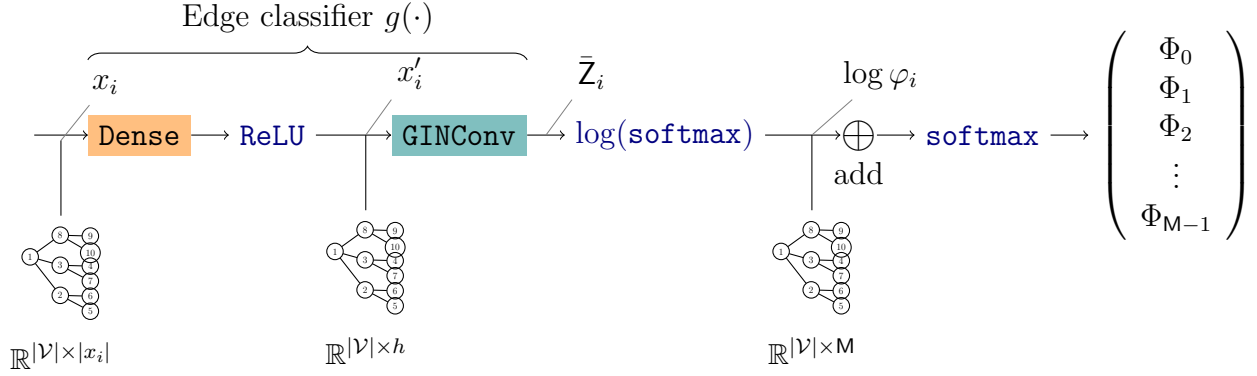

\paragraph{Architecture.} Below, we let $\s{Dense}$ denote a linear layer which operates locally on each node feature, and $\s{GINConv}$ denote the graph isomorphism network in \cite{DBLP:journals/corr/abs-1810-00826}. We let $h\in\mathbb{N}$ be the dimension of these hidden layers. Our architecture comprises of a total of four layers, with two of them depending on several trainable parameters. Specifically, the input is a graph, with node features, described by an $n \times d$ user-feature matrix $\mathbf{X} \triangleq [\mathbf{x}_1^T,\ldots,\mathbf{x}_n^T]^T$, and a list of edges in the (sub-)graph. In the first layer, we have a concatenation of a $\s{Dense}$ linear layer, operating locally on each feature vector, with $h\cdot(d+1)$ trainable parameters, and a rectified-linear unit (ReLU) activation function. Mathematically, for each $i\in[n]$, the output of the first layer is,
\begin{align}
    \mathbf{x}_{i}' = \s{ReLU}(\s{Dense}(\mathbf{x}_{i})).
\end{align}
Then, the second layer is the convolution layer $\s{GINConv}$ proposed in \cite{DBLP:journals/corr/abs-1810-00826}, whose output is the $M$-dimensional vector,
\begin{align}
    \bar{\s{Z}}_{i} = \Theta\left((1+\epsilon)\mathbf{x}_{i}' + \sum_{u \in \mathcal{N}_{i}} \mathbf{x}_{u}' \right),
\end{align}
for all $i\in[n]$, where $\Theta$ is a single linear layer, and $\epsilon$ is a trainable parameter. This layer embeds the features of each node with those of its parents. Thus, the $\s{GINConv}$ layer have $h \cdot M +M + 1 $ trainable parameters. Then, the output of the second layer serves as an input to a per-node $\s{log}$-$\s{softmax}$ layer, whose output is,
\begin{align}
\left[\varphi_i\right]_k=\log\left(\frac{\exp\p{\left[\bar{\s{Z}}_i\right]_{k}}}{\sum_{j=0}^M \exp\p{\left[\bar{\s{Z}}_i\right]_{j}}}\right),
\end{align}
for all $i \in [\ell]$, and $k \in \left[M\right]$. Finally, we perform graph-level aggregation via an addition-pooling layer, and then apply softmax again. The output $\Phi$ is an $M$-array vector, whose elements are given by,
\begin{align}
    \Phi_m = \frac{\prod_{i \in [\ell]} \left[\varphi_i\right]_{m}}{\sum_{j=0}^{M-1}\prod_{i \in [\ell]} \left[\varphi_i\right]_{j}},\label{eqn:phidef}
\end{align}
for all $m \in \left[M\right]$. Fig.~\ref{fig:msprtGNN} provides an illustration for the architecture described above; we will refer to this architecture as ``msprtGNN".

\paragraph{Inference procedure.} Our inference procedure is in fact the same as the MSPRT in \eqref{eqn:decproblem}, but instead of using the posterior probability associated with each hypothesis, we use the network outputs $\Phi_m$, for $m\in[M]$. Specifically, the stopping time $\s{T}_{\s{GNN}}$, and the final decision $\delta_{\s{GNN}}$, can be described as follows,
\begin{equation}
\boxed{\begin{aligned}\label{eqn:decproblem2}
    \s{T}_{\s{GNN}} &= \inf\ppp{\ell\in\mathbb{N}:\;\Phi_\ell^{(m)} \geq \frac{1}{1+a_m},\;\s{for}\;\s{some}\;m},\\
    \delta_{\s{GNN}}& = \calH_{m^\star},\;\s{where}\;m^\star=\arg\max_m \Phi_{\s{T}_{\s{GNN}}}^{(m)}.
\end{aligned}}
\end{equation}

\begin{algorithm}[t!]
\caption{msprtGNN}\label{alg:gnn}
\textbf{Input} {$\{(\mathbf{x}_{u_\ell},\mathbf{x}_{v_\ell})\}_{\ell\geq1}$}.\begin{algorithmic}
    \State \textbf{Initialize:} $0 < a_j < 1, \Phi_0^{(m)} \leftarrow \pi_m, \calF_0 \leftarrow \emptyset$.
    \State \textbf{While}  $\Phi_\ell^{(m)} < \frac{1}{1+a_m} \quad \forall m\in[M]$ \textbf{do:}
        \State\hspace{\algorithmicindent} $\ell \leftarrow \ell+1, \calF_\ell \leftarrow {\calF_{\ell-1}, \s{x}_{v_\ell}}$
        \State\hspace{\algorithmicindent} Find $\forall m$:
            \State\hspace{\algorithmicindent}\hspace{\algorithmicindent} $x_{v_i}' \leftarrow \texttt{ReLU}(\texttt{Dense}(x_{v_i})) : \mathbb{R}^{|\calV| \times |x_v|} \to \mathbb{R}^{|\calV| \times h}$ 
            \State\hspace{\algorithmicindent}\hspace{\algorithmicindent} $\bar{\s{Z}}_{i} \leftarrow \texttt{GINConv}(x_{v_i}', x_{u_i}') : \mathbb{R}^{|\calV| \times h} \to \mathbb{R}^{|\calV| \times M}$
            \State\hspace{\algorithmicindent}\hspace{\algorithmicindent} $\varphi_i \leftarrow \texttt{softmax}(\bar{\s{Z}}_{v_i}) : \mathbb{R}^{|\calV| \times M} \to \mathbb{R}^{|\calV| \times  M}$
            \State\hspace{\algorithmicindent}\hspace{\algorithmicindent} $\Phi_m \leftarrow \frac{\prod_{i \in [\ell]} \varphi_{i}^{(m)}}{\sum_{j=0}^{M-1}\prod_{i \in [\ell]} \varphi_{i}^{(j)}} : \mathbb{R}^{|\calV| \times M} \to \mathbb{R}^{M} $
\end{algorithmic}
\textbf{Return} Hypothesis $m$ for which $\Phi_{\ell}^{(m)} \geq \frac{1}{1+a_m}$.
\end{algorithm}

A few important comments are in order. Note that in order to apply the MSPRT procedure in Algorithm~\ref{alg:online}, the Markov kernels and the edges types are needed, and in the following subsection, we explain how these can be deduced from data. In msprtGNN, however, these are learnt implicitly; it can be seen that the msprtGNN procedure in Algorithm~\ref{alg:gnn} is independent of these unknowns. Intuitively speaking, the first two layers in our architecture act as an edge classifier. In fact, this is one of the main reasons why the msprtGNN procedure turns out to be more robust to model and estimation errors.

Similarly to MSPRT, we prove several statistical guarantees on the performance of msprtGNN. As in the previous subsection, for the following theoretical results, we assume that any pair of Markov kernels are ``sufficiently far". Again, we formulae this in terms of the conditional Hellinger distance; for any $z\in\bar{\calZ}$ and $k\neq j\in[M]$, define,
\begin{align}
    \bar{S}_{k,j}(z)&\triangleq1-\calH^2(\bar\alpha_k(\cdot \vert z), \bar\alpha_j(\cdot\vert z))\\
    &= \int_{z'\in\bar\calZ}\sqrt{\bar\alpha_k(z' \vert z)\bar\alpha_j(z' \vert z)}\mathrm{d}z'.\label{eqn:SkjDef2}
\end{align}
Then, we assume that $\max_{z\in\bar\calZ}\max_{k\neq j\in[M]}\bar{S}_{k,j}(z)<1$. Furthermore, we define
\begin{align}
    \xi \triangleq \max_{j,k\in[M],\ell\geq 1, z_1^\ell\in\bar{\calZ}^\ell}  \left.\frac{\Phi_k(z_1^\ell)}{\Phi_j(z_1^\ell)}\right/\frac{f_k(z_1^\ell)}{f_j(z_1^t)},\label{eqn:xidef}
\end{align}
which plays a significant role in the following results. Intuitively, $\xi$ measures the similaritdiscrioinecny between the underlying likelihoods/probability posteriors and the outputs of the msprtGNN architecture. FIn fact, for our results to hold, we need to assume that $\xi$ is finite, and we claim that this is indeed reasonable. For exampleSpecifically, when neural network classifiers are trained using the cross-entropy loss, it is well-known that the output of the network approximates the posterior probability, see, e.g., \cite[Sec. 6.11]{10.5555/525960}, \cite{RichardMichaelD.1991NNCE}, that is, $\Phi_k(\bar{\s{Z}}_1^\ell) \approx \P(\calH_k \vert \bar{\s{Z}}_1^\ell)$, and as so, $\xi\approx1$.. We start with the following result, which shows that $\s{T}_{\s{GNN}}$ is bounded with high probability.
\begin{theorem}[Exponentially bounded stopping time]\label{th:gnn_bounded_stop_time}
Fix $k\in[M]$. Assume that $\xi<\infty$ and,
\begin{align}
\min_{j\neq k\in[M]}\max_{z\in\bar{\calZ}}\bar{S}_{k,j}(z)<1.
\end{align}
Then, for any $t\in\mathbb{R}_+$, we have,
\begin{align}
    \P\pp{\left.\s{T}_{\s{GNN}}>t\right|\calH_k}\leq \s{C}_1\cdot\exp(-\s{C}_2\cdot t),
\end{align}
for some $\s{C}_1,\s{C}_2\in\mathbb{R}_+$.
\end{theorem}
Next, in a similar fashion to Theorem~\ref{th:gnn_error_guarantee}, we have the following bounds on the error probabilities associated with the msprtGNN procedure. We let $\{\calE^{\s{GNN}}_{j,k}\}$, $\{\calE^{\s{GNN}}_k\}$, and $\calE^{\s{GNN}}$, be defined as in Theorem~\ref{th:gnn_error_guarantee}, but with $\hat{\calH}_{\s{MSPRT}}$ replaced by $\hat{\calH}_{\s{GNN}}$.
\begin{theorem}[Error guarantees]\label{th:gnn_error_guarantee}
Fix $k\in[M]$ and assume that $\xi<\infty$. Then,
\begin{align}
    \calE^{\s{GNN}}_{k}&=\sum_{j:j \neq k}\calE^{\s{GNN}}_{j,k}
  \le a_k \xi + \xi -1,\label{eqn:errb1}\\
  \calE^{\s{GNN}} &\triangleq \sum_{k\in[M]} \calE^{\s{GNN}}_k
  \leq \xi\norm{a}_1 + M(\xi -1).\label{eqn:errb2}
\end{align}
If, in addition, $a_\ell = \nu$, for all $\ell\in[M]$, and for some $\nu\in[0,1]$, then,
\begin{align}
    \calE^{\s{GNN}} \leq 1 - \frac{1}{\xi(1+\nu)}.\label{eqn:errb3}
\end{align}
\end{theorem}

Finally, similarly to Theorem~\ref{th:asymptotic_time}, it turns out that, at least asymptotically, the msprtGNN stopping time converges in expectation to a similar limit as that of MSPRT, in the small error probability regime; here, we can see from Theorem~\ref{th:gnn_error_guarantee} that this regime corresponds to the asymptotics $\norm{a}_\infty\to0$ and $\xi\to1$. For any $z\in\calZ$, we define,
\begin{align}
    \chi^2(\bar{\alpha}_k(\cdot\vert z),\bar{\alpha}_{j}(\cdot\vert z)) \triangleq \int_{z'\in\bar{\calZ}}\frac{\bar{\alpha}^2_k(z'\vert z)}{\bar{\alpha}_j(z'\vert z)}\mathrm{d}z'-1.
\end{align}
We have the following result.
\begin{theorem}[Asymptotic stopping time]\label{th:gnn_asymptotic_time}
Fix $k\in[M]$. Assume that $\xi$ is finite, and
\begin{align}
    &\min_{j\neq k\in[M]}\max_{z\in\calZ} \chi^2(\bar{\alpha}_k(\cdot\vert z)||\bar{\alpha}_{j}(\cdot\vert z)) < \infty.
\end{align}
Then,
\begin{align}    
    \lim_{\xi\to1}\lim_{\norm{a}_\infty\to0} \frac{\s{T}_{\s{GNN}}}{- \log a_k} = \frac{1}{\min_{j \neq k} d_{\s{KL}}\left(\bar{\alpha}_k||\bar{\alpha}_j\vert\pi_k^{\s{stat}}\right)},
\end{align}
$f_k$-almost surely. Furthermore,
\begin{align}
    \lim_{\norm{a}_\infty\to0} \frac{\E\left[\s{T}_{\s{GNN}}\right]}{- \log a_k} = \frac{1}{\min_{j \neq k} d_{\s{KL}}\left(\bar{\alpha}_k||\bar{\alpha}_j\vert\pi_k^{\s{stat}}\right)}.
\end{align}
\end{theorem}

\subsection{Offline algorithm}\label{subsection:dataAlgo}
\sloppy
In order to use the MSPRT algorithm, the transition probabilities and initial probabilities are needed. Furthermore, we need to classify/convert the information traces $\{(\mathbf{x}_{u_\ell},\mathbf{x}_{v_\ell})\}_{\ell\geq1}$ into their edge-representation sequence $\{\s{Z}_{\ell}\}_{\ell\geq1}$. Accordingly, we next describe an offline algorithm which learn the above unknowns from training. Specifically, the data required for our offline algorithm is a set of $N$ information traces, denoted by $\{\mathcal{R}_\ell\}_{\ell=1}^N$, where $\mathcal{R}_\ell$ is $\ell$th information trace. We assume that each trace is labeled, and denote by $\calL_\ell \in \left[M\right]$ the information type of the $\ell$th trace.

\paragraph{Assigning edge types.} It should be clear that many techniques are possible for classifying edges in the information traces. Here, we propose the following technique, which proves quite efficient and successful in practice. In fact, as will be seen in our experiments, this technique outperforms other methods in the literature, e.g., \cite{wei2019quickstop,orenloberman2023online}. Generally speaking, our classification procedure is based on applying a certain pairing function $\Theta$ on the outputs of the GNN architecture, proposed in the previous subsection. Specifically, we are given labeled training information traces $\{(\mathcal{R}_\ell,\mathcal{L}_\ell)\}_{\ell=1}^N$. We are also given a trained msprtGNN architecture; in Section~\ref{sec:exper} we explain the training process in detail. The classification procedure follows the following steps:
\begin{enumerate}
    \item We create a set $\calE_{\s{train}}$ of all edges in $\{{\mathcal{R}_\ell}\}_{\ell=1}^N$.
    \item For each pair of users $e=(u,v)\in\calE_{\s{train}}$, with features $\mathbf{x}_u$ and $\mathbf{x}_v$, we denote by $\Phi^{e} = g_{\s{GNN}}(\mathbf{x}_u,\mathbf{x}_v)$ the vector of $M$ outputs of the msprtGNN, as defined in \eqref{eqn:phidef}, using the already trained msprtGNN architecture.
    %\item \deleted{We cluster the set of $M$-dimensional vectors $\{\Phi^{(e)}\}_{e\in\calE_{\s{train}}}$ into $|\calZ|$ classes using the $k$-means algorithm \cite{Lloyd82}, with $k=|\calZ|$. We denote the resulted trained clustering algorithm by $\mathcal{C}_{\s{means}}$.}
    \item We define $\Theta$ to be the pairing function which takes the indices of the largest and second largest entries in the vector $\Phi^{(e)}$, and maps them to a number between in $[|\calZ|]$. Specifically, 
    \begin{align}
        \Theta\triangleq \ell_1\cdot (M-1)+\ell_2-\mathds{1}_{\ell_2>\ell_1},\label{eq:Theta}
    \end{align} 
    where $\ell_1$ and $\ell_2$ are the indices of the largest and second largest entries in $\Phi^{(e)}$, respectively. For example, for $\calZ=\{0,1,2\}$, the pairing function/mapping $\Theta$ works as follows: $(0,1)\to0$, $(0,2)\to1$, $(1,0)\to2$, $(1,2)\to3$, $(2,0)\to4$, and $(2,1)\to5$. We would like to emphasize here that other pairing functions are possible, but the above proved successful in our experiments. 
    \item Each edge is assigned with an edge type using $g \triangleq \Theta \circ g_{\s{GNN}} : \calE \to \calZ$.
\end{enumerate}
The above edge classification method is summarized in Algorithm~\ref{alg:edge}. 

%Therefore, when using it to classify edges, we refer to msprtGNN the edge classifier, denoted as $f((\mathbf{x}_u, \mathbf{x}_v))$. The inputs to the edge classifier are the feature vectors of both nodes of the edge, $(\mathbf{x}_u, \mathbf{x}_v)$. The output of the edge classifier is a vector $\Phi_j, j\in[\s{M}]$, defined in equation \eqref{eqn:phidef}. Next we aim to train a clustering function that will cluster $\Phi_j, j\in[\s{M}]$ into one of $|\calZ|$ clusters. For this purpose, we decompose the training data to a list of individual edges, and classify each edge $(u,v)=e$ using the edge classifier, resulting in a set of edge classifier outputs, $\{\Phi^{(e)}\}_{e\in\text{training}}$. We then use this set to train a k-means clustering model denoted $\Theta(\Phi)$. The number of clusters, $k=|\calZ|$, is a hyperparameter.

%K-means clustering is an unsupervised learning algorithm that partitions observations into  distinct k clusters, where each data point belongs to the cluster with the nearest mean. The algorithm iteratively updates the cluster centers (means) and reassigns data points to the nearest cluster until convergence.

\begin{algorithm}[t!]
\caption{Edge type classifier}\label{alg:edge}
\textbf{Input} {Edge classifier $g_{\s{GNN}}$, training information traces $\{{\mathcal{R}_\ell}\}_{\ell=1}^N$ with labels $\ppp{\mathcal{L}_\ell}_{\ell=1}^N$}
\begin{algorithmic}[1]
    \State Let $\calE_{\s{train}}$ be the set of all edges participating in the information traces $\{{\mathcal{R}_\ell}\}_{\ell=1}^N$.
    \State Find $\Phi^{(e)}=g_{\s{GNN}}(\mathbf{x}_u, \mathbf{x}_v)$, for all $e=(u, v)\in\calE_{\s{train}}$.
    %\State \deleted{Train the naive $|\calZ|$-means clustering model $\Theta$ using $\{\Phi^{(e)}\}_{e\in\calE}$.}
\end{algorithmic}
\textbf{Return} {Edge type classifier $g(\cdot,\cdot) = \Theta(g_{\s{GNN}}(\cdot,\cdot))$ where $\Theta$ is defined in \eqref{eq:Theta}.}
\end{algorithm}

\paragraph{Initial probabilities.} To estimate the probability of an edge $z \in \calZ$ to forward information of type $m\in[M]$ directly from the source $s$, we use the simple frequentist estimator as follows,
    \begin{equation} \label{eq:initial_p}
    \hat{\eta}_m(z) = \frac{\sum_{k=1}^N \sum_{\ell=1}^{|\mathcal{R}_k|-1}\mathds{1}_{\{e_\ell^{\mathcal{R}_k} \in \mathcal{N}_s, \s{Z}_{\ell}^{\mathcal{R}_k} = z\}}\mathds{1}_{\{\mathcal{L}_k = m\}}}{\sum_{k=1}^N \mathds{1}_{\{\mathcal{L}_k = m\}}},
    \end{equation}
where $e_{\ell}^{\mathcal{R}_k}$ is the $\ell$th edge in trace $\mathcal{R}_k$, and $\s{Z}_{\ell}^{\mathcal{R}_k}$ is the $\ell$th edge representation in the $k$th information trace $\mathcal{R}_k$. Indeed, the numerator counts the number neighbors/children of the source $s$ with edge-type $z$ over all information traces with label $m$, while the denominator counts the number of traces with label $m$.  

\paragraph{Transition probabilities.}
As described in Subsection~\ref{subsection:GNN}, the transition matrix has $\mathcal{O}(\vert\calZ\vert^2)$ entries, which induces a high computational and sample complexity. Since our dataset is of fixed size, this implies an increasingly worse estimate as $M$ increases. To deal with this issue, we follow a Bayesian estimation approach and apply the Dirichlet-Categorical Bayesian (DCB) model \cite{alma990021928420204146}, which we describe next. Let $\textbf{p}_i^{(m)}$ denote the $i$th row of the $m$th transition matrix $\alpha_m(\cdot\vert\cdot)$. Recall that in the Bayesian approach we use probabilities to describe our initial uncertainty about the unknown parameters (in our case the transition probabilities), i.e., a \emph{prior distribution}, and then use probabilistic reasoning (that is, Bayes rule) to take into account our observations, namely, a \emph{posterior distribution}. Accordingly, in the DCB model, we treat $\textbf{p}_i^{(m)}$ as a random vector, with a Dirichlet prior distribution with parameters $\theta_i^{(m)}$, for $i\in[|\calZ|]$ and $m\in[M]$. To find these parameters, we use the frequentist estimate as follows,
\begin{align}
    & \theta_{i}^{(m)} = \frac{\sum_{k=1}^N \sum_{\ell=1}^{|\mathcal{R}_k|-1}\mathds{1}_{\{\s{Z}_{\ell}^{\mathcal{R}_k} = i\}}\mathds{1}_{\{\calL_k = m\}}}{\sum_{k=1}^N \mathds{1}_{\{\calL_k = m\}}},
\end{align}
for all $m\in[M]$ and $i\in[|\calZ|]$. %, where $\s{Z}_{\ell}^{\mathcal{R}_k}$ is the $\ell$th edge representation in the $k$th information trace $\mathcal{R}_k$. 
Then, 
\begin{align}
    &\textbf{p}_i^{(m)} \stackrel{\mathrm{i.i.d.}}{\sim} \s{Dirichlet}\left(\theta_i^{(m)}\right).
\end{align}
The sample complexity of this estimator is $\mathcal{O}(\vert\calZ\vert)$, and thus scales better than the $\mathcal{O}(\vert\calZ\vert^2)$ complexity we discussed above. Given the Dirichlet prior, it is well-known that the posterior distribution is Dirichlet as well, see, e.g., \cite[Prop. 17.3]{alma990021928420204146}. The parameters of this posterior distributions are calculated as follows. We first find the number of transitions $\s{N}_{ij}$ from state $i\in[|\calZ|]$ to state $j\in[|\calZ|]$, and the total number $\s{N}_i$ of state $i\in[|\calZ|]$ in the training information traces $\{\mathcal{R}_\ell\}_{\ell=1}^N$. We also introduce ``pseudo-count" hyperparamter $s$, which controls the weight given to the prior distribution. Then, in the DCB model, the posterior estimate for the $m$th matrix transition probabilities is,
\begin{align}\label{eq:transition_p}
    &\hat{\alpha}_m(j\vert i) = \frac{\s{N}_{ij} + s\cdot\theta_{i}^{(m)}}{\s{N}_i + s},
\end{align}
for $i,j\in[|\calZ|]$ and $m\in[M]$. Finally, a pseudo-code for the data-driven learning procedure proposed above is given in Algorithm~\ref{alg:offline}.

\begin{algorithm}[t!]
\caption{Offline Algorithm}\label{alg:offline}
\textbf{Input} {Edge classifier $g_{\s{GNN}}(\cdot,\cdot)$, training information traces $\{{\mathcal{R}_\ell}\}_{\ell=1}^N$ with labels $\ppp{\mathcal{L}_\ell}_{k=1}^N$.}
\begin{algorithmic}[1]
    \State Train the edge type classifier $g(\cdot,\cdot)$ using Algorithm \ref{alg:edge}.
    \State Assign each edge $e = (u, v) \in \calE$ an edge type $\s{Z}$ using $g(\cdot,\cdot)$.
    \State For all $j\in[M]$ and $z \in \calZ$ estimate $\hat{\eta_j}(z)$ using \eqref{eq:initial_p}.
    \State For all $j\in[M]$ and $z,z' \in \calZ$ estimate $\hat{\alpha_j}(z\vert z')$ using \eqref{eq:transition_p}.
\end{algorithmic}
\textbf{Return} {$g(\cdot,\cdot), \hat{\alpha}_j(\cdot|\cdot), \hat{\eta}_j(\cdot)$.}
\end{algorithm}

\section{Experiments}\label{sec:exper}

In this section, we test our algorithms over real-world datasets, and compare them to several state-of-the-art algorithms. We start by describing the datasets we use, followed by a description of the experimental setting we rely on, and finally, we present and discuss our results.

\subsection{Datasets}
\paragraph{UPFD dataset.} The UPFD dataset \cite{dou2021user} is composed of tree structured retweet graphs, crawled from the Twitter platform, and labeled as fake or real news, using the fact-checking websites ``Politifact" and ``Gossipcop". To obtain multiple classes, we merged the UPFD-Politifact and UPFD-Gossipcop datasets into a single 4-class dataset, i.e., $M=4$, with labels: ``true gossip news", ``fake gossip news", ``true politics news", and ``fake politics news". We also created a 3-class dataset, i.e., $M=3$, with labels: ``true news", "false gossip news", and "false politics news". Since the datasets are unbalanced we downsampled Gossipcop which is the larger one. Each node in the tree has ten profile features: ``verified", ``enabled geo-spatial positioning", ``followers count", "friends count", ``status count", ``favorite count", ``number of lists", ``created time", ``number of words in description", and ``number of words in screen name". We also perform an ablation study using ``content" features, a concatenation of the ``profile" features with ``Spacy features" \cite{honnibal_spacy_2018}; Spacy features are 300 features which are an average of the word vectors used in the last 200 tweets of each user. We will present the results of this ablation study towards the end of this section.  We split the datasets into 80\% for training and 20\% for testing. Table~\ref{tab:comparison} includes several statistics for the above UPFD dataset.

\paragraph{Weibo dataset.} The Weibo dataset \cite{10.5555/3061053.3061153} is also composed of tree-structured graphs, representing the retweet graph, similarly to UPFD. It also holds the original posts. To get multiple classes, we used openAI API to perform sentiment analysis on the original posts. We created a 3-class dataset, with labels: ``true news", ``false subjective news", and ``false objective news". Each node has four profile features: ``statuses count", ``friends count", ``followers count", and ``user's age". We split the dataset into 80\% for training and 20\% for testing. Table~\ref{tab:comparison} show several statistics for the Weibo dataset.

\begin{table}[t!]
    \centering
    \begin{tabular}{|c|c|c|c|c|}
        \hline
        \textbf{Dataset} & \textbf{\#edges/graph} & \textbf{\#nodes/graph} & \textbf{\#graphs/class} & \textbf{\#graphs}\\
        \hline
        \texttt{UPFD3}        & 87.5  & 88.5  & 157   & 471\\
        \texttt{UPFD4}        & 93    & 94    & 157   & 628\\
        \texttt{Weibo2}     & 988   & 517   & 1647   & 3294\\
        \texttt{Weibo3}     & 995   & 515   & 585   & 1755\\
        \hline
    \end{tabular}
    \caption{UPFD and Weibo datasets statistics.}
    \label{tab:comparison}
\end{table}

\subsection{Baselines}

In this subsection, we present our baselines for comparison. In general, we train all those baselines using the same train-test split mentioned above. We fine-tuned the hyper-parameters (e.g., hidden layers dimension sizes, learning rates, weight decay) for each baseline, to achieve maximal accuracy on the test set. For the purpose of evaluating the test risk, we set the propagation cost to be $c_j=10^{-3}$, for all $j\in[M]$. All the baselines, except HGFND \cite{10020234}, which we describe below, receive the propagation tree as an input. The baselines we compare our algorithms to are:
\begin{itemize}
    \item Na\"ive i.i.d. MSPRT \cite{340472}: 
    This baseline model applies the vanilla MSPRT method \cite{340472}, while assuming that $\{\s{Z}_\ell\}_{\ell\geq1}$ is a sequence of i.i.d. random variables. It uses the same $\s{Z}$ values computed for MSPRT, though it learns the frequencies of these $\s{Z}$ values, instead of a transtiion matrix. The online algorithm for the naive method is similar to Algorithm~\ref{alg:online}, though when applying Equation \eqref{eq:msprt} it assumes the sequence to be a sequence of i.i.d random variabbles.
    
    \item Quickstop \cite{wei2019quickstop}: This baseline models the \emph{temporal} sequence of edge-types $\{\s{Z}_\ell\}_{\ell\geq1}$ as a single Markov Chain. We modified the original Quickstop method, designed for the two classes, to apply for multiple classes.%, using \texttt{MSPRT}\cite{340472}.

    %\itemmarkovMSPRT: This baseline is a generalized version of \cite{orenloberman2023online}. We used \texttt{MSPRT} \cite{340472} instead of \texttt{SPRT} \cite{10.1214/aoms/1177730197}. Furthermore, we improved it by we using msprtGNN for assigning edge-types, instead of an SVM. Furthermore, we estimate the transition probabilities using a Dirichlet-Categorical Bayesian (DCB) model \cite{alma990021928420204146} instead of a simple frequentist estimation.

    \item UPFD-Sage \cite{dou2021user}: This model is a fake-news classifier, and has three versions: using GCN \cite{DBLP:journals/corr/KipfW16}, GAT \cite{veličković2018graphattentionnetworks}, and GraphSAGE \cite{DBLP:journals/corr/HamiltonYL17} convolutional layer. We chose the GraphSAGE version since it performed best on the test set. We modified the output to support for multiple classes.

    \item GCNFN \cite{monti2019fake}: This model uses two \textbf{GCN} \cite{DBLP:journals/corr/KipfW16} layers. We modified the output to support for multiple classes.

    \item HGFND \cite{10020234}: This method models the social network as a hypergraph \cite{DBLP:journals/corr/abs-1809-09401}. The nodes are news stories, connected by hyperedges. A hyperedge connects two news stories if they were shared by the same user, posted at approximately the same time, or mention similar entities. Because this method requires preprocessing a large portion of the test set prior to testing, it is not suitable for the sequential setting. Nevertheless, because this method achieved state-of-the-art results on the binary UPFD \cite{dou2021user} dataset, it is interesting to compare it to other methods in the multiclass, non-sequential setting (see, Table~\ref{tab:ablation1}).
\end{itemize}

\subsection{Experimental setting}

\paragraph{Training.} To train msprtGNN we used the cross entropy loss as our loss function, and used the AdamW optimizer \cite{DBLP:journals/corr/abs-1711-05101}, with learning rate $10^{-3}$ and weight-decay $10^{-2}$. Training was done in a non-sequential setting, using the whole information trace as input. We noticed that in the non-sequential setting, training msprtGNN with a mean-pooling layer \cite{DBLP:journals/corr/abs-1810-00826} achieves better accuracy, while during inference, we get better accuracy if we use the add-pooling layer \cite{DBLP:journals/corr/abs-1810-00826}; therefore, we follow this approach. This method works because the difference between mean-pooling and add-pooling is merely a multiplication by a constant, and thus does not change the order of outputs of the proceeding \texttt{softmax} layer. For the Weibo dataset, we use a hidden dimension of size $h=64$, while for the UPFD dataset, we use $h=128$. In the ablation study, where we use augmented content features, we use a hidden dimension of size $h=512$. Finally, we set $|\calZ|=2\cdot\s{M}$, for the $k$-means clustering step.

\paragraph{Experiment procedure.} To compare our methods against the baselines, we perform the experiment in the following manner. Each information trace in the test set is tested sequentially. A predication is made for each time step. We estimate the accuracy for each time step $t$ by counting the overall correct predictions done at $t$, and dividing it by the overall number of predicitions done at $t$. Another metric we use to compare the different models is the average accuracy, also known as the area under curve (AUC). This metric is sensitive both to accuracy and detection time, and therefore adequate in evaluating the overall performance of each classifier.

\subsection{Results}

Table~\ref{fig:comparison_tables} shows the accuracy achieved by the different baselines and our algorithms, at different time steps, as well as the AUC, for four databases described above. It is clear that msprtGNN achieves the highest AUC across the fours scenarios we tested. It is interesting to note that when examining the performance of MSPRT, na\"ive MSPRT, and Quickstop, they tend to flatline, and even decrease in accuracy after 40 reposts in the information trace. We believe this happens because these methods assume the process to be time-invariant/homogeneous, whereas in reality it may not be the case; msprtGNN do not rely on such an assumption, which may explain its superiority.

\begin{figure}[t!]
\captionsetup[subfigure]{justification=centering}
\begin{subfigure}[b]{0.3\textwidth}
%\begin{table}[H]
%    \centering
\begin{tabular}{|c|c|c|c|c|c|c|c|c|c|}
\hline
\textbf{Alg.} & $t=1$ & $5$ & $10$ & $15$ & $20$ & $30$ & $40$ & Full & AUC\\
\hline
\texttt{msprtGNN}   & 0.737 & \textbf{0.856} & \textbf{0.901} & \textbf{0.882} & \textbf{0.875} & \textbf{0.908} & 0.906 & \textbf{0.905} & \textbf{0.884}\\
\texttt{UPFD-Sage}  & \textbf{0.768} & 0.822 & 0.864 & 0.855 & 0.861 & \textbf{0.908} & 0.925& \textbf{0.905} & 0.878\\
\texttt{GCNFN}      & 0.758 & 0.833 & 0.864 & 0.882 & 0.861 & \textbf{0.908} & \textbf{0.943} & 0.884 & 0.878\\
\texttt{Na\"ive}      & 0.737 & 0.822 & 0.877 & 0.855 & 0.847 & 0.815 & 0.868 & - & 0.849\\
\texttt{MSPRT}& 0.737 & 0.822 & 0.877 & 0.855 & 0.861 & 0.815 & 0.849 & - & 0.846\\
\texttt{Quickstop}  & 0.737 & 0.822 & 0.827 & 0.829 & 0.833 & 0.815 & 0.887 & - & 0.832\\
\hline
\end{tabular}
\caption{3-Class UPFD.}
\label{tab:comparison1}
%\end{table}
\end{subfigure}

\begin{subfigure}[b]{0.3\textwidth}
%\begin{table}[H]
%    \centering
\begin{tabular}{|c|c|c|c|c|c|c|c|c|c|}
\hline
\textbf{Alg.} & $t=1$ & $5$ & $10$ & $15$ & $20$ & $30$ & $40$ & Full & AUC\\
\hline
\texttt{msprtGNN}       & \textbf{0.786} & \textbf{0.846} & \textbf{0.861} & \textbf{0.863} & \textbf{0.878} & \textbf{0.890} & 0.868 & 0.881& \textbf{0.863} \\
\texttt{UPFD-Sage}      & 0.778 & 0.812 & 0.778 & 0.824 & 0.827 & 0.854 & \textbf{0.868} & \textbf{0.897}& 0.829 \\
\texttt{GCNFN}          & \textbf{0.786} & 0.821 & 0.806 & 0.824 & 0.847 & 0.841 & 0.853 & 0.873& 0.836 \\
\texttt{Na\"ive}          & \textbf{0.786} & 0.838 & 0.833 & 0.853 & 0.857 & 0.866 & 0.868 & -& 0.851 \\
\texttt{MSPRT}    & \textbf{0.786} & \textbf{0.846} & 0.833 & 0.853 & 0.847 & 0.841 & 0.838 & -& 0.839 \\
\texttt{Quickstop}      & 0.770 & 0.795 & 0.778 & 0.775 & 0.755 & 0.768 & 0.765 & -& 0.768 \\

\hline
\end{tabular}
\caption{4-class UPFD.}
\label{tab:comparison2}
%\end{table}
\end{subfigure}

\begin{subfigure}[b]{0.3\textwidth}
%\begin{table}[H]
%\centering
\begin{tabular}{|c|c|c|c|c|c|c|c|c|c|}
\hline
\textbf{Alg.} & $t=1$ & $5$ & $10$ & $15$ & $20$ & $30$ & $40$ & Full & AUC\\
\hline
\texttt{msprtGNN}       & \textbf{0.797} & \textbf{0.841} & \textbf{0.853} & \textbf{0.858} & \textbf{0.853} & \textbf{0.856} & 0.854 & 0.862& \textbf{0.851} \\
\texttt{UPFD-Sage}      & 0.730 & 0.768 & 0.783 & 0.785 & 0.810 & 0.823 & 0.832 & \textbf{0.900}& 0.799 \\
\texttt{GCNFN}          & 0.676 & 0.733 & 0.759 & 0.782 & 0.802 & 0.812 & 0.821 & 0.888& 0.784 \\
\texttt{Na\"ive}          & 0.794 & 0.805 & 0.823 & 0.820 & 0.828 & \textbf{0.838} & \textbf{0.837} & -& 0.827 \\
\texttt{MSPRT}    & 0.794 & 0.815 & 0.830 & 0.829 & 0.833 & 0.832 & 0.830 & -& 0.830 \\
\texttt{Quickstop}     & 0.206 & 0.202 & 0.197 & 0.208 & 0.415 & 0.471 & 0.482 & -& 0.355 \\

\hline
\end{tabular}
\caption{2-class weibo dataset.}
\label{tab:comparison3}
%\end{table}
\end{subfigure}

\begin{subfigure}[b]{0.3\textwidth}
%\begin{table}[H]
%\centering
\begin{tabular}{|c|c|c|c|c|c|c|c|c|c|}
\hline
\textbf{Alg.} & $t=1$ & $5$ & $10$ & $15$ & $20$ & $30$ & $40$ & Full & AUC \\
\hline
\texttt{msprtGNN} & 0.588 & \textbf{0.616} & \textbf{0.613} & \textbf{0.611} & 0.609 & \textbf{0.608} & 0.611 & 0.507& \textbf{0.609} \\
\texttt{UPFD-Sage} & 0.449 & 0.480 & 0.510 & 0.540 & 0.554 & 0.566 & 0.577 & \textbf{0.514} & 0.540 \\
\texttt{GCNFN} & 0.494 & 0.509 & 0.538 & 0.560 & 0.571 & 0.596 & \textbf{0.614} & 0.495& 0.565 \\
\texttt{Na\"ive} & 0.574 & 0.602 & 0.598 & 0.603 & 0.606 & 0.590 & 0.571 & - & 0.594 \\
\texttt{MSPRT} & \textbf{0.591} & \textbf{0.616} & 0.598 & 0.606 & \textbf{0.612} & 0.584 & 0.583 & -& 0.597 \\
\texttt{Quickstop} & \textbf{0.591} & 0.585 & 0.581 & 0.583 & 0.583 & 0.578 & 0.567 & -& 0.581 \\

\hline
\end{tabular}
\caption{3-class weibo dataset.}
\label{tab:comparison4}
%\end{table}
\end{subfigure}
\caption{Accuracy as a function of time step.}
\label{fig:comparison_tables}
\end{figure}

\begin{figure}[t!]
%\captionsetup[subfigure]{justification=centering}
\begin{subfigure}{0.3\textwidth}
    \pgfplotsset{compat=1.18}

%\begin{document}
\begin{tikzpicture}
    \begin{axis}[
        width=7cm,
        height=7cm,
        xlabel={Decision deadline},
        ylabel={Accuracy},
        legend style={at={(1.05,1)},anchor=north west},
        grid=major,
        xmin=0, xmax=39,
        ymin=0.7, ymax=0.95,
        xtick={0, 5, 10, 15, 20, 25, 30, 35, 40},
        ytick={0.7, 0.75, 0.8, 0.85, 0.9, 0.95, 1.00},
        legend cell align={left}
    ]

\addplot[color=blue, mark=square*] coordinates
{(1,0.7857142857142857) (2,0.8015873015873016) (3,0.8278688524590164) (4,0.8512396694214877) (5,0.8461538461538461) (6,0.8632478632478633) (7,0.8558558558558559) (8,0.8558558558558559) (9,0.8545454545454545) (10,0.8611111111111112) (11,0.8504672897196262) (12,0.8598130841121495) (13,0.8666666666666667) (14,0.8557692307692307) (15,0.8627450980392157) (16,0.8811881188118812) (17,0.8787878787878788) (18,0.8787878787878788) (19,0.8787878787878788) (20,0.8775510204081632) (21,0.8762886597938144) (22,0.8586956521739131) (23,0.8478260869565217) (24,0.8478260869565217) (25,0.8539325842696629) (26,0.8539325842696629) (27,0.8505747126436781) (28,0.8690476190476191) (29,0.8674698795180723) (30,0.8902439024390244) (31,0.8888888888888888) (32,0.8987341772151899) (33,0.8961038961038961) (34,0.881578947368421) (35,0.88) (36,0.8767123287671232) (37,0.8767123287671232) (38,0.875) (39,0.8732394366197183) (40,0.8676470588235294)};
%\addlegendentry{msprtGNN}
\addplot[color=red, mark=] coordinates {(1,0.7777777777777778) (2,0.8015873015873016) (3,0.7950819672131147) (4,0.8016528925619835) (5,0.811965811965812) (6,0.811965811965812) (7,0.8108108108108109) (8,0.8108108108108109) (9,0.7909090909090909) (10,0.7777777777777778) (11,0.7757009345794392) (12,0.8037383177570093) (13,0.819047619047619) (14,0.8269230769230769) (15,0.8235294117647058) (16,0.8217821782178217) (17,0.8282828282828283) (18,0.8383838383838383) (19,0.8282828282828283) (20,0.826530612244898) (21,0.8144329896907216) (22,0.8152173913043478) (23,0.8152173913043478) (24,0.8152173913043478) (25,0.8202247191011236) (26,0.8202247191011236) (27,0.8275862068965517) (28,0.8333333333333334) (29,0.8313253012048193) (30,0.8536585365853658) (31,0.8518518518518519) (32,0.8607594936708861) (33,0.8701298701298701) (34,0.881578947368421) (35,0.88) (36,0.8767123287671232) (37,0.8767123287671232) (38,0.875) (39,0.8732394366197183) (40,0.8676470588235294)};
%\addlegendentry{upfd-sage}
\addplot[color=green, mark=triangle] coordinates
{(1,0.7857142857142857) (2,0.7777777777777778) (3,0.7786885245901639) (4,0.8099173553719008) (5,0.8205128205128205) (6,0.8376068376068376) (7,0.8288288288288288) (8,0.8198198198198198) (9,0.8181818181818182) (10,0.8055555555555556) (11,0.8130841121495327) (12,0.8130841121495327) (13,0.8285714285714286) (14,0.8269230769230769) (15,0.8235294117647058) (16,0.8217821782178217) (17,0.8383838383838383) (18,0.8383838383838383) (19,0.8383838383838383) (20,0.8469387755102041) (21,0.8350515463917526) (22,0.8369565217391305) (23,0.8369565217391305) (24,0.8478260869565217) (25,0.8426966292134831) (26,0.8426966292134831) (27,0.8390804597701149) (28,0.8452380952380952) (29,0.8433734939759037) (30,0.8414634146341463) (31,0.8641975308641975) (32,0.8734177215189873) (33,0.8701298701298701) (34,0.868421052631579) (35,0.8666666666666667) (36,0.863013698630137) (37,0.863013698630137) (38,0.8611111111111112) (39,0.8591549295774648) (40,0.8529411764705882)};
%\addlegendentry{gcnfn}
\addplot[color=orange, mark=] coordinates {(1,0.7857142857142857) (2,0.7857142857142857) (3,0.8442622950819673) (4,0.7933884297520661) (5,0.8376068376068376) (6,0.8461538461538461) (7,0.8468468468468469) (8,0.8468468468468469) (9,0.8454545454545455) (10,0.8333333333333334) (11,0.8317757009345794) (12,0.8411214953271028) (13,0.8380952380952381) (14,0.8461538461538461) (15,0.8529411764705882) (16,0.8613861386138614) (17,0.8686868686868687) (18,0.8585858585858586) (19,0.8686868686868687) (20,0.8571428571428571) (21,0.8556701030927835) (22,0.8478260869565217) (23,0.8478260869565217) (24,0.8478260869565217) (25,0.8651685393258427) (26,0.8651685393258427) (27,0.8505747126436781) (28,0.8571428571428571) (29,0.8554216867469879) (30,0.8658536585365854) (31,0.8641975308641975) (32,0.8734177215189873) (33,0.8701298701298701) (34,0.868421052631579) (35,0.88) (36,0.8767123287671232) (37,0.863013698630137) (38,0.8611111111111112) (39,0.8591549295774648) (40,0.8676470588235294)};
%\addlegendentry{naive}
\addplot[color=purple, mark=o] coordinates
{(1,0.7857142857142857) (2,0.7698412698412699) (3,0.8442622950819673) (4,0.8347107438016529) (5,0.8461538461538461) (6,0.8461538461538461) (7,0.8378378378378378) (8,0.8378378378378378) (9,0.8363636363636363) (10,0.8333333333333334) (11,0.8317757009345794) (12,0.8317757009345794) (13,0.8285714285714286) (14,0.8461538461538461) (15,0.8529411764705882) (16,0.8514851485148515) (17,0.8383838383838383) (18,0.8484848484848485) (19,0.8282828282828283) (20,0.8469387755102041) (21,0.845360824742268) (22,0.8478260869565217) (23,0.8369565217391305) (24,0.8586956521739131) (25,0.8651685393258427) (26,0.8426966292134831) (27,0.8390804597701149) (28,0.8214285714285714) (29,0.8192771084337349) (30,0.8414634146341463) (31,0.8395061728395061) (32,0.8481012658227848) (33,0.8441558441558441) (34,0.8552631578947368) (35,0.8533333333333334) (36,0.8356164383561644) (37,0.8493150684931506) (38,0.8611111111111112) (39,0.8450704225352113) (40,0.8382352941176471)};
%%\addlegendentry{markovMSPRT}

    \end{axis}
\end{tikzpicture}
%\end{document}
    \vspace*{-5mm}
         \caption{4-class UPFD dataset.}
         \label{fig:acc_time_upfd3_plot}
\end{subfigure}
\hspace{2.5cm}
\begin{subfigure}{0.3\textwidth}
         \pgfplotsset{compat=1.18}

%\begin{document}
\begin{tikzpicture}
    \begin{axis}[
        width=7cm,
        height=7cm,
        xlabel={Decision deadline},
        legend style={at={(0.305,0)},anchor=south west},
        grid=major,
        xmin=0, xmax=39,
        ymin=0.7, ymax=0.95,
        xtick={0, 5, 10, 15, 20, 25, 30, 35, 40},
        ytick={0.7, 0.75, 0.8, 0.85, 0.9, 0.95, 1},
        legend cell align={left}
    ]

\addplot[color=blue, mark=square*] coordinates
{(1,0.7368421052631579) (2,0.7684210526315789) (3,0.8279569892473119) (4,0.8369565217391305) (5,0.8555555555555555) (6,0.8295454545454546) (7,0.8705882352941177) (8,0.8470588235294118) (9,0.8795180722891566) (10,0.9012345679012346) (11,0.8607594936708861) (12,0.8846153846153846) (13,0.8831168831168831) (14,0.8831168831168831) (15,0.881578947368421) (16,0.8947368421052632) (17,0.8918918918918919) (18,0.8767123287671232) (19,0.8767123287671232) (20,0.875) (21,0.9154929577464789) (22,0.9154929577464789) (23,0.9014084507042254) (24,0.9014084507042254) (25,0.8985507246376812) (26,0.8985507246376812) (27,0.9117647058823529) (28,0.8939393939393939) (29,0.8923076923076924) (30,0.9076923076923077) (31,0.90625) (32,0.9032258064516129) (33,0.9032258064516129) (34,0.9166666666666666) (35,0.9310344827586207) (36,0.9310344827586207) (37,0.9285714285714286) (38,0.9285714285714286) (39,0.9259259259259259) (40,0.9056603773584906)};
\addlegendentry{msprtGNN}
\addplot[color=red, mark=] coordinates {(1,0.7684210526315789) (2,0.7894736842105263) (3,0.8064516129032258) (4,0.8260869565217391) (5,0.8222222222222222) (6,0.8295454545454546) (7,0.8352941176470589) (8,0.8470588235294118) (9,0.8433734939759037) (10,0.8641975308641975) (11,0.8607594936708861) (12,0.8717948717948718) (13,0.8701298701298701) (14,0.8831168831168831) (15,0.8552631578947368) (16,0.868421052631579) (17,0.8648648648648649) (18,0.863013698630137) (19,0.863013698630137) (20,0.8611111111111112) (21,0.8873239436619719) (22,0.8873239436619719) (23,0.9014084507042254) (24,0.9014084507042254) (25,0.8985507246376812) (26,0.8985507246376812) (27,0.9117647058823529) (28,0.9090909090909091) (29,0.9076923076923077) (30,0.9076923076923077) (31,0.90625) (32,0.9032258064516129) (33,0.9354838709677419) (34,0.9333333333333333) (35,0.9310344827586207) (36,0.9310344827586207) (37,0.9107142857142857) (38,0.9285714285714286) (39,0.9259259259259259) (40,0.9245283018867925)}; \addlegendentry{UPFD-Sage}
\addplot[color=green, mark=triangle] coordinates
{(1,0.7578947368421053) (2,0.7684210526315789) (3,0.7956989247311828) (4,0.8260869565217391) (5,0.8333333333333334) (6,0.8295454545454546) (7,0.8235294117647058) (8,0.8470588235294118) (9,0.8433734939759037) (10,0.8641975308641975) (11,0.8481012658227848) (12,0.8589743589743589) (13,0.8701298701298701) (14,0.8831168831168831) (15,0.881578947368421) (16,0.8552631578947368) (17,0.8513513513513513) (18,0.8493150684931506) (19,0.8493150684931506) (20,0.8611111111111112) (21,0.8873239436619719) (22,0.8873239436619719) (23,0.9014084507042254) (24,0.9154929577464789) (25,0.9130434782608695) (26,0.9130434782608695) (27,0.9117647058823529) (28,0.9090909090909091) (29,0.9076923076923077) (30,0.9076923076923077) (31,0.90625) (32,0.9193548387096774) (33,0.9193548387096774) (34,0.9166666666666666) (35,0.9137931034482759) (36,0.9310344827586207) (37,0.9285714285714286) (38,0.9464285714285714) (39,0.9444444444444444) (40,0.9433962264150944)};
\addlegendentry{GCNFN}
\addplot[color=orange, mark=] coordinates {(1,0.7368421052631579) (2,0.7894736842105263) (3,0.8387096774193549) (4,0.8152173913043478) (5,0.8222222222222222) (6,0.8295454545454546) (7,0.8235294117647058) (8,0.8470588235294118) (9,0.8674698795180723) (10,0.8765432098765432) (11,0.8354430379746836) (12,0.8846153846153846) (13,0.8571428571428571) (14,0.8701298701298701) (15,0.8552631578947368) (16,0.8552631578947368) (17,0.8648648648648649) (18,0.863013698630137) (19,0.863013698630137) (20,0.8472222222222222) (21,0.8732394366197183) (22,0.8732394366197183) (23,0.8591549295774648) (24,0.8732394366197183) (25,0.8405797101449275) (26,0.8405797101449275) (27,0.8382352941176471) (28,0.8333333333333334) (29,0.8153846153846154) (30,0.8153846153846154) (31,0.84375) (32,0.8548387096774194) (33,0.8548387096774194) (34,0.8666666666666667) (35,0.8793103448275862) (36,0.8620689655172413) (37,0.875) (38,0.875) (39,0.8703703703703703) (40,0.8679245283018868)}; \addlegendentry{Na\"ive}
\addplot[color=purple, mark=o] coordinates
{(1,0.7368421052631579) (2,0.7894736842105263) (3,0.8494623655913979) (4,0.8260869565217391) (5,0.8222222222222222) (6,0.8295454545454546) (7,0.8352941176470589) (8,0.8470588235294118) (9,0.8674698795180723) (10,0.8765432098765432) (11,0.8607594936708861) (12,0.8846153846153846) (13,0.8571428571428571) (14,0.8701298701298701) (15,0.8552631578947368) (16,0.8552631578947368) (17,0.8378378378378378) (18,0.863013698630137) (19,0.8493150684931506) (20,0.8611111111111112) (21,0.8591549295774648) (22,0.8591549295774648) (23,0.8309859154929577) (24,0.8450704225352113) (25,0.8260869565217391) (26,0.8260869565217391) (27,0.8235294117647058) (28,0.8181818181818182) (29,0.8153846153846154) (30,0.8153846153846154) (31,0.84375) (32,0.8387096774193549) (33,0.8548387096774194) (34,0.8666666666666667) (35,0.896551724137931) (36,0.8793103448275862) (37,0.875) (38,0.8928571428571429) (39,0.8518518518518519) (40,0.8490566037735849)};
\addlegendentry{MSPRT}
\end{axis}
\end{tikzpicture}
%\end{document}
         \vspace*{-5mm}
         \caption{3-class UPFD dataset}
         \label{fig:acc_time_upfd4_plot}
\end{subfigure}

\begin{subfigure}{0.3\textwidth}
         %\documentclass{standalone}
%\usepackage{pgfplots}
%\pgfplotsset{compat=1.18}
%\begin{document}
\begin{tikzpicture}
    \begin{axis}[
        width=7cm,
        height=7cm,
        xlabel={Decision deadline},
        ylabel={Accuracy},
        legend style={at={(0.3,0.01)},anchor=south west},
        grid=major,
        xmin=0, xmax=39,
        ymin=0.75, ymax=0.9,
        xtick={0, 5, 10, 15, 20, 25, 30, 35, 40},
        ytick={0.7, 0.75, 0.8, 0.85, 0.9, 0.95, 1},
        legend cell align={left}
    ]

\addplot[color=blue, mark=square*] coordinates
{(1,0.796969696969697) (2,0.8181818181818182) (3,0.8378787878787879) (4,0.8378787878787879) (5,0.8409090909090909) (6,0.843939393939394) (7,0.8469696969696969) (8,0.85) (9,0.85) (10,0.853030303030303) (11,0.8528072837632777) (12,0.8541033434650456) (13,0.8584474885844748) (14,0.8612804878048781) (15,0.8580152671755725) (16,0.8610687022900764) (17,0.8610687022900764) (18,0.8577981651376146) (19,0.8545176110260337) (20,0.8525345622119815) (21,0.8540706605222734) (22,0.8551617873651772) (23,0.8544891640866873) (24,0.8540372670807453) (25,0.8564742589703588) (26,0.8560250391236307) (27,0.8557993730407524) (28,0.8557993730407524) (29,0.8555729984301413) (30,0.8555729984301413) (31,0.8555729984301413) (32,0.8553459119496856) (33,0.8533123028391167) (34,0.8533123028391167) (35,0.8526148969889065) (36,0.8523809523809524) (37,0.8535031847133758) (38,0.8532695374800638) (39,0.8544) (40,0.8544)};
%\addlegendentry{msprtGNN}
\addplot[color=red, mark=] coordinates {(1,0.7303030303030303) (2,0.740909090909091) (3,0.7621212121212121) (4,0.7424242424242424) (5,0.7681818181818182) (6,0.7545454545454545) (7,0.7787878787878788) (8,0.7818181818181819) (9,0.7681818181818182) (10,0.7833333333333333) (11,0.7814871016691958) (12,0.7917933130699089) (13,0.776255707762557) (14,0.7942073170731707) (15,0.7847328244274809) (16,0.783206106870229) (17,0.7984732824427481) (18,0.8012232415902141) (19,0.8055130168453293) (20,0.8095238095238095) (21,0.8202764976958525) (22,0.8228043143297381) (23,0.8157894736842105) (24,0.8074534161490683) (25,0.8096723868954758) (26,0.8153364632237872) (27,0.8119122257053292) (28,0.8213166144200627) (29,0.8116169544740973) (30,0.8226059654631083) (31,0.8178963893249608) (32,0.8191823899371069) (33,0.8233438485804416) (34,0.8217665615141956) (35,0.8225039619651348) (36,0.8222222222222222) (37,0.8296178343949044) (38,0.8325358851674641) (39,0.8352) (40,0.832)}; %\addlegendentry{UPFD-Sage}
\addplot[color=green, mark=] coordinates {(1,0.6757575757575758) (2,0.706060606060606) (3,0.7424242424242424) (4,0.7439393939393939) (5,0.7333333333333333) (6,0.7484848484848485) (7,0.7333333333333333) (8,0.746969696969697) (9,0.7515151515151515) (10,0.759090909090909) (11,0.7693474962063733) (12,0.7720364741641338) (13,0.771689497716895) (14,0.7850609756097561) (15,0.7816793893129771) (16,0.7847328244274809) (17,0.7786259541984732) (18,0.8058103975535168) (19,0.8116385911179173) (20,0.8018433179723502) (21,0.8033794162826421) (22,0.802773497688752) (23,0.7894736842105263) (24,0.7903726708074534) (25,0.8018720748829953) (26,0.8012519561815337) (27,0.7962382445141066) (28,0.8056426332288401) (29,0.8053375196232339) (30,0.8116169544740973) (31,0.8116169544740973) (32,0.8128930817610063) (33,0.8170347003154574) (34,0.8170347003154574) (35,0.8129952456418383) (36,0.807936507936508) (37,0.8136942675159236) (38,0.8181818181818182) (39,0.824) (40,0.8208)}; %\addlegendentry{GCNFN}
\addplot[color=orange, mark=star] coordinates
{(1,0.793939393939394) (2,0.8121212121212121) (3,0.8015151515151515) (4,0.8090909090909091) (5,0.8045454545454546) (6,0.8136363636363636) (7,0.8196969696969697) (8,0.8136363636363636) (9,0.8257575757575758) (10,0.8227272727272728) (11,0.8270106221547799) (12,0.8237082066869301) (13,0.8280060882800608) (14,0.8307926829268293) (15,0.8198473282442749) (16,0.8259541984732824) (17,0.8244274809160306) (18,0.8211009174311926) (19,0.8192955589586524) (20,0.8248847926267281) (21,0.8341013824884793) (22,0.8335901386748844) (23,0.8312693498452013) (24,0.8291925465838509) (25,0.828393135725429) (26,0.8294209702660407) (27,0.829153605015674) (28,0.8338557993730408) (29,0.8351648351648352) (30,0.8367346938775511) (31,0.8367346938775511) (32,0.8380503144654088) (33,0.8391167192429022) (34,0.8359621451104101) (35,0.838351822503962) (36,0.8380952380952381) (37,0.839171974522293) (38,0.8389154704944178) (39,0.8368) (40,0.8368)};
%\addlegendentry{Na\"ive}
\addplot[color=purple, mark=o] coordinates
{(1,0.793939393939394) (2,0.8121212121212121) (3,0.8227272727272728) (4,0.8090909090909091) (5,0.8151515151515152) (6,0.8196969696969697) (7,0.8196969696969697) (8,0.8242424242424242) (9,0.8287878787878787) (10,0.8303030303030303) (11,0.8315629742033384) (12,0.8297872340425532) (13,0.832572298325723) (14,0.8323170731707317) (15,0.8290076335877863) (16,0.8290076335877863) (17,0.8305343511450382) (18,0.8302752293577982) (19,0.8346094946401225) (20,0.8325652841781874) (21,0.8356374807987711) (22,0.8335901386748844) (23,0.8359133126934984) (24,0.8385093167701864) (25,0.8393135725429017) (26,0.8419405320813772) (27,0.8401253918495298) (28,0.8385579937304075) (29,0.8383045525902669) (30,0.8320251177394035) (31,0.8320251177394035) (32,0.8333333333333334) (33,0.832807570977918) (34,0.832807570977918) (35,0.8351822503961965) (36,0.8365079365079365) (37,0.8407643312101911) (38,0.8373205741626795) (39,0.8352) (40,0.8304)};
%\addlegendentry{markovMSPRT}
\end{axis}
\end{tikzpicture}
%\end{document}
         \vspace*{-5mm}
         \caption{2-class Weibo dataset}
         \label{fig:acc_time_weibo2_plot}
\end{subfigure}
\hspace{2.5cm}
\begin{subfigure}{0.3\textwidth}
         %\documentclass{standalone}
%\usepackage{pgfplots}
%\pgfplotsset{compat=1.18}
%\begin{document}
\begin{tikzpicture}
    \begin{axis}[
        width=7cm,
        height=7cm,
        xlabel={Decision deadline},
        ylabel={},
        legend style={at={(0.305,0)},anchor=south west},
        grid=major,
        xmin=0, xmax=39,
        ymin=0.5, ymax=0.65,
        xtick={0, 5, 10, 15, 20, 25, 30, 35, 40},
        ytick={0.4, 0.45, 0.5, 0.55, 0.6, 0.65},
        legend cell align={left}
    ]

\addplot[color=blue, mark=square*] coordinates
{(1,0.5880681818181818) (2,0.59375) (3,0.6051136363636364) (4,0.59375) (5,0.6164772727272727) (6,0.5965909090909091) (7,0.6107954545454546) (8,0.6136363636363636) (9,0.6051136363636364) (10,0.6125356125356125) (11,0.5982905982905983) (12,0.6096866096866097) (13,0.6125356125356125) (14,0.6142857142857143) (15,0.6114285714285714) (16,0.62) (17,0.6189111747851003) (18,0.6109510086455331) (19,0.6115942028985507) (20,0.60932944606414) (21,0.6129032258064516) (22,0.6264705882352941) (23,0.6242603550295858) (24,0.6160714285714286) (25,0.6137724550898204) (26,0.6216216216216216) (27,0.6186186186186187) (28,0.6036036036036037) (29,0.6024096385542169) (30,0.608433734939759) (31,0.6114457831325302) (32,0.6097560975609756) (33,0.6123076923076923) (34,0.6018518518518519) (35,0.5987654320987654) (36,0.6006191950464397) (37,0.5975232198142415) (38,0.6006191950464397) (39,0.6) (40,0.6112852664576802)};
\addlegendentry{msprtGNN}
\addplot[color=red, mark=] coordinates {(1,0.44886363636363635) (2,0.4460227272727273) (3,0.4659090909090909) (4,0.4602272727272727) (5,0.48011363636363635) (6,0.4943181818181818) (7,0.4943181818181818) (8,0.5142045454545454) (9,0.5284090909090909) (10,0.50997150997151) (11,0.5071225071225072) (12,0.5384615384615384) (13,0.5441595441595442) (14,0.5342857142857143) (15,0.54) (16,0.5485714285714286) (17,0.5702005730659025) (18,0.5504322766570605) (19,0.5565217391304348) (20,0.5539358600583091) (21,0.5689149560117303) (22,0.5470588235294118) (23,0.5443786982248521) (24,0.5565476190476191) (25,0.5479041916167665) (26,0.5555555555555556) (27,0.5615615615615616) (28,0.5585585585585585) (29,0.5481927710843374) (30,0.5662650602409639) (31,0.5753012048192772) (32,0.5640243902439024) (33,0.5723076923076923) (34,0.5740740740740741) (35,0.5771604938271605) (36,0.5820433436532507) (37,0.5820433436532507) (38,0.5851393188854489) (39,0.578125) (40,0.5768025078369906)}; \addlegendentry{UPFD-Sage}
\addplot[color=green, mark=] coordinates {(1,0.4943181818181818) (2,0.48863636363636365) (3,0.4943181818181818) (4,0.5170454545454546) (5,0.5085227272727273) (6,0.5085227272727273) (7,0.5113636363636364) (8,0.5227272727272727) (9,0.5369318181818182) (10,0.5384615384615384) (11,0.5327635327635327) (12,0.5299145299145299) (13,0.5498575498575499) (14,0.5571428571428572) (15,0.56) (16,0.5571428571428572) (17,0.5472779369627507) (18,0.5619596541786743) (19,0.5681159420289855) (20,0.5714285714285714) (21,0.5630498533724341) (22,0.5794117647058824) (23,0.5769230769230769) (24,0.5833333333333334) (25,0.5958083832335329) (26,0.5855855855855856) (27,0.5885885885885885) (28,0.5945945945945946) (29,0.5963855421686747) (30,0.5963855421686747) (31,0.5963855421686747) (32,0.6036585365853658) (33,0.6061538461538462) (34,0.6018518518518519) (35,0.6018518518518519) (36,0.6099071207430341) (37,0.6130030959752322) (38,0.6130030959752322) (39,0.61875) (40,0.6144200626959248)}; \addlegendentry{GCNFN}
\addplot[color=orange, mark=star] coordinates
{(1,0.5738636363636364) (2,0.5909090909090909) (3,0.59375) (4,0.6079545454545454) (5,0.6022727272727273) (6,0.6022727272727273) (7,0.5909090909090909) (8,0.5994318181818182) (9,0.6051136363636364) (10,0.5982905982905983) (11,0.6011396011396012) (12,0.5954415954415955) (13,0.5897435897435898) (14,0.6) (15,0.6028571428571429) (16,0.6) (17,0.6045845272206304) (18,0.6051873198847262) (19,0.6086956521739131) (20,0.6064139941690962) (21,0.6041055718475073) (22,0.6088235294117647) (23,0.6094674556213018) (24,0.6041666666666666) (25,0.6017964071856288) (26,0.6006006006006006) (27,0.5945945945945946) (28,0.5915915915915916) (29,0.5903614457831325) (30,0.5903614457831325) (31,0.5933734939759037) (32,0.5884146341463414) (33,0.5815384615384616) (34,0.5771604938271605) (35,0.5771604938271605) (36,0.5758513931888545) (37,0.56656346749226) (38,0.56656346749226) (39,0.571875) (40,0.5705329153605015)};
\addlegendentry{Na\"ive}
\addplot[color=purple, mark=o] coordinates
{(1,0.5909090909090909) (2,0.6107954545454546) (3,0.5965909090909091) (4,0.6193181818181818) (5,0.6164772727272727) (6,0.6051136363636364) (7,0.6193181818181818) (8,0.6107954545454546) (9,0.6136363636363636) (10,0.5982905982905983) (11,0.5982905982905983) (12,0.6011396011396012) (13,0.6011396011396012) (14,0.6028571428571429) (15,0.6057142857142858) (16,0.6057142857142858) (17,0.6045845272206304) (18,0.6080691642651297) (19,0.6115942028985507) (20,0.6122448979591837) (21,0.5982404692082112) (22,0.5911764705882353) (23,0.591715976331361) (24,0.5922619047619048) (25,0.5958083832335329) (26,0.5885885885885885) (27,0.5915915915915916) (28,0.5885885885885885) (29,0.5933734939759037) (30,0.5843373493975904) (31,0.5783132530120482) (32,0.5853658536585366) (33,0.5969230769230769) (34,0.5895061728395061) (35,0.5833333333333334) (36,0.5820433436532507) (37,0.5820433436532507) (38,0.5820433436532507) (39,0.584375) (40,0.5830721003134797)};
\addlegendentry{MSPRT}
\end{axis}
\end{tikzpicture}
%\end{document}
         \vspace*{-5mm}
         \caption{3-class Weibo dataset}
         \label{fig:acc_time_weibo3_plot}
\end{subfigure}
\caption{Accuracy as a function of the decision deadline (number of tweets).}
\label{fig:acc_time_plot}
\end{figure}
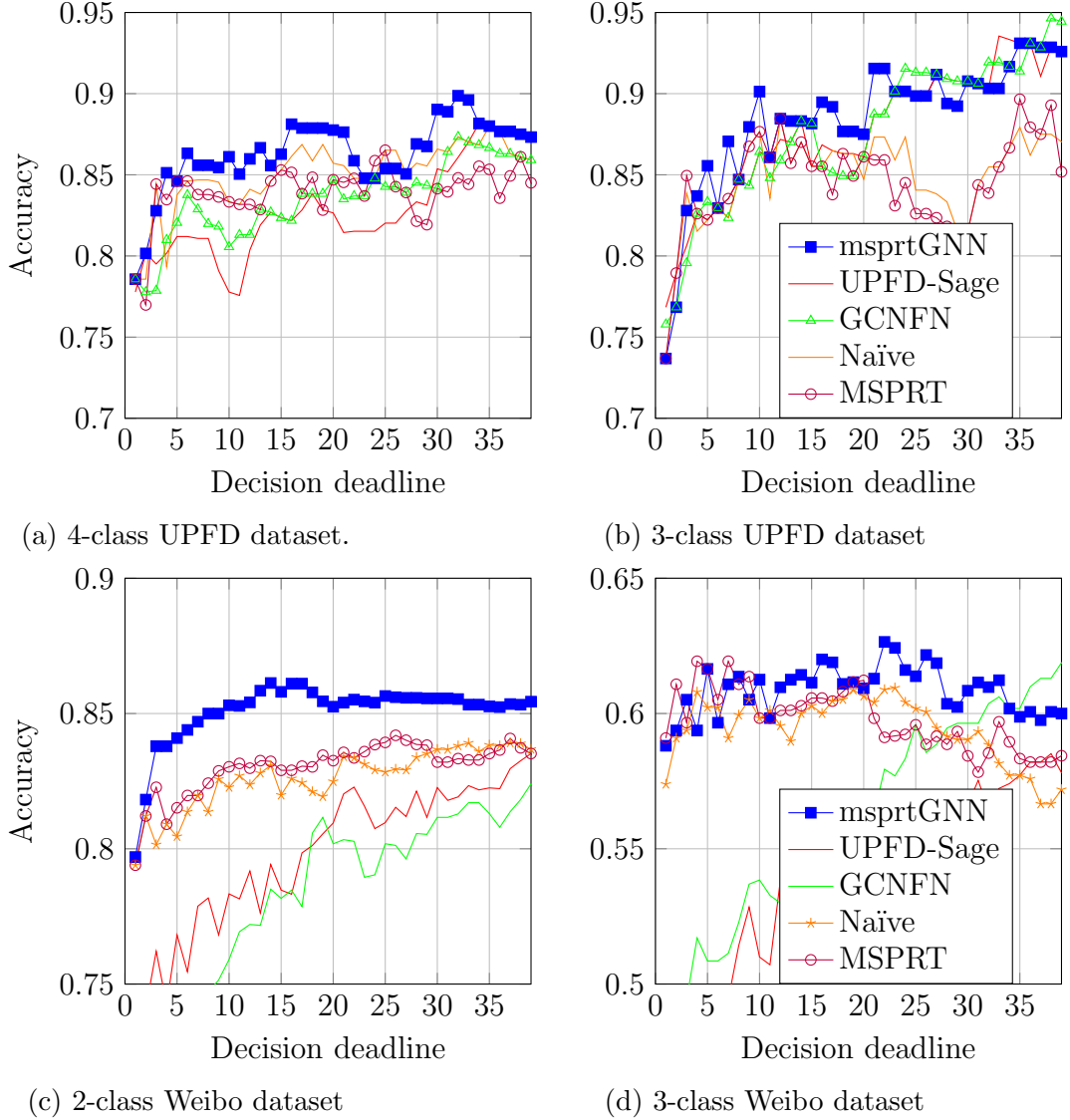

Next, Figure~\ref{fig:acc_time_plot} shows the accuracy achieved by the various algorithms, at different stopping times. Again, it can be seen that 
msprtGNN achieves the best performance as compared to the baselines, across all the scenarios we tested. This result is even more interesting given the fact the msprtGNN is not the best classifier in the non-sequential setting. Indeed, given a full information traces, UPFD-Sage outperforms slightly. Also, given single edges (i.e., at $t=1$), other baselines perform better as well. It is in the sequential setting where the msprtGNN is superior.

Finally, Table~\ref{tab:ablation1} shows our ablation study results. The purpose of this study is to show that msprtGNN \emph{outperforms} in the sequential setting, even though it underperforms in the non-sequential setting. Accordingly, this shows that the good performance of the msprtGNN algorithm in the sequential setting is not a result of being an overall better classifier. Instead, it is due to the way the architecture is designed for decreasing the test risk sequentially (see, \eqref{eqn:riskTest}). The first test we performed measures the accuracy when a decision is made after a single repost. In the second test we measure the accuracy when the decision is taken after seeing the whole information trace. In both tests we used the 4-class UPDF dataset with augmented ``content" node features. In both cases, msprtGNN is slightly inferior. 

\begin{table}[t!]
    \centering
    \begin{tabular}{|c|c|c|}
        \hline
        \textbf{Alg.} & \textbf{Accuracy (full trace)} & \textbf{Accuracy (single repost)}\\
        \hline
        \texttt{msprtGNN} & 0.95 & 0.91\\
        \hline
        \texttt{UPFD-sage} & \textbf{0.96} & \textbf{0.96} \\
        \hline
        \texttt{GCNFN} & \textbf{0.96} & 0.94\\
        \hline
        \texttt{HGFND} & 0.93 & - \\
        \hline
    \end{tabular}
    \caption{Accuracy in the 4-class UPFD dataset, with 310 node features.}
    \label{tab:ablation1}
\end{table}

\section{Proofs} \label{section:proofs}

In this section, we prove our main results.

\subsection{Proof of Theorem~\ref{th:bounded_stop_time}}

We begin by following the footsteps of \cite{340472}. We first note the simple fact that $\s{T}_{\s{MSPRT}}\leq\bar{\s{T}}$ with probability one, where
\begin{align}
\bar{\s{T}}\triangleq\inf\ppp{\ell\in\mathbb{N}:\frac{\pi_j}{\pi_k} \frac{f_j(\s{Z}_1^\ell)}{f_k(\s{Z}_1^\ell)}< \frac{\min_{\ell\in[M]} a_\ell}{M-1} \quad \forall j \neq k}.
\end{align}
Therefore, for any $k\in[M]$,
\begin{align}
\P(\s{T}_{\s{MSPRT}}>t\vert\calH_k)&\leq \P(\bar{\s{T}}>t\vert\calH_k)\\
&\leq \P\left(\left.\bigcup_{j: j \neq k} \frac{\pi_j}{\pi_k}\frac{f_j(\s{Z}_1^t)}{f_k(\s{Z}_1^t)} \geq \frac{\min_{\ell\in[M]} a_\ell}{M-1}\right|\calH_k\right) \\
&\leq \sum_{j: j \neq k}  \P\left(\left.\frac{\pi_j}{\pi_k}\frac{f_j(\s{Z}_1^t)}{f_k(\s{Z}_1^t)} \geq \frac{\min_{\ell\in[M]} a_\ell}{M-1}\right|\calH_k \right) \\ 
&= \sum_{j: j \neq k}  \P\left(\left.\sqrt{\frac{f_k(\s{Z}_1^t)}{f_j(\s{Z}_1^t)}} \geq \sqrt{\frac{\pi_k}{\pi_j}\frac{\min_{\ell\in[M]} a_\ell}{M-1}} \right|\calH_k\right)\\
&\leq \sum_{j: j \neq k} \sqrt{\frac{\pi_j}{\pi_k}\frac{M-1}{\min_{\ell\in[M]} a_\ell}} \bE\left[\left.\sqrt{ \frac{f_j(\s{Z}_1^t)}{f_k(\s{Z}_1^t)}}\right|\calH_k \right]\label{eqn:FirstUpperBoundTailTMSPRT0}\\
& = \sum_{j: j \neq k} \sqrt{\frac{\pi_j}{\pi_k}\frac{M-1}{\min_{\ell\in[M]} a_\ell}}[1-\calH^2\left(f_j(\s{Z}_1^t), f_k(\s{Z}_1^t) \right)]\\
& =  \sum_{j: j \neq k} \sqrt{\frac{\pi_j}{\pi_k}\frac{M-1}{\min_{\ell\in[M]} a_\ell}}\s{U}^{(t)}_{k,j}\label{eqn:FirstUpperBoundTailTMSPRT},
\end{align}
where the third inequality follows from the union bound, and the fourth inequality is due to Markov's inequality, and we have defined,
\begin{align}
\s{U}^{(t)}_{k,j} &\triangleq \bE\left[\left. \sqrt{\frac{f_j(\s{Z}_1^t)}{f_k(\s{Z}_1^t)}}\right|\calH_k \right].   
\end{align}
Next, for $z\in\calZ$ and $k\neq j\in[M]$, define the conditional Hellinger distance as
\begin{align}
    S_{k,j}(z)&\triangleq1-\calH^2(\alpha_k(\cdot \vert z), \alpha_j(\cdot\vert z))\\
    &= \sum_{z'\in\calZ}\sqrt{\alpha_k(z' \vert z)\alpha_j(z' \vert z)}.
\end{align}
Recall that if the $\ell$th observation $\s{Z}_\ell$ is the $j$th edge in path $\s{P}\in\calP_\ell$, we define the ancestor of $\s{Z}_\ell$ as $\calA_\ell\triangleq\s{Z}_{j-1}^{\s{P}}$, namely, it is the observation that precedes $\s{Z}_\ell$ in its corresponding path. Then, we note that
\begin{align}
\s{U}^{(t)}_{k,j} &= \bE\left[\left. \sqrt{\frac{f_j(\s{Z}_1^t)}{f_k(\s{Z}_1^t)}}\right|\calH_k \right]\\
&= \sum_{z_1^t} \prod_{i=1}^t\sqrt{\alpha_k(z_i \vert \calA_i)\alpha_j(z_i \vert \calA_i)}\\
&=\sum_{z_1^{t-1}} \sum_{z_t} \sqrt{\alpha_k(z_t \vert \calA_t)\alpha_j(z_t \vert \calA_t)} \prod_{i=1}^{t-1} \sqrt{\alpha_k(z_i \vert \calA_i)\alpha_j(z_i \vert \calA_i)}\\
&\leq \sum_{z_1^{t-1}} \prod_{i=1}^{t-1} \sqrt{\alpha_k(z_i \vert \calA_i)\alpha_j(z_i \vert \calA_i)} \cdot \max_{z\in\calZ} \sum_{z_t} \sqrt{\alpha_k(z_t \vert z)\alpha_j(z_t \vert z)} \\
&= \s{U}^{(t-1)}_{k,j}\cdot \max_{z\in\calZ} S_{k,j}(z).
\end{align}
Thus, applying the same chain of inequalities it is clear that,
\begin{align}
    \s{U}^{(t)}_{k,j} &\leq\pp{\max_{z\in\calZ}S_{k,j}(z)}^t.\label{eqn:ExponentialU}
\end{align}
Accordingly, we obtain that,
\begin{align}
\P(\s{T}_{\s{MSPRT}}>t\vert\calH_k)&\leq \sum_{j: j \neq k} \sqrt{\frac{\pi_j}{\pi_k}\frac{M-1}{\min_{\ell\in[M]} a_\ell}} \cdot\pp{\max_{z\in\calZ}S_{k,j}(z)}^t\\
&\leq (M-1)^{3/2}\max_{j\neq k}\sqrt{\frac{\pi_j}{\pi_k\min_{\ell\in[M]} a_\ell}}\pp{\max_{z\in\calZ}S_{k,j}(z)}^t\\
& = \s{C}_1\cdot\exp\p{-\s{C}_2\cdot t},\label{eqn:expofast}
%\P(\s{T}_{\s{MSPRT}}>t\vert\calH_k)&\leq \sum_{j: j \neq k} \sqrt{\frac{\pi_j}{\pi_k}\frac{M-1}{\min_{l} a_l}} \cdot\max_{\calF_{t-1}}\pp{\max_{\ell}S_\ell}^t\\
%&\leq (M-1)^{3/2}\max_{j\neq k}\sqrt{\frac{\pi_j}{\pi_k\min_{l} a_l}}\max_{\s{Z}_{u_t}}\pp{\max_{\ell}S_\ell}^t.\label{eqn:expofast}
\end{align}
where $\s{C}_1\triangleq(M-1)^{3/2}\max_{j\neq k}\sqrt{\frac{\pi_j}{\pi_k\min_{\ell\in[M]} a_\ell}}$ and $\s{C}_2\triangleq \max_{j\neq k}\max_{z\in\calZ}\log\frac{1}{S_{k,j}(z)}$. Since we assume that for any $k\neq j$ and any $z\in\calZ$, the transition probability distributions $\alpha_k(\cdot\vert z)$ and $\alpha_j(\cdot\vert z)$ are not the same, it follows that $\max_{k\neq j}\max_{z\in\calZ}S_{k,j}(z)<1\label{eqn:bounded_hellinger}$, which implies that $\s{C}_2>0$. Thus, the right-hand-side of \eqref{eqn:expofast} decays exponentially fast with $t$, which concludes the proof.

\subsection{Proof of Theorem~\ref{th:asymptotic_time}}
To prove Theorem~\ref{th:asymptotic_time} we need a few auxiliary results. Recall that for $m\in[M]$ and $\ell\in\mathbb{N}$ we define $ f_m(\s{Z}_1^\ell) \triangleq \P(\s{Z}_1^\ell \vert \calH_m)$, and that $\pi^{\s{stat}}_k$ is the stationary distribution of the $k$th irreducible Markov chain with transition probabilities $\alpha_k(\cdot\vert\cdot)$. We start with the following result which proves an asymptotic equipartition property (AEP) of $\frac{1}{\ell}\log\frac{f_k(\s{Z}_1^l)}{f_j(\s{Z}_1^l)}$, as $\ell\to\infty$. 

\begin{lemma}[AEP for Markov edges]\label{lem:AEP}
Let $\s{Z}_1^\ell$ be a sequence of edge types sampled from $\P(\cdot\vert\calH_k)$. %Let us assume that the length of each disjoint path $P\in\calP_\ell$, tends to $\infty$ as $\ell\to\infty$. 
Then, $f_k$-almost surely,
\begin{align}
\frac{1}{\ell}\log\frac{f_k(\s{Z}_1^\ell)}{f_j(\s{Z}_1^\ell)} \xrightarrow[\ell \to \infty]{} d_{\s{KL}}(\alpha_k||\alpha_j\vert\pi_k^{\s{stat}}).
\end{align}
\end{lemma}
\begin{proof}[Proof of Lemma~\ref{lem:AEP}]
Let $\mathscr{N}_{z}$ be the number of edges in $\s{Z}_1^\ell$ whose parent edge is $z$, for $z\in\calZ$. We have,
\begin{align}
    \lim_{\ell\to\infty}\frac{\mathscr{N}_z}{\ell}
    &=\lim_{\ell\to\infty}\frac{\sum_{\s{P}\in\calP_\ell}\sum_{i: \s{Z}_{i-1}^P=z} 1}{\ell}\\    &=\lim_{\ell\to\infty}\frac{\sum_{\s{P}\in\calP_\ell}\left[|\s{P}|\frac{1}{|\s{P}|}\sum_{i: \s{Z}_{i-1}^P=z} 1\right]}{\ell}\\
    &=\lim_{\ell\to\infty}\sum_{\s{P}\in\calP_\ell}\left[\frac{|\s{P}|}{\ell}\cdot\frac{1}{|\s{P}|}\sum_{i: \s{Z}_{i-1}^P=z} 1\right]\\
    &=\lim_{\ell\to\infty}\sum_{\s{P}\in\calP_\ell} w_{\s{P}} \cdot \calS_{\s{P},z},\label{eqn:Cesaro}
\end{align}
for any $z\in\calZ$, where we have defined $w_{\s{P}} \triangleq |\s{P}|/\ell$, and $\calS_{\s{P},z}\triangleq\frac{1}{|\s{P}|}\sum_{i: \s{Z}_{i-1}^P=z} 1$. Now, because we assume that $\alpha_k$ is irreducible, by the ergodic theorem for Markov chains \cite[Theorem 1.10.2]{Norris_1997}, we have that $\calS_{\s{P},z}\to\pi^{\s{stat}}_k(z)$ almost surely, for any $\s{P}\in\calP_\ell$, as $\ell\to\infty$, and for any $z\in\calZ$. Thus, since the sequence of random variables $\{\calS_{\s{P},z}\}_{\s{P}}$ converges to a limit almost-surely, then their weighted Cesaro-mean in \eqref{eqn:Cesaro} converges, almost surely, to the same limit as well \cite{bibaut2020sufficientinsufficientconditionsstochastic}. Thus, we finally get that,
\begin{align}
    \lim_{\ell\to\infty}\frac{\mathscr{N}_z}{\ell}
    &=\pi^{\s{stat}}_k(z).
\end{align}
Using the above result, we have,
\begin{align}
\frac{1}{\ell}\log\frac{f_k(\s{Z}_1^\ell)}{f_j(\s{Z}_1^\ell)}
&=\frac{1}{\ell}\sum_{i=1}^\ell\log{\frac{\alpha_k(\s{Z}_{i}\vert\calA_i)}{\alpha_j(\s{Z}_{i}\vert\calA_i)}}\\ 
&= \frac{1}{\ell}\sum_{z \in \calZ}\sum_{i: \calA_i=z}\log{\frac{\alpha_k(\s{Z}_{i}\vert\calA_i)}{\alpha_j(\s{Z}_{i}\vert\calA_i)}}\\
&= \sum_{z \in \calZ} \frac{\mathscr{N}_{z}}{\ell}\cdot\left(\frac{1}{\mathscr{N}_{z}}\sum_{i:\calA_i=z} \log{\frac{\alpha_k(\s{Z}_{i}\vert\calA_i)}{\alpha_j(\s{Z}_{i}\vert\calA_i)}}\right)\\
&\xrightarrow[\ell \to \infty]{} \sum_{z \in \calZ}\pi^{\s{stat}}_k(z)d_{\s{KL}}\left(\alpha_k(\cdot\vert z)||\alpha_j(\cdot\vert z)\right)\\
&=d_{\s{KL}}(\alpha_k||\alpha_j\vert\pi_k^{\s{stat}}),
\end{align}
almost surely, where we have applied the vanilla AEP \cite[Theorem 11.8.1]{10.5555/1146355} on $\{\s{Z}_j\}_{j\geq1}$, which is an i.i.d. sequence when conditioned on their parent edge.
\end{proof}

\begin{lemma}\label{lem:lemma_time}
Fix $k\in[M]$. Recall the definition of $\s{T}_{\s{MSPRT}}$ in \eqref{eqn:decproblem}, and assume that
\begin{align}
%\exists j\in[\s{M}]: d_{\s{KL}}(\alpha_k(\cdot\vert z)||\alpha_{j_k^\star}(\cdot\vert z)) < \infty \quad\forall z\in\calZ
\min_{j\neq k\in[M]}\max_{z\in\calZ} \chi^2(\alpha_k(\cdot\vert z),\alpha_{j}(\cdot\vert z)) < \infty\label{eqn:finite_stat_kl_div}.
\end{align}
%and
%\begin{align}
%\max_{z\in\calZ}\max_{k\neq j\in[M]}S_{k,j}(z)<1\label{eqn:hellinger_distance_time}.
%\end{align}
Then,
\begin{align}
    \s{T}_{\s{MSPRT}}\to\infty,
\end{align}
$f_k$-almost surely, as $\norm{a}_\infty \to 0$.
\end{lemma}

\begin{proof}[Proof of Lemma~\ref{lem:lemma_time}]
Fix $n\in\mathbb{N}$, and let $j_k^\star$ be a hypothesis $j\in[\s{M}]$ which achieves the minimum in \eqref{eqn:finite_stat_kl_div}. Using the definition of $\s{T}_{\s{MSPRT}}$, we have,
\begin{align}
\P(\s{T}_{\s{MSPRT}}<n \vert \calH_k) &= \P \left(\left.\exists \ell\in[M]: \max_{1 \leq m \leq n} \frac{\pi_\ell f_\ell(\s{Z}_1^m)}{\sum_j \pi_j f_j(\s{Z}_1^m)} > \frac{1}{1+a_\ell} \right|\calH_k\right)\\
&=\P \left(\left.\exists \ell\in[M]: \min_{1 \leq m \leq n} \frac{\sum_j \pi_j f_j(\s{Z}_1^m)}{\pi_\ell f_\ell(\s{Z}_1^m)} < 1+a_\ell\right|\calH_k\right)\\
&=\P \left(\left.\exists \ell\in[M]: \min_{1 \leq m \leq n} \sum_{j: j \neq l} \pi_j \frac{f_j(\s{Z}_1^m)}{f_\ell(\s{Z}_1^m)} < \pi_\ell a_\ell\right|\calH_k\right)\\
&\leq \P \left(\left.\exists \ell\in[M]: \min_{1 \leq m \leq n} \frac{f_j(\s{Z}_1^m)}{f_\ell(\s{Z}_1^m)} < \frac{\pi_\ell a_\ell}{\pi_j},\;\forall j \neq \ell\right|\calH_k\right)\\
& = \P \left(\left.\bigcup_{\ell\in[M]}: \min_{1 \leq m \leq n} \frac{f_j(\s{Z}_1^m)}{f_\ell(\s{Z}_1^m)} < \frac{\pi_\ell a_\ell}{\pi_j},\;\forall j \neq \ell\right|\calH_k\right)\\
&\leq\sum_{\ell\in[M]} \P \left(\left.\min_{1 \leq m \leq n} \frac{f_j(\s{Z}_1^m)}{f_\ell(\s{Z}_1^m)} < \frac{\pi_\ell a_\ell}{\pi_j},\;\forall j \neq \ell\right|\calH_k\right)\\
&=\sum_{\ell\neq k} \P \left(\left.\min_{1 \leq m \leq n} \frac{f_j(\s{Z}_1^m)}{f_\ell(\s{Z}_1^m)} < \frac{\pi_\ell a_\ell}{\pi_j},\;\forall j \neq \ell\right|\calH_k\right)\nonumber\\
&\quad\quad\quad\quad+\P \left(\left.\min_{1 \leq m \leq n} \frac{f_j(\s{Z}_1^m)}{f_k(\s{Z}_1^m)} < \frac{\pi_k a_k}{\pi_j},\;\forall j \neq k\right|\calH_k\right)\label{eqn:beforejkstar}\\
&\leq \sum_{\ell \neq k} \P \left(\left.\max_{1 \leq m \leq n} \frac{f_\ell(\s{Z}_1^m)}{f_k(\s{Z}_1^m)} > \frac{\pi_k}{\pi_\ell a_\ell} \right|\calH_k\right)\nonumber\\
&\quad\quad\quad\quad+ \P \left(\left.\max_{1 \leq m \leq n} \frac{f_k(\s{Z}_1^m)}{f_{j_k^\star}(\s{Z}_1^m)} > \frac{\pi_{j_k^\star}}{\pi_k a_k} \right|\calH_k\right)\label{eqn:unionDouble},
\end{align}
where the first inequality is because the sum of positive terms is larger than only a single term in the sum, the second inequality follows from the union bound, and the last inequality is by replacing the intersection over all $j \neq \ell$, with a single index $j=k$ in the first term at the right-hand-side of \eqref{eqn:beforejkstar}, and with $j=j_k^\star$ in the second term at the right-hand-side of \eqref{eqn:beforejkstar}. Before proving almost-sure convergence, we will prove convergence in probability. Specifically, applying  Markov's inequality on the first term at the right-hand-side of \eqref{eqn:unionDouble}, we get for any $\ell\neq k$,
\begin{align}
\P \left(\left.\max_{1 \leq m \leq n}\frac{f_\ell(\s{Z}_1^m)}{f_k(\s{Z}_1^m)} > \frac{\pi_k}{\pi_\ell a_\ell} \right|\calH_k\right)
&\leq \sum_{m=1}^n\frac{\pi_\ell a_\ell}{\pi_k}\E\pp{\left.\frac{f_\ell(\s{Z}_1^m)}{f_k(\s{Z}_1^m)}\right|\calH_k}\\
& =n\frac{\pi_\ell a_\ell}{\pi_k},
\end{align}
where the last equality is due to the fact that
\begin{align}
\E\pp{\left.\frac{f_\ell(\s{Z}_1^m)}{f_k(\s{Z}_1^m)}\right|\calH_k} = \sum_{z_1^m\in\calZ^m}\frac{f_\ell(z_1^m)}{f_k(z_1^m)}f_k(z_1^m) =1.
\end{align}
Thus, the first term at the right-hand-side of \eqref{eqn:unionDouble} can be upper bounded as,
\begin{align}
    \sum_{\ell \neq k} \P \left(\left.\max_{1 \leq m \leq n}\frac{f_\ell(\s{Z}_1^m)}{f_k(\s{Z}_1^m)} > \frac{\pi_k}{\pi_\ell a_\ell} \right|\calH_k\right)\leq \p{\frac{n}{\pi_k}\sum_{\ell\neq k}\pi_k}\cdot\norm{a}_\infty,\label{eqn:secConvA2}
\end{align}
which goes to zero, as $\norm{a}_\infty\to0$. 

\begin{comment}
\begin{align}
\P \left(\left.\min_{1 \leq m \leq n} \log\frac{f_k(\s{Z}_1^m)}{f_\ell(\s{Z}_1^m)} < \log\frac{\pi_\ell a_\ell}{\pi_k} \right|\calH_k\right)  &= \P \left(\left.\max_{1 \leq m \leq n} \log\frac{f_\ell(\s{Z}_1^m)}{f_k(\s{Z}_1^m)} > \log\frac{\pi_k}{\pi_\ell a_\ell} \right|\calH_k\right) \\
&\leq \sum_{m=1}^n\P \left(\left. \log\frac{f_\ell(\s{Z}_1^m)}{f_k(\s{Z}_1^m)} > \log\frac{\pi_k}{\pi_\ell a_\ell} \right|\calH_k\right)\\
&\leq \sum_{m=1}^n \inf_{t>0}e^{-t \log\frac{\pi_k}{\pi_\ell a_\ell}}\E{\left.e^{t\log\frac{f_\ell(\s{Z}_1^m)}{f_k(\s{Z}_1^m)}}\right|\calH_k}\\
&=\sum_{m=1}^n \inf_{t>0}\left(\frac{\pi_\ell a_\ell}{\pi_k}\right)^t \E{\left. \left(\frac{f_\ell(\s{Z}_1^m)}{f_k(\s{Z}_1^m)}\right)^t \right|\calH_k}\\
&\leq \sum_{m=1}^n \sqrt{\frac{\pi_\ell a_\ell}{\pi_k}} \E{\left. \sqrt{\frac{f_\ell(\s{Z}_1^m)}{f_k(\s{Z}_1^m)}} \right|\calH_k}\\
&= \sqrt{\frac{\pi_\ell a_\ell}{\pi_k}} \sum_{m=1}^n \left(1-\calH^2(f_\ell(\s{Z}_1^m), f_k(\s{Z}_1^m))\right)\\
&\leq \sqrt{\frac{\pi_\ell a_\ell}{\pi_k}} \sum_{m=1}^n\left(1-\calH^2(\alpha_k(\cdot \vert z))\right)^m \xrightarrow[\norm{a}_\infty]{} 0.
\end{align}
The first inequality is a union bound. The second inequality - chernoff bound. In the third inequality we choose specifically t=$0.5$. In the last inequality we used the asuumption in equation \eqref{eqn:hellinger_distance_time}. 
\end{comment}

Next, let us analyze the second term in \eqref{eqn:unionDouble}. %We will use here the assumption in equation \eqref{eqn:finite_stat_kl_div}. Specifically, it is not difficult to see that \eqref{eqn:finite_stat_kl_div} combined with the fact that $\calZ$ is finite, implies that $\frac{\alpha_k(z\vert z')}{\alpha_{j_k^\star}(z\vert z')}\leq\s{C}$, is bounded for any $z,z'\in\calZ$, for some $\s{C}>0$. 
Using Markov's inequality again, we have,
\begin{align}
\P \left(\left.\max_{1 \leq m \leq n} \frac{f_k(\s{Z}_1^m)}{f_{j_k^\star}(\s{Z}_1^m)} > \frac{\pi_{j_k^\star}}{\pi_k a_k} \right|\calH_k\right)&\leq\sum_{m=1}^n\P{\left. \frac{f_k(\s{Z}_1^m)}{f_{j_k^\star}(\s{Z}_1^m)} > \frac{\pi_{j_k^\star}}{\pi_k a_k} \right|\calH_k}\\
&\leq \sum_{m=1}^n\frac{\pi_ka_k}{\pi_{j_k^\star}}\E\pp{\left.\frac{f_k(\s{Z}_1^m)}{f_{j_k^\star}(\s{Z}_1^m)}\right|\calH_k}\\
&= \sum_{m=1}^n\frac{\pi_ka_k}{\pi_{j_k^\star}}\pp{1+\chi^2(f_k(\s{Z}_1^m),f_{j_k^\star}(\s{Z}_1^m))}.
%&=\sum_{m=1}^n\P{\left.\sum_{i=1}^m\log\frac{\alpha_k(\s{Z}_i\vert\calA_i)}{\alpha_{j_k^\star}(\s{Z}_i\vert\calA_i)} > \log\frac{\pi_k a_k}{\pi_{j_k^\star}} \right|\calH_k}.
\end{align}
Next, we note that,
\begin{align}
    1+\chi^2(f_k({\s{Z}}_1^m),f_{j_k^\star}({\s{Z}}_1^m)) &= \sum_{z_1^m\in{\calZ}^m}\frac{f_k^2(z_1^m)}{f_{j_k^\star}(z_1^m)}\\
    & = \sum_{z_1^{m-1}\in{\calZ}^{m-1}}\prod_{i=1}^{m-1}\frac{\alpha_k^2(z_i\vert\calA_i)}{\alpha_{j_k^\star}(z_i\vert \calA_i)}\sum_{z_m\in{\calZ}}\frac{\alpha_k^2(z_m\vert\calA_m)}{\alpha_{j_k^\star}(z_m\vert \calA_m)}\\
    &\leq \sum_{z_1^{m-1}\in{\calZ}^{m-1}}\prod_{i=1}^{m-1}\frac{\alpha_k^2(z_i\vert\calA_i)}{\alpha_{j_k^\star}(z_i\vert \calA_i)}\max_{z'\in\calZ}\sum_{z_m\in{\calZ}}\frac{\alpha_k^2(z_m\vert z')}{\alpha_{j_k^\star}(z_m\vert z')}\\
    & = \sum_{z_1^{m-1}\in{\calZ}^{m-1}}\prod_{i=1}^{m-1}\frac{\alpha_k^2(z_i\vert\calA_i)}{\alpha_{j_k^\star}(z_i\vert \calA_i)}\cdot\max_{z'\in\calZ}\chi^2(\alpha_k(\cdot\vert z'),\alpha_{j_k^\star}(\cdot\vert z'))\\
    &\leq\cdots\\
    &\leq\pp{\max_{z'\in\calZ}\chi^2(\alpha_k(\cdot\vert z'),\alpha_{j_k^\star}(\cdot\vert z'))}^m\\
    &\leq \s{C}^m,
\end{align}
where the last inequality follows from the assumption in \eqref{eqn:finite_stat_kl_div}, which implies that $\max_{z'\in\calZ}\chi^2(\alpha_k(\cdot\vert z'),\alpha_{j_k^\star}(\cdot\vert z'))\leq\s{C}$, for some $\s{C}>0$. Therefore,
\begin{align}
    \P \left(\left.\max_{1 \leq m \leq n} \frac{f_k(\s{Z}_1^m)}{f_{j_k^\star}(\s{Z}_1^m)} > \frac{\pi_{j_k^\star}}{\pi_k a_k} \right|\calH_k\right)&\leq \p{\frac{\pi_k}{\pi_{j_k^\star}}\sum_{m=1}^n\s{C}^m}a_k\\
    &\leq \p{\frac{\pi_k}{\pi_{j_k^\star}}\sum_{m=1}^n\s{C}^m}\norm{a}_\infty,\label{eqn:secConvA3}
\end{align}
which goes to zero, as $\norm{a}_\infty\to0$. Therefore, combining \eqref{eqn:unionDouble}, \eqref{eqn:secConvA2}, and \eqref{eqn:secConvA3}, we obtain that $\s{T}_{\s{MSPRT}}\to\infty$ in probability as $\norm{a}_\infty\to0$. Now, convergence in probability implies that there must be a sub-sequence of $\s{T}_{\s{MSPRT}}$ which converge to infinity $f_k$-almost surely. Because $\s{T}_{\s{MSPRT}}$ is non-decreasing as each $a_\ell\to0$, we may conclude that $\s{T}_{\s{MSPRT}}\to\infty$ $f_k$-almost surely, as $\norm{a}_\infty\to0$.

%\begin{align}    
%&\leq\sum_{m=1}^n\P{\left. \left|\frac{1}{m}\sum_{i=1}^m\log\frac{\alpha_k(\s{Z}_i\vert\calA_i)}{\alpha_{j_k^\star}(\s{Z}_i\vert\calA_i)}\right| > \left|\frac{1}{m}\log\frac{\pi_k a_k}{\pi_{j_k^\star}}\right| \right|\calH_k}\\
%&\leq\sum_{m=1}^n\frac{\E{\left. \left|\frac{1}{m}\sum_{i=1}^m\log\frac{\alpha_k(\s{Z}_i\vert\calA_i)}{\alpha_{j_k^\star}(\s{Z}_i\vert\calA_i)}\right|\right|\calH_k}}{\left|\frac{1}{m}\log\frac{\pi_k a_k}{\pi_{j_k^\star}}\right|} \to 0\\
%&\leq\sum_{m=1}^n\frac{\E{\left. \frac{1}{m}\sum_{i=1}^m\left|\log\frac{\alpha_k(\s{Z}_i\vert\calA_i)}{\alpha_{j_k^\star}(\s{Z}_i\vert\calA_i)}\right|\right|\calH_k}}{\left|\frac{1}{m}\log\frac{\pi_k a_k}{\pi_{j_k^\star}}\right|} \to 0.
%\end{align}
\end{proof}
%Regarding the assumption given in equation \eqref{eqn:finite_stat_kl_div}. while it is desirable to have infinite kl-divergence between the likelihoods of hypotheses, it is unhelpful for asymptotic analysis, as it implies there is a sequence $\s{Z}_1^m$ for which $\P(\calH_{j_k^\star} \vert \s{Z}_1^m, \calH_k)=0$. If this sequence is observed, $\calH_{j_k^\star}$ can be dropped, and the stopping time would not increase asymptotically as $\norm{a}_\infty\to0$. Without this assumption, Theorem~\ref{th:asymptotic_time} serves as an asymptotic upper bound to the stopping time.

\begin{lemma}\label{lem:lemma_time_k}
Fix $k\in[M]$, and let
\begin{align}
    \s{T}_k \triangleq \inf\ppp{n\in\mathbb{N}:\frac{\pi_k f_k(\s{Z}_1^n)}{\sum_{j=0}^{M-1} \pi_{j} f_j(\s{Z}_1^n)} > \frac{1}{1+a_k}}.
\end{align}
Assume that
\begin{align}
%\exists j\in[\s{M}]: d_{\s{KL}}(\alpha_k(\cdot\vert z)||\alpha_{j_k^\star}(\cdot\vert z)) < \infty \quad\forall z\in\calZ
\min_{j\neq k\in[M]}\max_{z\in\calZ} \chi^2(\alpha_k(\cdot\vert z),\alpha_{j}(\cdot\vert z)) < \infty\label{eqn:finite_stat_kl_div_k}.
\end{align}
Then,
\begin{align}
    \lim_{\norm{a}_\infty \to 0} \frac{\s{T}_k}{- \log a_k} = \frac{1}{\min_{j:j \neq k} d_{\s{KL}}(\alpha_k||\alpha_j\vert\pi_k^{\s{stat}})},
\end{align}
$f_k$-almost surely. 
\end{lemma}

\begin{proof}[Proof of Lemma~\ref{lem:lemma_time_k}]
For simplicity of notation we define $p_k^{(n)} \triangleq \frac{\pi_k f_k(\s{Z}_1^n)}{\sum_{j=0}^{M-1} \pi_{j} f_j(\s{Z}_1^n)}$. Note that $\s{T}_k \geq \s{T}_{\s{MSPRT}}$, because $\s{T}_{\s{MSPRT}} = \min_{j\in[M]} \s{T}_j$. Therefore, Lemma~\ref{lem:lemma_time} implies that $\s{T}_k \to\infty$, as $\norm{a}_\infty \to 0$, for any $k\in[M]$. A little bit of straightforward algebra steps reveal that $\s{T}_k$ can be represented as,
\begin{align}
    \s{T}_k \triangleq \inf\ppp{n\in\mathbb{N}:\mathscr{W}_n(\s{Z}_1^n)> -\frac{\log a_k}{n}},
\end{align}
where
\begin{align}
    \mathscr{W}_n(\s{Z}_1^n)\triangleq-\frac{1}{n}\log{\sum_{j:j \neq k} \exp(-n\left[\frac{1}{n}\log\frac{\pi_k}{\pi_j} + \frac{1}{n}\log\frac{f_k(\s{Z}_1^n)}{f_j(\s{Z}_1^n)}\right])}.
\end{align}
Then, at $n = \s{T}_k$, we note that,
\begin{align}
\lim_{\norm{a}_\infty \to 0}\mathscr{W}_{\s{T}_k}(\s{Z}_1^{\s{T}_k})& = \lim_{\s{T}_k\to\infty}\mathscr{W}_{\s{T}_k}(\s{Z}_1^{\s{T}_k})\\
&=\lim_{\s{T}_k\to\infty}-\frac{1}{\s{T}_k}\log{\sum_{j: j \neq k} \exp(-\s{T}_k\left[\frac{1}{\s{T}_k}\log\frac{\pi_k}{\pi_j} + \frac{1}{\s{T}_k}\log\frac{f_k(\s{Z}_1^{\s{T}_k})}{f_j(\s{Z}_1^{\s{T}_k})}\right])}\\
&=\lim_{\s{T}_k\to\infty}\min_{j:j \neq k}\pp{\frac{1}{\s{T}_k}\log\frac{\pi_k}{\pi_j} + \frac{1}{\s{T}_k}\log\frac{f_k(\s{Z}_1^{\s{T}_k})}{f_j(\s{Z}_1^{\s{T}_k})}}\\
&=\min_{j:j \neq k} d_{\s{KL}}(\alpha_k||\alpha_j\vert\pi_k^{\s{stat}}),\label{eqn:KLabove}
\end{align}
almost surely, where the first equality is because of Lemma~\ref{lem:lemma_time}. To apply Lemma~\ref{lem:lemma_time} we used the assumption in \eqref{eqn:finite_stat_kl_div_k}. The third equality follows from \cite[Lemma 5.2]{340472}. The last equality is due to Lemma~\ref{lem:AEP}. Next, at $n=\s{T}_k-1$ we by continuity, we have,
\begin{align}
\lim_{\norm{a}_\infty \to 0}\mathscr{W}_{\s{T}_k-1}(\s{Z}_1^{\s{T}_k-1})& = \lim_{\s{T}_k\to\infty}\mathscr{W}_{\s{T}_k-1}(\s{Z}_1^{\s{T}_k-1})\\
&=\lim_{\s{T}_k\to\infty}-\frac{1}{\s{T}_k}\log{\sum_{j: j \neq k} \exp(-\s{T}_k\left[\frac{1}{\s{T}_k}\log\frac{\pi_k}{\pi_j} + \frac{1}{\s{T}_k}\log\frac{f_k(\s{Z}_1^{\s{T}_k-1})}{f_j(\s{Z}_1^{\s{T}_k-1})}\right])}\\
&=\lim_{\s{T}_k\to\infty}\min_{j:j \neq k}\pp{\frac{1}{\s{T}_k}\log\frac{\pi_k}{\pi_j} + \frac{1}{\s{T}_k}\log\frac{f_k(\s{Z}_1^{\s{T}_k-1})}{f_j(\s{Z}_1^{\s{T}_k-1})}}\\
&=\min_{j:j \neq k} d_{\s{KL}}(\alpha_k||\alpha_j\vert\pi_k^{\s{stat}}),\label{eqn:KLbelow}
\end{align}
almost surely. Therefore, since $\s{T}_k$ is defined as the first time $n$ for which $p_k^{(n)} > \frac{1}{1+a_k}$, we have the following two inequalities simultaneously,
\begin{align}
\mathscr{W}_{\s{T}_k}(\s{Z}_1^{\s{T}_k}) &\geq -\frac{\log a_k}{\s{T}_k},\label{eqn:asymp1}\\
\mathscr{W}_{\s{T}_k-1}(\s{Z}_1^{\s{T}_k-1}) &\leq -\frac{\log a_k}{\s{T}_k-1}.\label{eqn:asymp2}
\end{align}
Applying the squeeze theorem on \eqref{eqn:asymp1}--\eqref{eqn:asymp2}, and using \eqref{eqn:KLabove} and \eqref{eqn:KLbelow}, we finally obtain that,
\begin{align}
    \frac{\s{T}_k}{- \log a_k} \to \frac{1}{\min_{j:j \neq k} d_{\s{KL}}(\alpha_k||\alpha_j\vert\pi_k^{\s{stat}})},
\end{align}
almost surely w.r.t. $f_k$, as $\norm{a}_\infty \to 0$, which concludes the proof.
\end{proof}

We are now in a position to prove Theorem \ref{th:asymptotic_time}. Specifically, this theorem guarantees both almost-sure in expectation types of convergence. We begin with the former. To that end, we note that for all $\epsilon>0$,
\begin{align}
&\P\pp{\left.\left|\frac{\s{T}_{\s{MSPRT}}}{-\log a_k} - d_{\s{KL}}(\alpha_k||\alpha_j\vert\pi_k^{\s{stat}})\right| > \epsilon\right|\calH_k}=\nonumber\\
%&= \sum_{l=0}^{\s{M}-1} \P{\left|\frac{\s{T}_{\s{MSPRT}}}{-\log a_k} - d_{\s{KL}}(\alpha_k||\alpha_j\vert\pi_k^{\s{stat}})\right| > \epsilon \text{ and accept } \calH_l}\\
&\quad\quad=\P{\left.\left|\frac{\s{T}_k}{-\log a_k} - d_{\s{KL}}(\alpha_k||\alpha_j\vert\pi_k^{\s{stat}})\right| > \epsilon\right|\calH_k}\nonumber\\
&\quad\quad\quad\quad+ \sum_{\ell\neq k} \P{\left.\left|\frac{\s{T}_{\s{MSPRT}}}{-\log a_k} - d_{\s{KL}}(\alpha_k||\alpha_j\vert\pi_k^{\s{stat}})\right| > \epsilon,\; \s{accept}\;\calH_\ell\right|\calH_k}\\
&\quad\quad\leq \P{\left.\left|\frac{\s{T}_k}{-\log a_k} - d_{\s{KL}}(\alpha_k||\alpha_j\vert\pi_k^{\s{stat}})\right| > \epsilon\right|\calH_k} + \sum_{\ell\neq k} \P{\s{accept}\;\calH_\ell\vert\calH_k}.\label{eqn:inprobab}
\end{align}
Theorem~\ref{th:error_guarantee} implies that $\P{\s{accept}\;\calH_\ell\vert\calH_k}\to0$, as $\norm{a}_\infty\to0$, while Lemma~\ref{lem:lemma_time_k} implies that the first term at the right-hand-side of \eqref{eqn:inprobab} converges to zero, as $\norm{a}_\infty\to0$; this proves the convergence in probability of $\frac{\s{T}_{\s{MSPRT}}}{-\log a_k}$ to $d_{\s{KL}}(\alpha_k||\alpha_j\vert\pi_k^{\s{stat}})$. Now, using the same arguments as in the proof of Lemma~\ref{lem:lemma_time_k}, since $\s{T}_{\s{MSPRT}}$ is non-decreasing, as $\norm{a}_\infty\to0$, we also have $\frac{\s{T}_{\s{MSPRT}}}{-\log a_k}$ converges $f_k$-almost surely to $d_{\s{KL}}(\alpha_k||\alpha_j\vert\pi_k^{\s{stat}})$.

As for convergence in expectation, we will prove that $\frac{\s{T}_{\s{MSPRT}}}{-\log a_k}$ is uniformly integrable, and then the almost sure convergence will imply the convergence in expectation \cite[Theorem 16.4]{Bill86}. We have,
\begin{align}
    \E [\s{T}_{\s{MSPRT}} | \s{T}_{\s{MSPRT}} \geq \ell,\calH_k] &= \sum_{t=\ell}^{\infty} {t\cdot\P(\s{T}_{\s{MSPRT}}=t\vert\calH_k)}\\
    &\leq \sum_{t=\ell}^{\infty} {t\cdot\P(\s{T}_{\s{MSPRT}} \geq t\vert\calH_k)}\\
    &\leq \sum_{t=\ell}^{\infty} t \sum_{j: j \neq k} \sqrt{\frac{\pi_j}{\pi_k}\frac{M-1}{\min_{\ell\in[M]} a_\ell}} \cdot\s{U}^{(t)}_{k,j} \\
    &= \sum_{j: j \neq k} \sqrt{\frac{\pi_j}{\pi_k}\frac{M-1}{\min_{\ell\in[M]} a_\ell}} \sum_{t=\ell}^{\infty} t\cdot\s{U}^{(t)}_{k,j}\\
    &\leq \sum_{j: j \neq k} \sqrt{\frac{\pi_j}{\pi_k}\frac{M-1}{\min_{\ell\in[M]} a_\ell}} \sum_{t=\ell}^{\infty} t\pp{\max_{z\in\calZ}S_{k,j}(z)}^t,
\end{align}
where the second inequality follows from \eqref{eqn:FirstUpperBoundTailTMSPRT}, and the last inequality follows from \eqref{eqn:ExponentialU}. For $|r|<1$, recall the identity,
\begin{align}
    %&\sum_{n=K}^{\infty} {r^n} = \frac{r^{K}}{1-r}\\
    %&\sum_{n=K}^{\infty} {n r^{n-1}} = \frac{Kr^{K-1} - r^K(K+1)}{(1-r)^2}\\
    \sum_{n=\ell}^{\infty} {nr^n = \frac{\ell - r(\ell+1)}{(1-r)^2}r^\ell} \to 0,\;\s{as}\; \ell \to \infty.\label{eqn:GeoSer}
\end{align}
Accordingly, let $r_{k,j}\triangleq\max_{z\in\calZ}S_{k,j}(z)$. Since we assume that for any $k\neq j$ and any $z\in\calZ$, the transition probability distributions $\alpha_k(\cdot\vert z)$ and $\alpha_j(\cdot\vert z)$ are not the same, it follows that $\max_{k\neq j}\max_{z\in\calZ}S_{k,j}(z)<1$. Thus $\max_{k\neq j}r_{k,j}<1$, and we get,
\begin{align}
    \E [\s{T}_{\s{MSPRT}} | \s{T}_{\s{MSPRT}} \geq \ell,\calH_k]\leq \sum_{j: j \neq k} \sqrt{\frac{\pi_j}{\pi_k}\frac{M-1}{\min_{\ell\in[M]} a_\ell}} \frac{\ell - r_{k,j}(\ell+1)}{(1-r_{k,j})^2}r_{k,j}^\ell,
\end{align}
which clearly converges to zero, as $\ell\to\infty$. Thus, by definition, we get that $\s{T}_{\s{MSPRT}}$ is uniformly integrable, which concludes the proof.

\subsection{Proof of Theorem~\ref{th:gnn_bounded_stop_time}}

\begin{proof}
The layout of this proof is similar to that of Theorem~\ref{th:bounded_stop_time}.
We again begin by noting that $\s{T}_{\s{GNN}}\leq\hat{\s{T}}$ with probability one, where
\begin{align}
\hat{\s{T}}\triangleq\inf\ppp{\ell\in\mathbb{N}:\prod_{i=1}^{\ell}\frac{\varphi_j(\bar{\s{Z}}_i)}{\varphi_k(\bar{\s{Z}}_i)}< \frac{\min_{\ell\in[M]} a_\ell}{M-1} \quad \forall j \neq k}.
\end{align}
Then, using the same arguments that lead to \eqref{eqn:FirstUpperBoundTailTMSPRT0}, we get,
\begin{align}
    \P(\s{T}_{\s{GNN}}>t\vert\calH_k)&\leq
\sum_{j: j \neq k} \sqrt{\frac{\pi_j}{\pi_k}\frac{M-1}{\min_{\ell\in[M]} a_\ell}}\cdot \s{V}^t_{k,j},\label{eqn:GNNUniformBound}
\end{align}
where
\begin{align}
    \s{V}^t_{k,j} \triangleq\bE\left[\left.\sqrt{\frac{\Phi_j(\bar{\s{Z}}_1^t)}{\Phi_k(\bar{\s{Z}}_1^t)}}\right|\calH_k \right]
\end{align}
Recall that
\begin{align}
    \s{U}^{(t)}_{k,j} &\triangleq \bE\left[\left. \sqrt{\frac{f_j(\bar{\s{Z}}_1^t)}{f_k(\bar{\s{Z}}_1^t)}}\right|\calH_k \right].
\end{align}
Then, we have,
\begin{align}
\s{V}^t_{k,j} &= \bE\left[\left. \sqrt{\frac{\Phi_j(\bar{\s{Z}}_1^t)}{\Phi_k(\bar{\s{Z}}_1^t)}}\right|\calH_k \right]\\
&=\int \sqrt{\frac{\Phi_j(\s{z}_1^t)}{\Phi_k(\s{z}_1^t)}} f_k(\s{z}_1^t) \mathrm{d}\s{z}_1^t\\
&=\int \sqrt{\frac{\Phi_j(\s{z}_1^t)}{\Phi_k(\s{z}_1^t)}}\sqrt{\frac{f_j(\s{z}_1^t)f_k(\s{z}_1^t)}{f_j(\s{z}_1^t)f_k(\s{z}_1^t)}} f_k(\s{z}_1^t) \mathrm{d}\s{z}_1^t\\
&=\int \sqrt{\frac{\Phi_j(\s{z}_1^t)f_k(\s{z}_1^t)}{\Phi_k(\s{z}_1^t)f_j(\s{z}_1^t)}} \sqrt{f_j(\s{z}_1^t)f_k(\s{z}_1^t)} \mathrm{d}\s{z}_1^t\\
&\leq \sqrt{\xi} \int \sqrt{f_j(\s{z}_1^t)f_k(\s{z}_1^t)} \mathrm{d}\s{z}_1^t\\
&= \sqrt{\xi}\cdot \s{U}^{(t)}_{k,j},\label{eqn:xiUpperBoundHeling}
\end{align}
where the inequality follows from the definition of $\xi$ in \eqref{eqn:xidef}. Now, using exactly the same steps that lead to \eqref{eqn:ExponentialU}, we may deduce that,
\begin{align}
  \s{V}^t_{k,j}\leq   \sqrt{\xi}\cdot \pp{\max_{z\in\bar\calZ}\bar{S}_{k,j}(z)}^t,\label{eqn:ExponentialU2}
\end{align}
where
\begin{align}
\bar{S}_{k,j}(z)&\triangleq1-\calH^2(\bar{\alpha}_k(\cdot \vert z), \bar{\alpha}_j(\cdot\vert z))\\
&= \int_{z'\in\bar\calZ}\sqrt{\alpha_k(z' \vert z)\alpha_j(z' \vert z)}\mathrm{d}z'.
\end{align}
Accordingly, we obtain that,
\begin{align}
\P(\s{T}_{\s{GNN}}>t\vert\calH_k)&\leq \sqrt{\xi}\cdot\sum_{j: j \neq k} \sqrt{\frac{\pi_j}{\pi_k}\frac{M-1}{\min_{\ell\in[M]} a_\ell}} \cdot\pp{\max_{z\in\bar\calZ}\bar{S}_{k,j}(z)}^t\\
&\leq \sqrt{\xi}(M-1)^{3/2}\max_{j\neq k}\sqrt{\frac{\pi_j}{\pi_k\min_{\ell\in[M]} a_\ell}}\pp{\max_{z\in\bar\calZ}\bar{S}_{k,j}(z)}^t\\
& = \s{C}_1\cdot\exp\p{-\s{C}_2\cdot t},\label{eqn:expofast2}
%\P(\s{T}_{\s{MSPRT}}>t\vert\calH_k)&\leq \sum_{j: j \neq k} \sqrt{\frac{\pi_j}{\pi_k}\frac{M-1}{\min_{l} a_l}} \cdot\max_{\calF_{t-1}}\pp{\max_{\ell}S_\ell}^t\\
%&\leq (M-1)^{3/2}\max_{j\neq k}\sqrt{\frac{\pi_j}{\pi_k\min_{l} a_l}}\max_{\s{Z}_{u_t}}\pp{\max_{\ell}S_\ell}^t.\label{eqn:expofast}
\end{align}
where $\s{C}_1\triangleq\sqrt{\xi}(M-1)^{3/2}\max_{j\neq k}\sqrt{\frac{\pi_j}{\pi_k\min_{\ell\in[M]} a_\ell}}$ and $\s{C}_2\triangleq \max_{j\neq k}\max_{z\in\bar\calZ}\log\frac{1}{\bar{S}_{k,j}(z)}$. Recall that we assume that $\xi$ is finite, and thus $\s{C}_1<\infty$. Furthermore, we assume that $\max_{k\neq j}\max_{z\in\calZ}\bar{S}_{k,j}(z)<1$, which implies that $\s{C}_2>0$. Therefore, the right-hand-side of \eqref{eqn:expofast2} decays exponentially fast with $t$, which concludes the proof.
\end{proof}

\subsection{Proof of Theorem~\ref{th:gnn_error_guarantee}}
\begin{proof}
We follow the proof of \cite[Theorem 4.2]{340472}. Since conditioned on $\calH_k$, the stop timing $\s{T}_{\s{GNN}}$ is finite, we may write,
\begin{align}
\calE^{\s{GNN}}_{k} &= \sum_{n=1}^\infty \P(\s{accept}\;\calH_k, \s{T}_{\s{GNN}}=n\vert\calH_k)\\
&= \sum_{n=1}^\infty \int_{\s{accept}\;\calH_k, \s{T}=n} f_k(z_1^n) \mathrm{d}z_1^n\\
&= \sum_{n=1}^\infty \int_{\s{accept}\;\calH_k, \s{T}=n} \frac{\Phi_k(z_1^n)}{\Phi_k(z_1^n)}f_k(z_1^n) \mathrm{d}z_1^n\\
&\geq \frac{1}{1+a_k} \sum_{j\in[M]} \sum_{n=1}^{\infty} \int_{\s{accept}\;\calH_k, \s{T}=n} \frac{\Phi_k(z_1^n)}{\Phi_j(z_1^n)}f_k(z_1^n) \mathrm{d}z_1^n\\
&= \frac{1}{1+a_k} \sum_{j\in[M]} \sum_{n=1}^{\infty} \int_{\s{accept}\;\calH_k, \s{T}=n} \frac{\Phi_k(z_1^n)}{\Phi_j(z_1^n)}\frac{f_j(z_1^n)}{f_j(z_1^n)}f_k(z_1^n) \mathrm{d}z_1^n\\
&= \frac{1}{1+a_k} \sum_{j\in[M]} \sum_{n=1}^{\infty} \int_{\s{accept}\;\calH_k, \s{T}=n} \frac{\Phi_k(z_1^n)}{\Phi_j(z_1^n)}\frac{f_k(z_1^n)}{f_j(z_1^n)}f_j(z_1^n) \mathrm{d}z_1^n\\
&\geq \xi^{-1} \frac{1}{1+a_k} \sum_{j\in[M]} \sum_{n=1}^{\infty} \int_{\s{accept}\;\calH_k, \s{T}=n} f_j(z_1^n) \mathrm{d}z_1^n\\
&= \xi^{-1} \frac{1}{1+a_k} \sum_{j\in[M]} \calE_{j,k}^{\s{GNN}}\label{eqn:errB0}\\
&= \xi^{-1} \frac{1}{1+a_k} (\calE_{k, k}^{\s{GNN}} + \calE_{k}^{\s{GNN}}),
\end{align}
where the first inequality follows from the GNN SDR in \eqref{eqn:decproblem2}, and the second inequality follows from the definition of $\xi$ in \eqref{eqn:xidef}. We therefore conclude that,
\begin{align}
\calE_k^{\s{GNN}} &\leq \calE_{k,k}^{\s{GNN}} \pp{(1+a_k)\xi - 1}\\
&\leq a_k \xi + \xi -1,
\end{align}
where we have used the fact that $\calE_{k,k}^{\s{GNN}}\leq1$, for any $k\in[M]$. This proves \eqref{eqn:errb1}, from which \eqref{eqn:errb2} follows trivially. Finally, we prove \eqref{eqn:errb3}. To that end, if we assume $a_\ell = \nu$, for all $\ell\in[M]$, then from \eqref{eqn:errB0}, we get that,
\begin{align}
\calE^{\s{GNN}} &= 1 - \sum_{k\in[M]} \calE_{k}^{\s{GNN}} \\
&\leq 1 - \frac{1}{\xi(1+\nu)} \sum_{k\in[M]} \sum_{j\in[M]} \calE_{j,k}^{\s{GNN}}\\
&=1 - \frac{1}{\xi(1+\nu)}.
\end{align}
\end{proof}

\subsection{Proof of Theorem~\ref{th:gnn_asymptotic_time}}
As in the proof of Theorem~\ref{th:asymptotic_time}, to prove Theorem~\ref{th:gnn_asymptotic_time} we first establish a few auxiliary results.
\begin{lemma}\label{lem:gnn_aep}
Let $\bar{\s{Z}}_1^\ell$ be a sequence sampled from $\P(\cdot\vert\calH_k)$. Let us assume that the length of each disjoint path $P\in\calP_\ell$, tends to $\infty$ as $\ell\to\infty$. If $\xi<\infty$, then,
\begin{align}
\frac{1}{\ell}\log\frac{\Phi_k(\bar{\s{Z}}_1^\ell)}{\Phi_j(\bar{\s{Z}}_1^\ell)} \xrightarrow[\ell \to \infty]{} d_{\s{KL}}(\bar\alpha_k||\bar\alpha_j\vert\pi_k^{\s{stat}}),
\end{align}
$f_k$-almost surely.
\end{lemma}
\begin{proof}
We note that,
\begin{align}
\frac{1}{\ell}\log\frac{\Phi_k(\bar{\s{Z}}_1^\ell)}{\Phi_j(\bar{\s{Z}}_1^\ell)}
&=\frac{1}{\ell}\log\left[\frac{\Phi_j(\bar{\s{Z}}_1^\ell)}{\Phi_k(\bar{\s{Z}}_1^\ell)}\frac{f_j(\bar{\s{Z}}_1^\ell)}{f_j(\bar{\s{Z}}_1^\ell)}\frac{f_k(\bar{\s{Z}}_1^\ell)}{f_k(\bar{\s{Z}}_1^\ell)}\right]\\
&=\frac{1}{\ell}\log\left[\frac{\Phi_j(\bar{\s{Z}}_1^\ell)}{\Phi_k(\bar{\s{Z}}_1^\ell)}\frac{f_k(\bar{\s{Z}}_1^\ell)}{f_j(\bar{\s{Z}}_1^\ell)}\right] + \frac{1}{\ell}\log\frac{f_k(\bar{\s{Z}}_1^\ell)}{f_j(\bar{\s{Z}}_1^\ell)}.
%&= \frac{1}{\ell}\log\left[\frac{\Phi_j(\bar{\s{Z}}_1^\ell)}{\Phi_k(\bar{\s{Z}}_1^\ell)}\frac{f_k(\bar{\s{Z}}_1^\ell)}{f_j(\bar{\s{Z}}_1^\ell)}\right] + \frac{1}{\ell}\sum_{i=1}^{\ell} \log{\frac{\bar{\alpha}_j(\s{Z}_{i} \vert \calA_i)}{\bar{\alpha}_k(\s{Z}_{i} \vert \calA_i)}}\\
%&\xrightarrow[\ell \to \infty]{}d_{\s{KL}}(\bar{\alpha}_k||\bar{\alpha}_j\vert\pi_k^{\s{stat}}),
\end{align}
Since we assume that $\xi<\infty$, it is clear that
\begin{align}
    \lim_{\ell\to\infty}\frac{1}{\ell}\log\frac{\Phi_k(\bar{\s{Z}}_1^\ell)}{\Phi_j(\bar{\s{Z}}_1^\ell)} = \lim_{\ell\to\infty}\frac{1}{\ell}\log\frac{f_k(\bar{\s{Z}}_1^\ell)}{f_j(\bar{\s{Z}}_1^\ell)}.
\end{align}
While at this point it is tempting to apply the Markov AEP property in Lemma~\ref{lem:AEP}, this is not possible, because its proof rely on the fact that $\{\s{Z}_\ell\}_{\ell\geq0}$ is a Markov chain on finite state space, while here we deal with a Markov process $\{\bar{\s{Z}}_\ell\}_{\ell\geq0}$ defined on (possibly) uncountable alphabets. Nonetheless, an AEP property for such processes (and, in fact, even much more general ones) is known. Indeed, using \cite[Theorem 7.5.1]{Gray11} (see, also, \cite[Theorem 437]{CosmaAryeh}), we have that,
\begin{align}
    \lim_{\ell\to\infty}\frac{1}{\ell}\log\frac{f_k(\bar{\s{Z}}_1^\ell)}{f_j(\bar{\s{Z}}_1^\ell)} = d_{\s{KL}}(\bar\alpha_k||\bar\alpha_j\vert\pi_k^{\s{stat}}),
\end{align}
$f_k$-almost surely. 
\end{proof}

\begin{lemma}\label{lem:gnn_time}
Fix $k\in[M]$. Recall the definition of $\s{T}_{\s{GNN}}$ in \eqref{eqn:decproblem2}. Assume that $\xi<\infty$, and
\begin{align}
%\exists j\in[\s{M}]: d_{\s{KL}}(\alpha_k(\cdot\vert z)||\alpha_{j_k^\star}(\cdot\vert z)) < \infty \quad\forall z\in\calZ
\min_{j\neq k\in[M]}\max_{z\in\calZ} \chi^2(\bar{\alpha}_k(\cdot\vert z)||\bar{\alpha}_{j}(\cdot\vert z)) < \infty\label{eqn:finite_stat_kl_div2}.
\end{align}
Then,
\begin{align}
    \s{T}_{\s{GNN}}\to\infty,
\end{align}
$f_k$-almost surely, as $\norm{a}_\infty \to 0$.
\end{lemma}

\begin{proof}[Proof of Lemma~\ref{lem:gnn_time}]
We follow the proof of Lemma~\ref{lem:lemma_time}, and as so we skip some of the more straightforward steps. Fix $n\in\mathbb{N}$, and let $j_k^\star$ be a hypothesis $j\in[\s{M}]$ which achieves the minimum in \eqref{eqn:finite_stat_kl_div2}. Using the same arguments that lead to \eqref{eqn:unionDouble}, we obtain,
\begin{align}
   \P (\s{T}_{\s{GNN}}<n \vert \calH_k) &= \P \left(\left.\exists \ell\in[M]: \max_{1 \leq m \leq n} \frac{\Phi_\ell(\bar{\s{Z}}_1^m)}{\sum_j \Phi_j(\bar{\s{Z}}_1^m)} > \frac{1}{1+a_\ell} \right|\calH_k\right)\\
   &\leq\sum_{\ell \neq k} \P \left(\left.\max_{1 \leq m \leq n} \frac{\Phi_\ell(\bar{\s{Z}}_1^m)}{\Phi_k(\bar{\s{Z}}_1^m)} > \frac{1}{a_\ell} \right|\calH_k\right)\nonumber\\
&\quad\quad\quad\quad+ \P \left(\left.\max_{1 \leq m \leq n} \frac{\Phi_k(\bar{\s{Z}}_1^m)}{\Phi_{j_k^\star}(\bar{\s{Z}}_1^m)} > \frac{1}{a_k} \right|\calH_k\right).\label{eqn:unionDouble2}
\end{align}
Applying  Markov's inequality on the first term at the right-hand-side of \eqref{eqn:unionDouble2}, we get for any $\ell\neq k$,
\begin{align}
\P \left(\left.\max_{1 \leq m \leq n} \frac{\Phi_\ell(\bar{\s{Z}}_1^m)}{\Phi_k(\bar{\s{Z}}_1^m)} > \frac{1}{a_\ell} \right|\calH_k\right)
&\leq \sum_{m=1}^na_\ell\E\pp{\left.\frac{\Phi_\ell(\bar{\s{Z}}_1^m)}{\Phi_k(\bar{\s{Z}}_1^m)}\right|\calH_k}\\
& \leq n\cdot a_\ell\cdot\xi,
\end{align}
where the last equality is due to the fact that
\begin{align}
\E\pp{\left.\frac{\Phi_\ell(\bar{\s{Z}}_1^m)}{\Phi_k(\bar{\s{Z}}_1^m)}\right|\calH_k} &= \E\pp{\left.\frac{\Phi_\ell(\bar{\s{Z}}_1^m)f_k(\bar{\s{Z}}_1^m)}{\Phi_k(\bar{\s{Z}}_1^m)f_\ell(\bar{\s{Z}}_1^m)}\cdot\frac{f_\ell(\bar{\s{Z}}_1^m)}{f_k(\bar{\s{Z}}_1^m)}\right|\calH_k}\\
&\leq \xi\cdot\E\pp{\left.\frac{f_\ell(\bar{\s{Z}}_1^m)}{f_k(\bar{\s{Z}}_1^m)}\right|\calH_k}\\
& = \xi,
\end{align}
where the inequality follows from the definition of $\xi$ in \eqref{eqn:xidef}. Thus, the first term at the right-hand-side of \eqref{eqn:unionDouble2} can be upper bounded as,
\begin{align}
    \sum_{\ell \neq k} \P \left(\left.\max_{1 \leq m \leq n} \frac{\Phi_\ell(\bar{\s{Z}}_1^m)}{\Phi_k(\bar{\s{Z}}_1^m)} > \frac{1}{a_\ell} \right|\calH_k\right)\leq n\cdot\xi\cdot(M-1)\cdot\norm{a}_\infty,\label{eqn:secConvA2GNN}
\end{align}
which in light of the fact that $\xi<\infty$, goes to zero, as $\norm{a}_\infty\to0$. 

Next, let us analyze the second term in  \eqref{eqn:unionDouble2}. Applying Markov's inequality, we have,
\begin{align}
\P \left(\left.\max_{1 \leq m \leq n} \frac{\Phi_k(\bar{\s{Z}}_1^m)}{\Phi_{j_k^\star}(\bar{\s{Z}}_1^m)} > \frac{1}{a_k} \right|\calH_k\right)&\leq\sum_{m=1}^n\P \left(\left.\frac{\Phi_k(\bar{\s{Z}}_1^m)}{\Phi_{j_k^\star}(\bar{\s{Z}}_1^m)} > \frac{1}{1a_k} \right|\calH_k\right)\\
&\leq \sum_{m=1}^na_k\E\pp{\left.\frac{\Phi_k(\bar{\s{Z}}_1^m)}{\Phi_{j_k^\star}(\bar{\s{Z}}_1^m)}\right|\calH_k}\\
&\leq \xi\sum_{m=1}^na_k\E\pp{\left.\frac{f_k(\bar{\s{Z}}_1^m)}{f_{j_k^\star}(\bar{\s{Z}}_1^m)}\right|\calH_k}\\
&=\xi\cdot a_k\cdot\sum_{m=1}^n\pp{1+\chi^2(f_k(\bar{\s{Z}}_1^m),f_{j_k^\star}(\bar{\s{Z}}_1^m))},
%&=\sum_{m=1}^n\P{\left.\sum_{i=1}^m\log\frac{\alpha_k(\s{Z}_i\vert\calA_i)}{\alpha_{j_k^\star}(\s{Z}_i\vert\calA_i)} > \log\frac{\pi_k a_k}{\pi_{j_k^\star}} \right|\calH_k}.
\end{align}
where the third inequality follow from the definition of $\xi$ in \eqref{eqn:xidef}. Next, we note that,
\begin{align}
    1+\chi^2(f_k(\bar{\s{Z}}_1^m),f_{j_k^\star}(\bar{\s{Z}}_1^m)) &= \int_{z_1^m\in\bar{\calZ}^m}\frac{f_k^2(z_1^m)}{f_{j_k^\star}(z_1^m)}\mathrm{d}z_1^m\\
    & = \int_{z_1^{m-1}\in\bar{\calZ}^{m-1}}\mathrm{d}z_1^{m-1}\prod_{i=1}^{m-1}\frac{\bar{\alpha}_k^2(z_i\vert\calA_i)}{\bar{\alpha}_{j_k^\star}(z_i\vert \calA_i)}\int_{z_m\in\bar{\calZ}}\frac{\bar{\alpha}_k^2(z_m\vert\calA_m)}{\bar{\alpha}_{j_k^\star}(z_m\vert \calA_m)}\mathrm{d}z_m\\
    &\leq \int_{z_1^{m-1}\in\bar{\calZ}^{m-1}}\mathrm{d}z_1^{m-1}\prod_{i=1}^{m-1}\frac{\bar{\alpha}_k^2(z_i\vert\calA_i)}{\bar{\alpha}_{j_k^\star}(z_i\vert \calA_i)}\max_{z'\in\calZ}\int_{z_m\in\bar{\calZ}}\frac{\bar{\alpha}_k^2(z_m\vert z')}{\bar{\alpha}_{j_k^\star}(z_m\vert z')}\mathrm{d}z_m\\
    & = \int_{z_1^{m-1}\in\bar{\calZ}^{m-1}}\mathrm{d}z_1^{m-1}\prod_{i=1}^{m-1}\frac{\bar{\alpha}_k^2(z_i\vert\calA_i)}{\bar{\alpha}_{j_k^\star}(z_i\vert \calA_i)}\cdot\max_{z'\in\calZ}\chi^2(\bar{\alpha}_k(\cdot\vert z'),\bar{\alpha}_{j_k^\star}(\cdot\vert z'))\\
    &\leq\cdots\\
    &\leq\pp{\max_{z'\in\calZ}\chi^2(\bar{\alpha}_k(\cdot\vert z'),\bar{\alpha}_{j_k^\star}(\cdot\vert z'))}^m\\
    &\leq \s{C}^m,
\end{align}
where the last inequality follows from the assumption in \eqref{eqn:finite_stat_kl_div2}, which implies that $\max_{z'\in\calZ}\chi^2(\bar{\alpha}_k(\cdot\vert z'),\bar{\alpha}_{j_k^\star}(\cdot\vert z'))\leq\s{C}$, for some $\s{C}>0$. Therefore,
\begin{align}
    \P \left(\left.\max_{1 \leq m \leq n} \frac{\Phi_k(\bar{\s{Z}}_1^m)}{\Phi_{j_k^\star}(\bar{\s{Z}}_1^m)} > \frac{1}{a_k} \right|\calH_k\right)&\leq \xi\p{\sum_{m=1}^n\s{C}^m}a_k\\
    &\leq \xi\p{\sum_{m=1}^n\s{C}^m}\norm{a}_\infty,\label{eqn:secConvAGNN}
\end{align}
which goes to zero, as $\norm{a}_\infty\to0$. Therefore, combining \eqref{eqn:unionDouble2}, \eqref{eqn:secConvA2GNN}, and \eqref{eqn:secConvAGNN}, we obtain that $\s{T}_{\s{GNN}}\to\infty$ in probability as $\norm{a}_\infty\to0$. Now, convergence in probability implies that there must be a sub-sequence of $\s{T}_{\s{GNN}}$ which converge to infinity $f_k$-almost surely. Because $\s{T}_{\s{GNN}}$ is non-decreasing as each $a_\ell\to0$, we may conclude that $\s{T}_{\s{GNN}}\to\infty$ $f_k$-almost surely, as $\norm{a}_\infty\to0$.

\end{proof}

\begin{lemma}\label{lem:gnn_time_k}
Fix $k\in[M]$, and let
\begin{align}
    \bar{\s{T}}_k \triangleq \inf\ppp{n\in\mathbb{N}:\frac{ \Phi_k(\s{Z}_1^n)}{\sum_{j=0}^{M-1}\Phi_j(\s{Z}_1^n)} > \frac{1}{1+a_k}}.
\end{align}
Assume that $\xi<\infty$ and
\begin{align}
\min_{j\neq k\in[M]}\max_{z\in\calZ} \chi^2(\alpha_k(\cdot\vert z),\alpha_{j}(\cdot\vert z)) < \infty\label{eqn:finite_stat_kl_div_kGNN}.
\end{align}
Then,
\begin{align}
    \lim_{\norm{a}_\infty \to 0} \frac{\bar{\s{T}}_k}{- \log a_k} = \frac{1}{\min_{j:j \neq k} d_{\s{KL}}(\bar{\alpha}_k||\bar{\alpha}_j\vert\pi_k^{\s{stat}})},
\end{align}
$f_k$-almost surely.
\end{lemma}

\begin{proof}[Proof of Lemma~\ref{lem:gnn_time_k}]
For simplicity of notation we define $\bar{p}_k^{(n)} \triangleq \frac{ \Phi_k(\s{Z}_1^n)}{\sum_{j=0}^{M-1}\Phi_j(\s{Z}_1^n)}$. Note that $\bar{\s{T}}_k \geq \s{T}_{\s{GNN}}$, because $\s{T}_{\s{GNN}} = \min_{{j\in[M]}} \bar{\s{T}}_j$. Therefore, Lemma~\ref{lem:gnn_time} implies that $\bar{\s{T}}_k \to\infty$, as $\norm{a}_\infty \to 0$, for any $k\in[M]$. A little bit of straightforward algebra steps reveal that $\bar{\s{T}}_k$ can be represented as,
\begin{align}
    \bar{\s{T}}_k \triangleq \inf\ppp{n\in\mathbb{N}:\bar{\mathscr{W}}_n(\bar{\s{Z}}_1^n)> -\frac{\log a_k}{n}},
\end{align}
where
\begin{align}
    \bar{\mathscr{W}}_n(\bar{\s{Z}}_1^n)\triangleq-\frac{1}{n}\log{\sum_{j:j \neq k} \exp(-n\left[\frac{1}{n}\log\frac{\Phi_k(\bar{\s{Z}}_1^n)}{\Phi_j(\bar{\s{Z}}_1^n)}\right])}.
\end{align}
Then, at $n = \bar{\s{T}}_k$, we note that,
\begin{align}
\lim_{\norm{a}_\infty \to 0}\bar{\mathscr{W}}_{\bar{\s{T}}_k}(\s{Z}_1^{\bar{\s{T}}_k})& = \lim_{\bar{\s{T}}_k\to\infty}\bar{\mathscr{W}}_{\bar{\s{T}}_k}(\bar{\s{Z}}_1^{\bar{\s{T}}_k})\\
&=\lim_{\bar{\s{T}}_k\to\infty}-\frac{1}{\bar{\s{T}}_k}\log{\sum_{j: j \neq k} \exp(-\bar{\s{T}}_k\left[\frac{1}{\bar{\s{T}}_k}\log\frac{\Phi_k(\bar{\s{Z}}_1^{\bar{\s{T}}_k})}{\Phi_j(\bar{\s{Z}}_1^{\bar{\s{T}}_k})}\right])}\\
&=\min_{j:j \neq k} \lim_{\bar{\s{T}}_k\to\infty}\frac{1}{\bar{\s{T}}_k}\log\frac{\Phi_k(\bar{\s{Z}}_1^{\bar{\s{T}}_k})}{\Phi_j(\bar{\s{Z}}_1^{\bar{\s{T}}_k})}
\\
&=\min_{j:j \neq k} d_{\s{KL}}(\bar{\alpha}_k||\bar{\alpha}_j\vert\pi_k^{\s{stat}}),\label{eqn:KLabove3}
\end{align}
almost surely, where the first equality is a result of Lemma~\ref{lem:gnn_time}, the second equality follows from \cite[Lemma 5.2]{340472}, and the last equality is due to Lemma~\ref{lem:gnn_aep}. Next, at $n=\s{T}_k-1$ we by continuity, we have,
\begin{align}
\lim_{\norm{a}_\infty \to 0}\bar{\mathscr{W}}_{\bar{\s{T}}_k-1}(\s{Z}_1^{\s{T}_k-1})& = \lim_{\bar{\s{T}}_k\to\infty}\bar{\mathscr{W}}_{\bar{\s{T}}_k-1}(\s{Z}_1^{\bar{\s{T}}_k-1})\\
&=\lim_{\bar{\s{T}}_k\to\infty}-\frac{1}{\bar{\s{T}}_k}\log{\sum_{j: j \neq k} \exp(-\bar{\s{T}}_k\left[\frac{1}{\bar{\s{T}}_k}\log\frac{\Phi_k(\bar{\s{Z}}_1^{\bar{\s{T}}_k-1})}{\Phi_j(\bar{\s{Z}}_1^{\bar{\s{T}}_k-1})}\right])}\\
&=\min_{j:j \neq k} \lim_{\bar{\s{T}}_k\to\infty}\frac{1}{\bar{\s{T}}_k}\log\frac{\Phi_k(\s{Z}_1^{\bar{\s{T}}_k-1})}{\Phi_j(\s{Z}_1^{\bar{\s{T}}_k-1})}\\
&=\min_{j:j \neq k} d_{\s{KL}}(\bar{\alpha}_k||\bar{\alpha}_j\vert\pi_k^{\s{stat}}),\label{eqn:KLbelow4}
\end{align}
almost surely. Therefore, since $\bar{\s{T}}_k$ is defined as the first time $n$ for which $\bar{p}_k^{(n)} > \frac{1}{1+a_k}$, we have the following two inequalities simultaneously,
\begin{align}
\bar{\mathscr{W}}_{\bar{\s{T}}_k}(\bar{\s{Z}}_1^{\bar{\s{T}}_k}) &\geq -\frac{\log a_k}{\bar{\s{T}}_k},\label{eqn:asymp3}\\
\bar{\mathscr{W}}_{\bar{\s{T}}_k-1}(\bar{\s{Z}}_1^{\bar{\s{T}}_k-1}) &\leq -\frac{\log a_k}{\bar{\s{T}}_k-1}.\label{eqn:asymp4}
\end{align}
Applying the squeeze theorem on \eqref{eqn:asymp3}--\eqref{eqn:asymp4}, and using \eqref{eqn:KLabove3} and \eqref{eqn:KLbelow4}, we finally obtain that,
\begin{align}
    \frac{\bar{\s{T}}_k}{- \log a_k} \to \frac{1}{\min_{j:j \neq k} d_{\s{KL}}(\bar{\alpha}_k||\bar{\alpha}_j\vert\pi_k^{\s{stat}})},
\end{align}
almost surely w.r.t. $f_k$, as $\norm{a}_\infty \to 0$, which concludes the proof.
\end{proof}

We are now in a position to prove Theorem \ref{th:gnn_asymptotic_time}. Specifically, this theorem guarantees both almost-sure in expectation types of convergence. We begin with the former. To that end, we note that for all $\epsilon>0$,
\begin{align}
&\P\pp{\left.\left|\frac{\s{T}_{\s{GNN}}}{-\log a_k} - d_{\s{KL}}(\bar{\alpha}_k||\bar{\alpha}_j\vert\pi_k^{\s{stat}})\right| > \epsilon\right|\calH_k}=\nonumber\\
%&= \sum_{l=0}^{\s{M}-1} \P{\left|\frac{\s{T}_{\s{MSPRT}}}{-\log a_k} - d_{\s{KL}}(\alpha_k||\alpha_j\vert\pi_k^{\s{stat}})\right| > \epsilon \text{ and accept } \calH_l}\\
&\quad\quad=\P{\left.\left|\frac{\s{T}_k}{-\log a_k} - d_{\s{KL}}(\bar{\alpha}_k||\bar{\alpha}_j\vert\pi_k^{\s{stat}})\right| > \epsilon\right|\calH_k}\nonumber\\
&\quad\quad\quad\quad+ \sum_{\ell\neq k} \P{\left.\left|\frac{\s{T}_{\s{GNN}}}{-\log a_k} - d_{\s{KL}}(\bar{\alpha}_k||\bar{\alpha}_j\vert\pi_k^{\s{stat}})\right| > \epsilon,\; \s{accept}\;\calH_\ell\right|\calH_k}\\
&\quad\quad\leq \P{\left.\left|\frac{\s{T}_k}{-\log a_k} - d_{\s{KL}}(\bar{\alpha}_k||\bar{\alpha}_j\vert\pi_k^{\s{stat}})\right| > \epsilon\right|\calH_k} + \sum_{\ell\neq k} \P{\s{accept}\;\calH_\ell\vert\calH_k}\\
&\quad\quad\leq \P{\left.\left|\frac{\s{T}_k}{-\log a_k} - d_{\s{KL}}(\bar{\alpha}_k||\bar{\alpha}_j\vert\pi_k^{\s{stat}})\right| > \epsilon\right|\calH_k}+\xi\norm{a}_\infty+\xi-1,\label{eqn:inprobab2}
\end{align}
where the last inequality follows from 
Theorem~\ref{th:gnn_error_guarantee}. Now, Lemma~\ref{lem:gnn_time_k} implies that the first term at the right-hand-side of \eqref{eqn:inprobab2} converges to zero, as $\norm{a}_\infty\to0$, while the leftover terms converge to zero, as $\norm{a}_\infty\to0$ and $\xi\to1$. This proves the convergence in probability of $\frac{\s{T}_{\s{GNN}}}{-\log a_k}$ to $d_{\s{KL}}(\bar{\alpha}_k||\bar{\alpha}_j\vert\pi_k^{\s{stat}})$. Now, using the same arguments as in the proof of Lemma~\ref{lem:gnn_time_k}, since $\s{T}_{\s{GNN}}$ is non-decreasing, as $\norm{a}_\infty\to0$, we also have $\frac{\s{T}_{\s{GNN}}}{-\log a_k}$ converges $f_k$-almost surely to $d_{\s{KL}}(\bar{\alpha}_k||\bar{\alpha}_j\vert\pi_k^{\s{stat}})$.

As for convergence in expectation, we will prove that $\frac{\s{T}_{\s{GNN}}}{-\log a_k}$ is uniformly integrable, and then the almost sure convergence will imply the convergence in expectation \cite[Theorem 16.4]{Bill86}. We have,
%\paragraph{Denote:} $r \triangleq \max_{j; Z_0}{\sqrt{1-\calH^2 (f_k(Z_2,Z_1|Z_0), f_j(Z_2,Z_1|Z_0)))}}$
\begin{align}
    \E [\s{T}_{\s{GNN}} | \s{T}_{\s{GNN}} \geq \ell,\calH_k] &= \sum_{t=\ell}^{\infty} {t\cdot\P(\s{T}_{\s{GNN}}=t\vert\calH_k)}\\
    &\leq \sum_{t=\ell}^{\infty} {t\cdot\P(\s{T}_{\s{GNN}} \geq t\vert\calH_k)}\\
    &\leq  \sum_{t=\ell}^{\infty} t \sum_{j: j \neq k} \sqrt{\frac{\pi_j}{\pi_k}\frac{M-1}{\min_{\ell\in[M]} a_\ell}} \cdot\s{V}^{(t)}_{k,j} \\
    &\leq \sqrt{\xi}\sum_{j: j \neq k} \sqrt{\frac{\pi_j}{\pi_k}\frac{M-1}{\min_{\ell\in[M]} a_\ell}} \sum_{t=\ell}^{\infty} t\cdot\s{U}^{(t)}_{k,j}\\
    &\leq \sqrt{\xi}\sum_{j: j \neq k} \sqrt{\frac{\pi_j}{\pi_k}\frac{M-1}{\min_{\ell\in[M]} a_\ell}} \sum_{t=\ell}^{\infty} t\pp{\max_{z\in\bar\calZ}\bar{S}_{k,j}(z)}^t,
\end{align}
where the second inequality follows from \eqref{eqn:GNNUniformBound}, the second inequality is due to \eqref{eqn:xiUpperBoundHeling}, and the last inequality follows from \eqref{eqn:ExponentialU2}. Let $\bar{r}_{k,j}\triangleq\max_{z\in\bar{\calZ}}\bar{S}_{k,j}(z)$. Since we assume that $\max_{k\neq j}\max_{z\in\bar{\calZ}}\bar{S}_{k,j}(z)<1$, we get,
\begin{align}
    \E [\s{T}_{\s{GNN}} | \s{T}_{\s{GNN}} \geq \ell,\calH_k]\leq \sqrt{\xi}\sum_{j: j \neq k} \sqrt{\frac{\pi_j}{\pi_k}\frac{M-1}{\min_{\ell\in[M]} a_\ell}} \frac{\ell - \bar{r}_{k,j}(\ell+1)}{(1-\bar{r}_{k,j})^2}\bar{r}_{k,j}^\ell,
\end{align}
which clearly converges to zero, as $\ell\to\infty$. Thus, by definition, we get that $\s{T}_{\s{GNN}}$ is uniformly integrable, which concludes the proof.

\section{Conclusion} \label{section:Conc}
This paper introduces multiclass information flow detection algorithms based on a realistic probabilistic model of information propagation over social media networks modeled by graphs. The learning task is to minimizes a risk defined as a combination of the classification error and the detection time. Our first algorithm is based on the well-known MSPRT, while the other is a novel graph neural network based sequential decision algorithm. For both algorithm we prove several statistical guarantees. Extensive experiments over two real-world datasets demonstrate that these algorithms outperform other state-of-the-art misinformation detection algorithms. 

%\section*{Acknowledgments}

%This work was supported by the ISRAEL SCIENCE FOUNDATION (grant No. 1734/21).

\bibliographystyle{ieeetr}
\bibliography{bibfile}

\begin{thebibliography}{10}

\bibitem{Harsin2018PostTruthAC}
J.~Harsin, ``Post-truth and critical communication studies,'' {\em Oxford Research Encyclopedia of Communication}, 2018.

\bibitem{Arendt2005-ARETAP}
H.~Arendt, ``Truth and politics,'' in {\em Truth} (J.~Medina and D.~Wood, eds.), pp.~295--314, Blackwell, 2005-01-01.

\bibitem{iq}
D.~Fallis, {\em The Varieties of Disinformation}, pp.~135--161.
\newblock 07 2014.

\bibitem{Kwon13}
S.~Kwon, M.~Cha, K.~Jung, W.~Chen, and Y.~Wang, ``Prominent features of rumor propagation in online social media,'' in {\em 2013 IEEE 13th International Conference on Data Mining}, pp.~1103--1108, 2013.

\bibitem{10.1007/s11042-020-10183-2}
R.~K. Kaliyar, A.~Goswami, and P.~Narang, ``Fakebert: Fake news detection in social media with a bert-based deep learning approach,'' {\em Multimedia Tools Appl.}, vol.~80, p.~11765–11788, mar 2021.

\bibitem{DBLP:journals/corr/RiedelASR17}
B.~Riedel, I.~Augenstein, G.~P. Spithourakis, and S.~Riedel, ``A simple but tough-to-beat baseline for the fake news challenge stance detection task,'' {\em CoRR}, vol.~abs/1707.03264, 2017.

\bibitem{DBLP:journals/corr/KipfW16}
T.~N. Kipf and M.~Welling, ``Semi-supervised classification with graph convolutional networks,'' {\em CoRR}, vol.~abs/1609.02907, 2016.

\bibitem{DBLP:journals/corr/abs-1810-00826}
K.~Xu, W.~Hu, J.~Leskovec, and S.~Jegelka, ``How powerful are graph neural networks?,'' {\em CoRR}, vol.~abs/1810.00826, 2018.

\bibitem{dou2021user}
Y.~Dou, K.~Shu, C.~Xia, P.~S. Yu, and L.~Sun, ``User preference-aware fake news detection,'' in {\em Proceedings of the 44th international ACM SIGIR conference on research and development in information retrieval}, pp.~2051--2055, 2021.

\bibitem{DBLP:journals/corr/abs-1902-06673}
F.~Monti, F.~Frasca, D.~Eynard, D.~Mannion, and M.~M. Bronstein, ``Fake news detection on social media using geometric deep learning,'' {\em CoRR}, vol.~abs/1902.06673, 2019.

\bibitem{devlin2019bertpretrainingdeepbidirectional}
J.~Devlin, M.-W. Chang, K.~Lee, and K.~Toutanova, ``Bert: Pre-training of deep bidirectional transformers for language understanding,'' 2019.

\bibitem{honnibal_spacy_2018}
M.~Honnibal and I.~Montani, ``{spaCy} 2: {Natural} language understanding with {Bloom} embeddings, convolutional neural networks and incremental parsing,'' {\em To appear}, 2017.

\bibitem{veličković2018graphattentionnetworks}
P.~Veličković, G.~Cucurull, A.~Casanova, A.~Romero, P.~Liò, and Y.~Bengio, ``Graph attention networks,'' 2018.

\bibitem{DBLP:journals/corr/HamiltonYL17}
W.~L. Hamilton, R.~Ying, and J.~Leskovec, ``Inductive representation learning on large graphs,'' {\em CoRR}, vol.~abs/1706.02216, 2017.

\bibitem{wei2019quickstop}
H.~Wei, X.~Kang, W.~Wang, and L.~Ying, ``Quickstop: A markov optimal stopping approach for quickest misinformation detection,'' {\em Proceedings of the ACM on Measurement and Analysis of Computing Systems}, vol.~3, no.~2, pp.~1--25, 2019.

\bibitem{10.1214/aoms/1177730197}
A.~Wald and J.~Wolfowitz, ``{Optimum Character of the Sequential Probability Ratio Test},'' {\em The Annals of Mathematical Statistics}, vol.~19, no.~3, pp.~326 -- 339, 1948.

\bibitem{orenloberman2023online}
M.~Oren-Loberman, V.~Azar, and W.~Huleihel, ``Online auditing of information flow,'' {\em arXiv:2310.14595}, 2023.

\bibitem{bayesian}
J.~Mockus, {\em Bayesian Approach to Global Optimization}, vol.~37, pp.~473--481.
\newblock 01 2006.

\bibitem{DBLP:journals/corr/abs-2002-04397}
Y.~Ren and J.~Zhang, ``{HGAT:} hierarchical graph attention network for fake news detection,'' {\em CoRR}, vol.~abs/2002.04397, 2020.

\bibitem{10020234}
U.~Jeong, K.~Ding, L.~Cheng, R.~Guo, K.~Shu, and H.~Liu, ``Nothing stands alone: Relational fake news detection with hypergraph neural networks,'' in {\em 2022 IEEE International Conference on Big Data (Big Data)}, (Los Alamitos, CA, USA), pp.~596--605, IEEE Computer Society, dec 2022.

\bibitem{DBLP:journals/corr/abs-1809-09401}
Y.~Feng, H.~You, Z.~Zhang, R.~Ji, and Y.~Gao, ``Hypergraph neural networks,'' {\em CoRR}, vol.~abs/1809.09401, 2018.

\bibitem{oren2023onlineconf}
M.~ren Loberman, V.~Azar, and W.~Huleihel, ``Online auditing of information flow,'' in {\em 2024 IEEE International Conference on Acoustics, Speech, and Signal Processing (ICASSP 2024)}, 2024.

\bibitem{340472}
C.~Baum and V.~Veeravalli, ``A sequential procedure for multihypothesis testing,'' {\em IEEE Transactions on Information Theory}, vol.~40, no.~6, pp.~1994--2007, 1994.

\bibitem{10.5555/3061053.3061153}
J.~Ma, W.~Gao, P.~Mitra, S.~Kwon, B.~J. Jansen, K.-F. Wong, and M.~Cha, ``Detecting rumors from microblogs with recurrent neural networks,'' in {\em Proceedings of the Twenty-Fifth International Joint Conference on Artificial Intelligence}, IJCAI'16, p.~3818–3824, AAAI Press, 2016.

\bibitem{10.5555/525960}
C.~M. Bishop, {\em Neural Networks for Pattern Recognition}.
\newblock USA: Oxford University Press, Inc., 1995.

\bibitem{RichardMichaelD.1991NNCE}
M.~D. Richard and R.~P. Lippmann, ``Neural network classifiers estimate bayesian a posteriori probabilities,'' {\em Neural computation}, vol.~3, no.~4, pp.~461--483, 1991.

\bibitem{alma990021928420204146}
D.~Koller and N.~Friedman, {\em Probabilistic graphical models : principles and techniques}.
\newblock Adaptive computation and machine learning, Cambridge, Mass: MIT Press, 2009.

\bibitem{monti2019fake}
F.~Monti, F.~Frasca, D.~Eynard, D.~Mannion, and M.~M. Bronstein, ``Fake news detection on social media using geometric deep learning,'' {\em arXiv:1902.06673}, 2019.

\bibitem{DBLP:journals/corr/abs-1711-05101}
I.~Loshchilov and F.~Hutter, ``Fixing weight decay regularization in adam,'' {\em CoRR}, vol.~abs/1711.05101, 2017.

\bibitem{Norris_1997}
J.~R. Norris, {\em Markov Chains}.
\newblock Cambridge Series in Statistical and Probabilistic Mathematics, Cambridge University Press, 1997.

\bibitem{bibaut2020sufficientinsufficientconditionsstochastic}
A.~F. Bibaut, A.~Luedtke, and M.~J. van~der Laan, ``Sufficient and insufficient conditions for the stochastic convergence of ces\`{a}ro means,'' 2020.

\bibitem{10.5555/1146355}
T.~M. Cover and J.~A. Thomas, {\em Elements of Information Theory (Wiley Series in Telecommunications and Signal Processing)}.
\newblock USA: Wiley-Interscience, 2006.

\bibitem{Bill86}
P.~Billingsley, {\em Probability and Measure}.
\newblock John Wiley and Sons, second~ed., 1986.

\bibitem{Gray11}
R.~M. Gray, {\em Entropy and Information Theory}.
\newblock Springer Publishing Company, Incorporated, 2nd~ed., 2011.

\bibitem{CosmaAryeh}
C.~R. Shalizi and A.~Kontorovich, {\em Almost None of the Theory of Stochastic Processes}.
\newblock A Course on Random Processes, for Students of Measure-Theoretic Probability, with a View to Applications in Dynamics and Statistics, 2007.

\end{thebibliography}
%24, 22, 32

%%%%%%%%%%%%%%%%%%%%%%%%%%%%%%%%%%%%%%%%%%%%%%%%%%%%%%%%%%%%%%%%%

%possible dataset: https://github.com/KaiDMML/FakeNewsNet

%\bibliographystyle{alpha}
%\bibliography{bibfile}

\end{document}